\documentclass[twocolumn]{svjour3}

\usepackage[latin1]{inputenc}
\usepackage{times}

\usepackage{psfrag, amsmath, amssymb, amscd, amsfonts, latexsym, epsf, graphicx}
\usepackage{natbib}

\usepackage[labelformat=simple]{subcaption}

\def\bbbr{{\mathbb R}} 
\def\bbbz{{\mathbb Z}}

\def\sep{\scriptsize\mbox{sep}}
\def\nonsep{\scriptsize\mbox{non-sep}}
\def\sampl{\scriptsize\mbox{sampl}}
\def\normsampl{\scriptsize\mbox{normsampl}}
\def\intdisc{\scriptsize\mbox{int}}
\def\disc{\scriptsize\mbox{disc}}
\def\boxfilt{\scriptsize\mbox{box}}
\def\norm{\scriptsize\mbox{norm}}
\def\relscale{\scriptsize\mbox{relscale}}
\def\cascade{\scriptsize\mbox{cascade}}
\def\scaleestrel{\scriptsize\mbox{scaleest,rel}}
\def\scaleref{\scriptsize\mbox{ref}}
\def\blob{\scriptsize\mbox{blob}}
\def\edge{\scriptsize\mbox{edge}}
\def\ridge{\scriptsize\mbox{ridge}}
\def\refscale{\scriptsize\mbox{ref}}
\def\acc{\scriptsize\mbox{acc}}
\def\aff{\scriptsize\mbox{aff}}
\def\affsampl{\scriptsize\mbox{affsampl}}
\def\affint{\scriptsize\mbox{affint}}
\def\hybrnormsampl{\scriptsize\mbox{hybr-sampl}}
\def\hybrint{\scriptsize\mbox{hybr-int}}

\journalname{arXiv preprint}

\begin{document}

\title{\bf Discrete approximations of Gaussian smoothing and Gaussian derivatives%
\thanks{The support from the Swedish Research Council 
              (contracts 2022-02969) is gratefully acknowledged. }}

\titlerunning{Discrete approximations of Gaussian smoothing and Gaussian derivatives}

\author{Tony Lindeberg}

\institute{Tony Lindeberg \at
              Computational Brain Science Lab,
              Division of Computational Science and Technology,
              KTH Royal Institute of Technology,
              SE-100 44 Stockholm, Sweden.
              \email{tony@kth.se}}

\date{Received: date / Accepted: date}

\maketitle

\begin{abstract}
\noindent
This paper develops an in-depth treatment concerning the problem of
approximating the Gaussian smoothing and Gaussian derivative
computations in scale-space theory for application on discrete data. With close connections to
previous axiomatic treatments of continuous and discrete scale-space theory,
we consider three main ways of discretizing these scale-space operations
in terms of explicit discrete convolutions, based on either
(i)~sampling the Gaussian kernels and the Gaussian derivative kernels,
(ii)~locally integrating the Gaussian kernels and the Gaussian
derivative kernels over each pixel support region, to aim at
suppressing some of the severe artefacts of sampled Gaussian kernels
and sampled Gaussian derivatives at very fine scales, or
(iii)~basing the scale-space analysis on the discrete analogue of the
Gaussian kernel, and then computing derivative approximations by
applying small-support central difference operators to the spatially
smoothed image data.

We study the properties of these three main
discretization methods both theoretically and experimentally, and
characterize their performance by quantitative measures, including the
results they give rise to with respect to the task of scale selection,
investigated for four different use cases, and with emphasis on the
behaviour at fine scales. The results show that the
sampled Gaussian kernels and the sampled Gaussian derivatives
as well as the integrated
Gaussian kernels and the integrated Gaussian derivatives
perform very poorly at very fine scales.
At very fine scales, the discrete analogue of the Gaussian kernel with
its corresponding discrete derivative approximations performs
substantially better. The sampled Gaussian kernel and the sampled
Gaussian derivatives do, on the other hand, lead to numerically very
good approximations of the corresponding continuous results, when the
scale parameter is sufficiently large, in most of the experiments presented in
the paper, when the scale parameter is greater than a value of about 1, 
in units of the grid spacing. Below a standard deviation of about
0.75, the derivative estimates obtained from convolutions with
the sampled Gaussian derivative kernels
are, however, not numerically accurate or consistent, while the results
obtained from the discrete analogue of the Gaussian kernel, with its
associated central difference operators applied to the spatially
smoothed image data, is then a much better choice.

\end{abstract}

\section{Introduction}

When operating on image data, the earliest layers of image operations
are usually expressed in terms of receptive fields, which means that
the image information is integrated over local support regions in image
space.
For modelling such operations, the notion of scale-space theory
(Iijima \citeyear{Iij62};
 Witkin \citeyear{Wit83};
 Koenderink \citeyear{Koe84};
 Koenderink and van Doorn \citeyear{KoeDoo87-BC,KoeDoo92-PAMI};
 Lindeberg \citeyear{Lin93-Dis,Lin94-SI,Lin10-JMIV};
 Florack \citeyear{Flo97-book};
 Sporring {\em et al.\/}\ \citeyear{SpoNieFloJoh96-SCSPTH};
 Weickert {\em et al.\/}\ \citeyear{WeiIshImi99-JMIV};
 ter Haar Romeny \citeyear{Haa04-book})
stands out as a principled theory, by which the shapes of the
receptive fields can be determined from axiomatic derivations,
that reflect desirable theoretical properties of the first stages of
visual operations.

In summary, this theory states that convolutions with Gaussian kernels
and Gaussian derivative constitutes a canonical class of image
operations as a first layer of visual processing.
Such spatial receptive fields, or approximations thereof,
can, in turn, be used as
basis for expressing a large variety of image operations, both in
classical computer vision
(Lindeberg \citeyear{Lin97-IJCV,Lin98-IJCV,Lin12-JMIV,Lin15-JMIV},
Bretzner and Lindeberg \citeyear{BL97-CVIU},
Schiele and Crowley \citeyear{SchCro00-IJCV},
Chomat {\em et al.\/} \citeyear{chomat:00},
Mikolajczyk and Schmid \citeyear{MikSch04-IJCV},
Lowe \citeyear{Low04-IJCV},
Bay {\em et al.\/} \citeyear{BayEssTuyGoo08-CVIU},
Tuytelaars and Mikolajczyk \citeyear{TuyMik08-Book},
Linde and Lindeberg \citeyear{LinLin12-CVIU})
and more recently in deep learning
(Jacobsen {\em et al.\/} \citeyear{JacGemLouSme16-CVPR},
Lindeberg \citeyear{Lin21-SSVM,Lin22-JMIV},
Pintea {\em et al.\/} \citeyear{PinTomGoeLooGem21-IP},
Sangalli {\em et al.\/} \citeyear{SanBluVelAng22-BMVC},
Penaud-Polge {\em et al.\/} \citeyear{PenVelAng22-ICIP},
Gavilima-Pilataxi and Ibarra-Fiallo \citeyear{GavIva23-ICPRS}).

The theory for the notion of scale-space representation does, however,
mainly concern continuous image data, while implementations of this
theory on digital computers requires a discretization over image space.
The subject of this article is to describe and compare a number of
basic approaches for discretizing the Gaussian convolution operation,
as well as convolutions with Gaussian derivatives.

While one could possibly argue that at sufficiently coarse scales,
where sampling effects ought to be small, the influence of choosing
one form of discrete implementation compared to some other ought to be
negligible, or at least of minor effect, there are situations where it is
desirable to apply scale-space operations at rather fine scales, and
then also to be reasonably
sure that one would obtain desirable response properties of the
receptive fields.

One such domain, and which motivates
the present deeper study of discretization effects for Gaussian
smoothing operations and Gaussian derivative computations at fine
scales, is when applying Gaussian derivative operations in deep
networks, as done in a recently developed subdomain of deep learning
(Jacobsen {\em et al.\/} \citeyear{JacGemLouSme16-CVPR},
Lindeberg \citeyear{Lin21-SSVM,Lin22-JMIV},
Pintea {\em et al.\/} \citeyear{PinTomGoeLooGem21-IP},
Sangalli {\em et al.\/} \citeyear{SanBluVelAng22-BMVC},
Penaud-Polge {\em et al.\/} \citeyear{PenVelAng22-ICIP},
Gavilima-Pilataxi and Ibarra-Fiallo \citeyear{GavIva23-ICPRS}).

A practical observation, that one may make, when working with deep
learning, is that deep networks may tend to have a preference to
computing image representations at very fine scale levels.
For example, empirical results indicate that deep networks often tend to perform image
classification based on very fine-scale image information,
corresponding to the local image texture on the surfaces of objects in
the world. Indirect support for such a view may also be taken from the
now well-established fact that deep networks may be very sensitive to adversarial
perturbations, based on adding deliberately designed noise patterns of
very low amplitude to the image data
(Szegedy {\em et al.\/} \citeyear{SzeZarSutBruErhGooFer13-arXiv},
Moosavi-Dezfooli {\em et al.\/} \citeyear{MooFawFawFro17-CVPR},
 Athalye {\em et al.\/} \citeyear{AthEngIlyKwo18-ICML}, 
 Baker {\em et al.\/} \citeyear{BakLuErlKel18-CompBiol},
 Geirhos {\em et al.\/} \citeyear{GeiRubMicBetWicBre18-arXiv},
 Hermann  {\em et ak.\/} \citeyear{HerCheKor20-NeurIPS},
 Hendrycks {\em et ak.\/} \citeyear{HenZhaBasSteSon21-CVPR},
 Veerabadran {\em et al.\/} \citeyear{VeeGolShaChePapKurGooShlSohMozGam23-NatureComm}).
That observation demonstrates that deep networks may be very strongly
influenced by fine-scale structures in the input image.
Another observation may be taken from working with deep networks based
on using Gaussian derivative kernels as the filter weights. If one
designs such a network with complementary training of the scale levels
for the Gaussian derivatives, then a common result is that the network
will prefer to base its decisions based on receptive fields at rather
fine scale levels.

When to implement such Gaussian derivative networks in practice, one
hence faces the need for being able to go below the rule of thumb for
classical computer vision, of not attempting to operate below a
certain scale threshold, where the standard deviation of the Gaussian
derivative kernel should then not be below a value of say $1/\sqrt{2}$ or 1,
in units of the grid spacing.

From a viewpoint of theoretical signal processing, one may possibly
take the view to argue that one should use the sampling theorem to express a lower
bound on the scale level, which one should then never go below.
For regular images, as obtained from digital cameras, or already
acquired data sets as compiled by the computer vision community, such an
approach based on the sampling theorem, is, however, not fully possible in
practice. First of all, we almost never, or at least very rarely, have
explicit information about the sensor characteristics of the image
sensor. Secondly, it would hardly be possible to model the imaging
process in terms of an ideal bandlimited filter with frequency
characteristics near the spatial sampling density of the image sensor.
Applying an ideal bandpass filter to an already given digital image
may lead to ringing phenomena near the discontinuities in
the image data, which
will lead to far worse artefacts for spatial image data than for
{\em e.g.\/} signal transmission over information carriers in terms of
sine waves.

Thus, the practical problem that one faces when designing and applying
a Gaussian derivative network to image data, is to in a practically
feasible manner express a spatial smoothing process, that can smooth a
given digital input image for any fine scale of the discrete
approximation to a Gaussian derivative filter. A theoretical problem,
that then arises, concerns how to design such a process, so that it can
operate from very fine scale levels, possibly starting even at scale
level zero corresponding to the original input data,
without leading to severe discretization artefacts.

A further technical problem that arises is that, even if one would
take the {\em a priori\/} view of basing the implementation on the
purely discrete theory for scale-space smoothing and scale-space
derivatives developed in (Lindeberg \citeyear{Lin90-PAMI,Lin93-JMIV}),
and as we have taken in our previous work on Gaussian derivative networks
(Lindeberg \citeyear{Lin21-SSVM,Lin22-JMIV}),
one then faces the problem of handling the special mathematical functions used
as smoothing primitives in this theory (the modified Bessel functions
of integer order) when propagating gradients for training deep
networks backwards by automatic differentiation, when performing
learning of the scale levels in the network. These necessary mathematical
primitives do not exist as built-in functions in {\em e.g.\/} PyTorch
(Paszke {\em et al.\/} \citeyear{PasGroChiChaYanDeVLinDesAntLer17-NIPS}),
which implies that the user then would have to implement a PyTorch
interface for these functions himself, or choose some other type of
discretization method, if aiming to learn the scale levels in the
Gaussian derivative networks by back propagation.
There are a few related studies of discretizations of scale-space
operations (Wang \citeyear{Wan00-TIP},
Lim and Stiehl \citeyear{LimSti03-ScSp},
Tschirsich and Kuijper \citeyear{TscKui15-JMIV},
Slav{\'\i}k and Stehl{\'\i}k \citeyear{SlaSte15-JMathAnalAppl},
Rey-Otero and Delbracio \citeyear{OteDel16-IPOL}).
These do not, however, answer the questions that need to be addressed
for the intended use cases for our developments.

Wang (\citeyear{Wan00-TIP}) proposed pyramid-like algorithms for computing multi-scale
  differential operators using a spline technique, however, then
  taking rather coarse steps in the scale direction.
Lim and Stiehl (\citeyear{LimSti03-ScSp}) studied properties of discrete scale-space
  representations under discrete iterations in the scale direction,
  based on Euler's forward method. For our purpose, we do, however, need
  to consider the scale direction as a continuum.
Tschirsich and Kuijper (\citeyear{TscKui15-JMIV}) investigated the compatibility of
  topological image descriptors with a discrete scale-space
  representations, and did also derive an eigenvalue decomposition in
  relation to the semi-discrete diffusion equation, that determines the
  evolution properties over scale, to enable efficient computations of
  discrete scale-space representations of the same image at multiple
  scales. With respect to our target application area, we are,
  however, more interested in computing image features based on
  Gaussian derivative responses, and then mostly also computing 
  discrete scale-space representation at a single scale only, for each input image.
  Slav{\'\i}k and Stehl{\'\i}k  (\citeyear{SlaSte15-JMathAnalAppl})
  developed a theory for more
  general evolution equations over semi-discrete domains, which
  incorporates the 1-D discrete scale-space evolution family, that we
  consider here, as corresponding to convolutions with the discrete
  analogue of the Gaussian kernel, as a special
  case. For our purposes, we are, however, more interested in
  performing an in-depth study of different discrete approximations of the
  axiomatically determined class of Gaussian
  smoothing operations and Gaussian derivative operators, than
  expanding the treatment to other possible evolution equations over
  discrete spatial domains.
  
Rey-Otero and Delbracio (\citeyear{OteDel16-IPOL}) assumed that the
image data can be regarded as bandlimited, and did then use a Fourier-based approach
for performing closed-form Gaussian convolution over a reconstructed
Fourier basis, which in that way constitutes a way to eliminate the
discretization errors, provided that a correct reconstruction of an underlying
continuous image, notably before the image acquisition step, can be performed.%
\footnote{For a number of clarifications with respect to an evaluation of what
  Rey-Otero and Delbracio (\citeyear{OteDel16-IPOL}) refer to as
``Lindeberg's smoothing method'' in that work, see
Appendix~\ref{app-clarif-lindebergs-smoothing-method}.}
A very closely related approach, for computing Gaussian
convolutions based on a reconstruction of an assumed-to-be bandlimited
signal, has also been previously outlined by
{\AA}str{\"o}m and Heyden (\citeyear{AstHey97-SCSPTH}).

As argued earlier in this introduction, the image
data that is processed in computer vision are, however, not generally
accompanied with characteristic information regarding the image
acquisition process, specifically not with regard to what extent
the image data could be regarded as bandlimited. Furthermore, one could question
if the image data obtained from a modern camera sensor could at all be modelled as
bandlimited, for a cutoff frequency very near the resolution of the
image. Additionally, with regard to our target application domain of
deep learning, one could also question if it would be manageable to
invoke a Fourier-based image reconstruction step for each convolution
operation in a deep network. In the work to be developed here, we are, on the other hand,
more interested in developing a theory for discretizing
the Gaussian smoothing and the Gaussian derivative operations at very
fine levels of scale, in terms of explicit convolution operations, and
based on as minimal as possible assumptions regarding the nature of the image data.
  
The purpose of this article is thus to perform a detailed theoretical analysis
of the properties of different discretizations of the Gaussian
smoothing operation and Gaussian derivative computations at any scale,
and with emphasis on reaching as near as possible to the desirable
theoretical properties of the underlying scale-space representation, to
hold also at very fine scales for the discrete implementation.

For performing such analysis, we will consider basic
approaches for discretizing the Gaussian kernel in terms of either pure
spatial sampling (the sampled Gaussian kernel) or local integration
over each pixel support region (the integrated Gaussian kernel) and
compare to the results of a genuinely discrete scale-space theory
(the discrete analogue of the Gaussian kernel),
see Figure~\ref{fig-kernel-graphs} for graphs of such kernels.d
After analysing and
numerically quantifying the properties of these basic types of
discretizations, we will then extend the analysis to discretizations
of Gaussian derivatives in terms of either sampled Gaussian derivatives,
integrated Gaussian derivatives,
and compare to the results of a genuinely discrete theory based on
convolutions with the discrete analogue of the Gaussian kernel
followed by discrete derivative approximations computed by applying
small-support central difference operators to the discrete scale-space
representation. We will also extend the analysis to the computation of
local directional derivatives, as a basis for filter-bank approaches
for receptive fields, based on either the scale-space representation
generated by convolution with rotationally symmetric Gaussian kernels,
or the affine Gaussian scale space.

It will be shown that, with regard to the topic of raw Gaussian
smoothing, the discrete analogue of the Gaussian kernel has the best
theoretical properties, out of the discretization methods considered.
For scale values when the standard deviation of the continuous
Gaussian kernel is above 0.75 or 1, the sampled Gaussian kernel does
also have very good properties, and leads to very good approximations
of the corresponding fully continuous results. The integrated Gaussian
kernel is better at handling fine scale levels than the sampled
Gaussian kernel, but does, however, comprise a scale offset that hampers its
accuracy in approximating the underlying continuous theory.

Concerning the topic of approximating the computation of Gaussian
derivative responses, it will be shown that the approach based on
convolution with the discrete analogue of the Gaussian kernel followed
by central difference operations has the clearly best properties at
fine scales, out of the studied three main approaches. In fact, when
the standard deviation of the underlying continuous Gaussian kernel is
a bit below about 0.75, the sampled Gaussian derivative kernels and the
integrated Gaussian derivative kernels do not lead to accurate
numerical estimates of derivatives, when applied to monomials of the
same order as the order of spatial differentiation, or lower.
Over an intermediate scale range in the upper part of this scale
interval, the integrated Gaussian derivative kernels do, however, have
somewhat better properties than the sampled Gaussian derivative kernels.
For the discrete approximations of Gaussian derivatives defined from
convolutions with the discrete analogue of the Gaussian kernel
followed by central differences, the numerical estimates of
derivatives obtained by applying this approach to monomials of the
same order as the order of spatial differentiation do, on the other
hand, lead to derivative estimates exactly equal to their continuous
counterparts, and also over the entire scale range.

For larger scale values, for standard deviations greater than about 1,
relative to the grid spacing, in the
experiments to be reported in the paper, the discrete approximations
of Gaussian derivatives obtained from convolutions with sampled
Gaussian derivatives do on the other hand lead to numerically very
accurate approximations of the corresponding results obtained from the
purely continuous scale-space theory. For the discrete derivative
approximations obtained by convolutions with the integrated
Gaussian derivatives, the box integration introduces a scale offset,
that hampers the accuracy of the approximation of the corresponding
expressions obtained from the fully continuous scale-space theory.
The integrated Gaussian derivative kernels do, however, degenerate less seriously
than the sampled Gaussian derivative kernels within a certain range of
very fine scales. Therefore, they may constitute an interesting alternative, if
the mathematical primitives needed for the discrete
analogues of the Gaussian derivative are not fully available within
a given system for programming deep networks.

For simplicity, we do in this treatment restrict ourselves to image operations that operate in terms of
discrete convolutions only. In this respect, we do not consider
implementations in terms of Fourier transforms, which are also
possible, while less straightforward in the context of deep learning.
We do furthermore not consider extensions to spatial interpolation
operations, which operate between the positions of the image
pixels, and which can be highly useful, for example, for locating the
positions of image features with subpixel accuracy
(Unser {\em et al.\/} \citeyear{UnsAldEde91-PAMI,UnsAldEde93-SP},
Wang and Lee \citeyear{WanLee98-PAMI},
Bouma {\em et al.\/} \citeyear{BouVilBesRomGer07-ScSp},
Bekkers \citeyear{Bek20-ICLR},
Zheng {\em et al.\/} \citeyear{ZheGonYouTao22-IJCV}).
We do additionally not consider
representations that perform subsamplings at coarser scales, which
can be useful for reducing the amount of computational work
(Burt and Adelson \citeyear{BA83-COM},
Crowley \citeyear{Cro84-dolp},
Simoncelli {\em et al.\/} \citeyear{SimFreAdeHee92-IT},  
Simoncelli and Freeman \citeyear{SimFre-ICIP95},
Lindeberg and Bretzner \citeyear{LinBre03-ScSp},
Crowley and Riff \citeyear{CroRif03-ScSp},
Lowe \citeyear{Low04-IJCV}),
or representations that aim at speeding up the spatial
convolutions on serial computers based on performing the computations
in terms of spatial recursive filters 
(Deriche \citeyear{Der92-ICIP},
 Young and van~Vliet \citeyear{YouVlie95-SP},
 van Vliet {\em et al.\/} \citeyear{VliYouVer98-PR},
 Geusebroek {\em et al.\/} \citeyear{GeuSmeWei03-TIP},
 Farneb{\"a}ck and Westin \citeyear{FarWes06-JMIV},
 Charalampidis \citeyear{Cha16-SP}).
For simplicity, we develop the theory for the special cases of 1-D
signals or 2-D images, while extensions to higher-dimensional
volumetric images is straightforward, as implied by separable
convolutions for the scale-space concept based on convolutions with
rotationally symmetric Gaussian kernels.

Concerning experimental evaluations, we do in this paper deliberately
focus on and restrict ourselves to the theoretical properties of
different discretization methods, and only report performance measures
based on such theoretical properties. One motivation for this approach
is that the integration with different types of visual modules may
call for different relative properties of the discretization methods.
We therefore want this treatment to be timeless, and not biased to
the integration with particular computer vision methods or
algorithms that operate on the output from Gaussian smoothing
operations or Gaussian derivatives.
Experimental evaluations
with regard to Gaussian derivative networks will be reported in
follow-up work. 
The results from this theoretical analysis should therefore
be more generally applicable to larger variety of approaches in classical
computer vision, as well as to other deep learning approaches that
involve Gaussian derivative operators.

\section{Discrete approximations of Gaussian smoothing}
\label{disc-approx-gauss-smooth}

The Gaussian scale-space representation $L(x, y;\; s)$ of a 2-D
spatial image $f(x, y)$ is defined by convolution with 2-D Gaussian
kernels of different sizes
\begin{equation}
  \label{eq-2D-gauss-kern}
  g_{2D}(x, y;\; s) = \frac{1}{2 \pi s} \, e^{-(x^2 + y^2)/2s}
\end{equation}
according to (Iijima \citeyear{Iij62}, Witkin \citeyear{Wit83},
Koenderink \citeyear{Koe84}, Lindeberg \citeyear{Lin93-Dis,Lin10-JMIV}, Florack
\citeyear{Flo97-book}, Weickert {\em et al.\/}
\citeyear{WeiIshImi99-JMIV}, ter Haar Romeny {\em et al.\/} \citeyear{Haa04-book})
\begin{equation}
  \label{eq-2D-gauss-conv}
  L(x, y;\; s)
  = \int_{\xi \in \bbbr} \int_{\eta \in \bbbr}
         g_{2D}(\xi, \eta;\; s) \, f(x - \xi, y - \eta) \, d\xi \, d\eta.
\end{equation}
Equivalently, this scale-space representation can be seen as defined by
the solution of the 2-D diffusion equation
\begin{equation}
  \label{eq-cont-2d-diffusion-eq}
  \partial_s L
  = \frac{1}{2} \, ( \partial_{xx} L + \partial_{yy} L)
\end{equation}
with initial condition $L(x, y;\; 0) = f(x, y)$.

\subsection{Theoretical properties of Gaussian scale-space
  representation}
\label{sec-theor-props-gauss-scsp}

\subsubsection{Non-creation of new structure with increasing scale}
\label{sec-non-creat-struct}

The Gaussian scale space, generated by convolving an image with 
Gaussian kernels, obeys a number of special properties, that ensure that
the transformation from any finer scale level to any coarser scale
level is guaranteed to always correspond to a simplification of the image information:
\begin{itemize}
  \item
  {\bf Non-creation of local extrema:\/}
  For any one-dimensional signal $f$, it can be shown that the number
  of local extrema in the 1-D Gaussian scale-space representation at
  any coarser scale $s_2$ is guaranteed to not be higher than the
  number of local extrema at any finer scale $s_1 < s_2$.
\item
  {\bf Non-enhancement of local extrema:\/}
  For any $N$-dimen\-sional signal, it can be shown that the derivative
  of the scale-space representation with respect to the scale
  parameter $\partial_s L$ is guaranteed to obey $\partial_s L \leq 0$
  at any local spatial maximum point and $\partial_s L \geq 0$ at any
  local spatial minimum point. In this respect, the Gaussian convolution
  operation has a strong smoothing effect.
\end{itemize}
In fact, the Gaussian kernel can be singled out as the unique choice
of smoothing kernel as having these properties, from axiomatic derivations,
if combined with the requirement of a {\em semi-group property\/}
over scales
\begin{equation}
   \label{eq-semi-group-gauss}
   g_{2D}(\cdot, \cdot;\; s_1) * g_{2D}(\cdot, \cdot;\; s_2) = g_{2D}(\cdot, \cdot;\; s_1 + s_2)
\end{equation}
and certain regularity assumptions, see Theorem~5 in
(Lindeberg \citeyear{Lin90-PAMI}), Theorem~3.25 in
(Lindeberg \citeyear{Lin93-Dis}) and Theorem~5 in
(Lindeberg \citeyear{Lin10-JMIV}) for more specific statements.

For related treatments about theoretically principled scale-space
axiomatics, see also Koenderink (\citeyear{Koe84}),
Babaud {\em et al.\/} (\citeyear{BWBD86-PAMI}),
Yuille and Poggio (\citeyear{YuiPog86-PAMI}),
Koenderink and van Doorn (\citeyear{KoeDoo92-PAMI}),
Pauwels {\em et al.\/} (\citeyear{PauFidMooGoo95-PAMI}),
Lindeberg (\citeyear{Lin96-ScSp}),
Weickert {\em et al.\/} (\citeyear{WeiIshImi99-JMIV})
and Duits {\em et al.\/} (\citeyear{DuiFloGraRom04-JMIV}).

\subsubsection{Cascade smoothing property}

Due to the semi-group property, it follows that the scale-space
representation at any coarser scale $L(x, y;\; s_2)$ can be obtained by
convolving the scale-space representation at any finer scale $L(x, y;\; s_1)$
with a Gaussian kernel parameterized by the scale difference $s_2 - s_1$:
\begin{equation}
   \label{eq-casc-prop-raw-scsp}
   L(\cdot, \cdot;\; s_2) = g_{2D}(\cdot, \cdot;\; s_2 - s_1) * L(\cdot, \cdot;\; s_1).
\end{equation}
This form of {\em cascade smoothing property\/} is an essential
property of a scale-space representation, since it implies that the
transformation from any finer scale level $s_1$ to any coarser scale
level $s_2$ will always be a simplifying transformation, provided that
the convolution kernel used for the cascade smoothing operation
corresponds to a simplifying transformation.

\subsubsection{Spatial averaging}
\label{sec-spat-avg}

The Gaussian kernel is non-negative
\begin{equation}
  g_{2D}(x, y;\; s) \geq 0
\end{equation}
and normalized to unit $L_1$-norm
\begin{equation}
  \int_{(x, y) \in \bbbr^2} g_{2D}(x, y;\; s) = 1.
\end{equation}
In these respects, Gaussian smoothing corresponds to a spatial
averaging process, which constitutes one of the desirable attributes
of a smoothing process intended to reflect different spatial scales in
image data.

\subsubsection{Separable Gaussian convolution}

Due to the separability of the 2-D Gaussian kernel
\begin{equation}
  g_{2D}(x, y;\; s) = g(x;\; s) \, g(y;\; s),
\end{equation}
where the 1-D Gaussian kernel is of the form
\begin{equation}
  g(x;\; s) = \frac{1}{\sqrt{2 \pi s}} \, e^{-x^2/2s},
\end{equation}
the 2-D Gaussian convolution operation (\ref{eq-2D-gauss-conv})
can also be written as two separable 1-D convolutions of the form
\begin{multline}
   L(x, y;\; s) = \\ = \int_{\xi \in \bbbr} g(\xi;\, s) \, \left( \int_{\eta
       \in \bbbr} g(\eta;\; s) \, f(x - \xi, y - \eta) \, d\eta \right) \, d\xi.
\end{multline}
Methods that implement Gaussian convolution in terms of explicit
discrete convolutions usually exploit this separability property, since if the
Gaussian kernel is truncated%
\footnote{\label{footnote-trunction-error}
  In the implementations underlying this work, we truncate the
  Gaussian kernel at the tails, such that
  $2 \int_{x = N}^{\infty} g(x;\; s) \, dx \leq \epsilon$, for a small
  value of $\epsilon$, of the order of $10^{-8}$.
  The reason for using such a small value of $\epsilon$ in our
  experiments is to ensure
  that the experimental results are not negatively affected by truncation
  effects. For practical implementations of Gaussian smoothing, one
  can often use a larger value of $\epsilon$, chosen so as to give a
  suitable trade-off between computational accuracy and computational efficiency.}
at the tails for $x = \pm N$, the
computational work for separable convolution will be of the order
\begin{equation}
  W_{\sep} = 2 \, (2 N + 1)
\end{equation}
per image pixel, whereas it would be of order
\begin{equation}
  W_{\nonsep} = (2 N + 1)^2
\end{equation}
for non-separable 2-D convolution.

\subsection{Modelling situation for theoretical analysis of different
  approaches for implementing Gaussian smoothing discretely}

From now on, we will, for
simplicity, only consider the case with 1-D Gaussian convolutions of
the form
\begin{equation}
  \label{eq-1D-cont-gauss-conv}
   L(x;\; s) = \int_{\xi \in \bbbr} g(\xi;\, s) \, f(x - \xi) \, d\xi,
 \end{equation}
which are to be implemented in terms of discrete convolutions of the form
\begin{equation}
  \label{eq-1D-disc-gauss-conv}
   L(x;\; s) = \sum_{n \in \bbbz} \, T(n;\, s) \,  f(x - n),
\end{equation}
for some family of discrete filter kernels $T(n;\; s)$.

\subsubsection{Measures of the spatial extent of smoothing kernels}

The spatial extent of these 1-D kernels can be described by the scale
parameter $s$, which represents the spatial
variance of the convolution kernel
\begin{multline}
  V(g(\cdot;\; s)) = \\
  = \frac{\int_{x \in \bbbr} x^2 \, g(x;\; s) \, dx}{\int_{x \in \bbbr} g(x;\; s) \, dx}
  - \left(
        \frac{\int_{x \in \bbbr} x \, g(x;\; s) \, dx}{\int_{x \in \bbbr} g(x;\; s) \, dx}
     \right)^2 = s,
\end{multline}
and which can also be parameterized in
terms of the standard deviation
\begin{equation}
  \sigma = \sqrt{s}.
\end{equation}
For the discrete kernels, the spatial variance is correspondingly measured as
\begin{multline}
  V(T(\cdot;\; s)) = \\
  = \frac{\sum_{n \in \bbbz} n^2 \, T(n;\; s)}{\sum_{n \in \bbbz} T(n;\; s)}
  - \left(
    \frac{\sum_{n \in \bbbz} n \, T(n;\; s)}{\sum_{n \in \bbbz} T(n;\; s)}
  \right)^2.
\end{multline}

\begin{figure*}[hbtp]
  \begin{center}
    \begin{tabular}{cccc}
      {\em Continuous Gauss\/} & {\em Sampled Gauss\/}
      & {\em Integrated Gauss\/} & {\em Discrete Gauss\/} \\
      \includegraphics[width=0.22\textwidth]{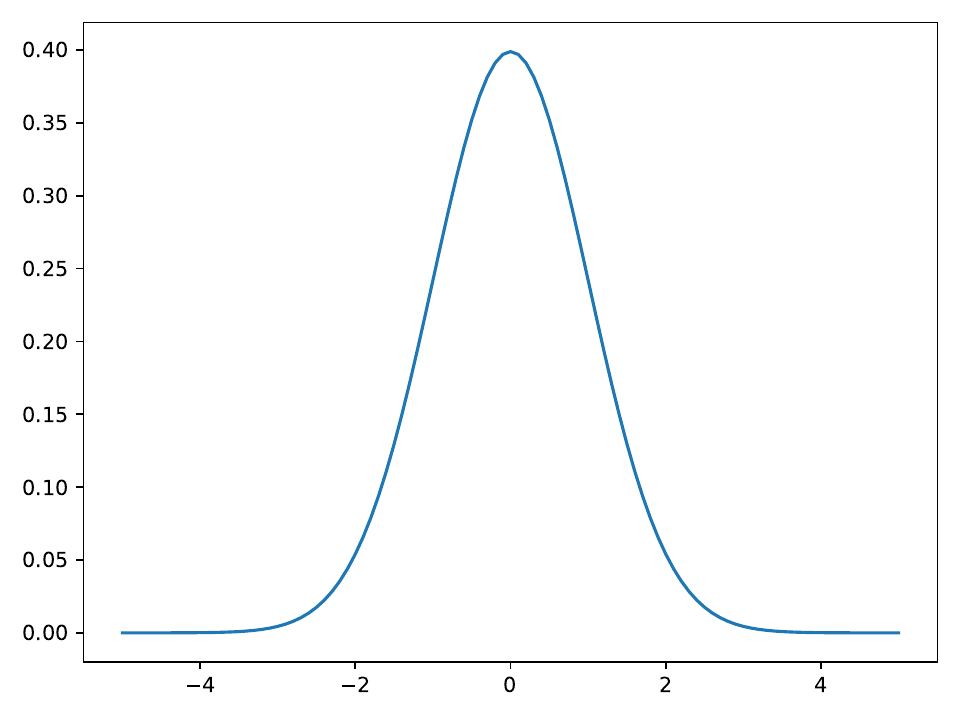}
      & \includegraphics[width=0.22\textwidth]{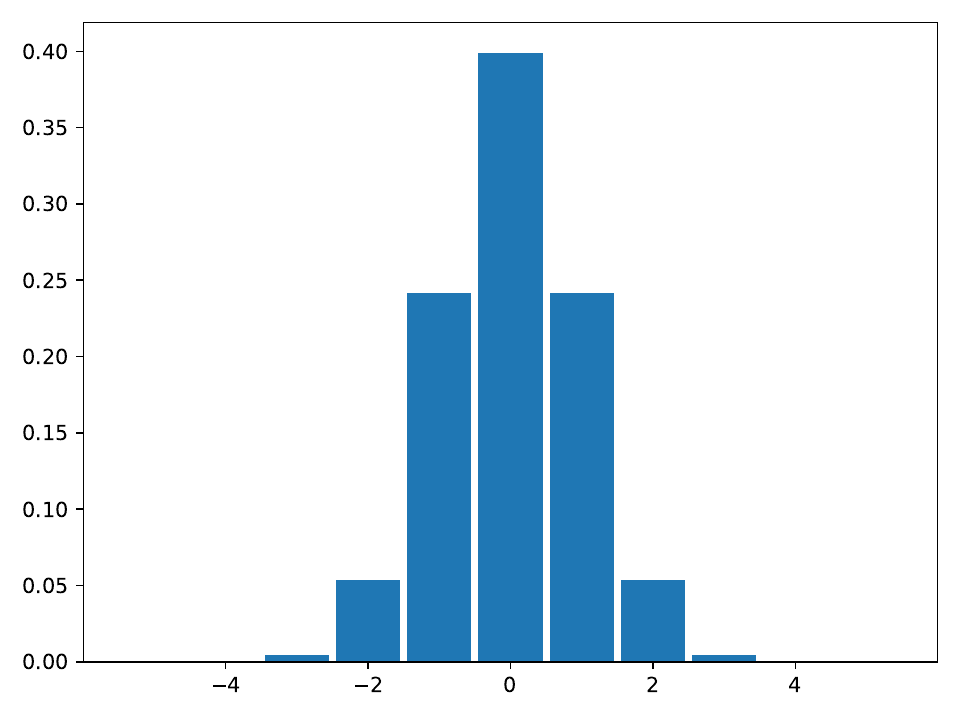}
      & \includegraphics[width=0.22\textwidth]{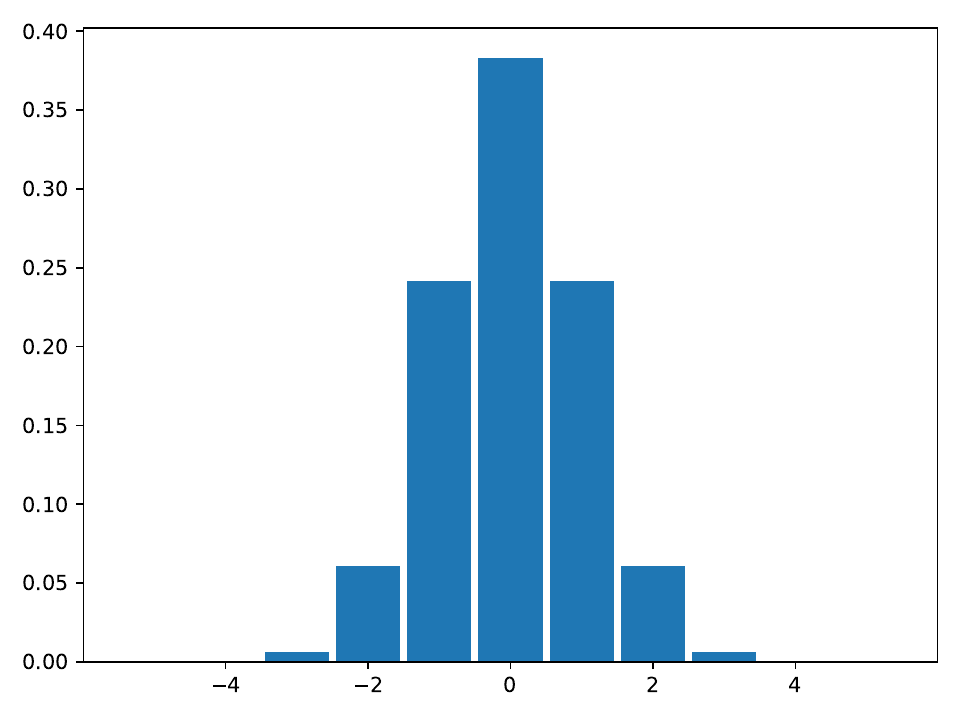}
      & \includegraphics[width=0.22\textwidth]{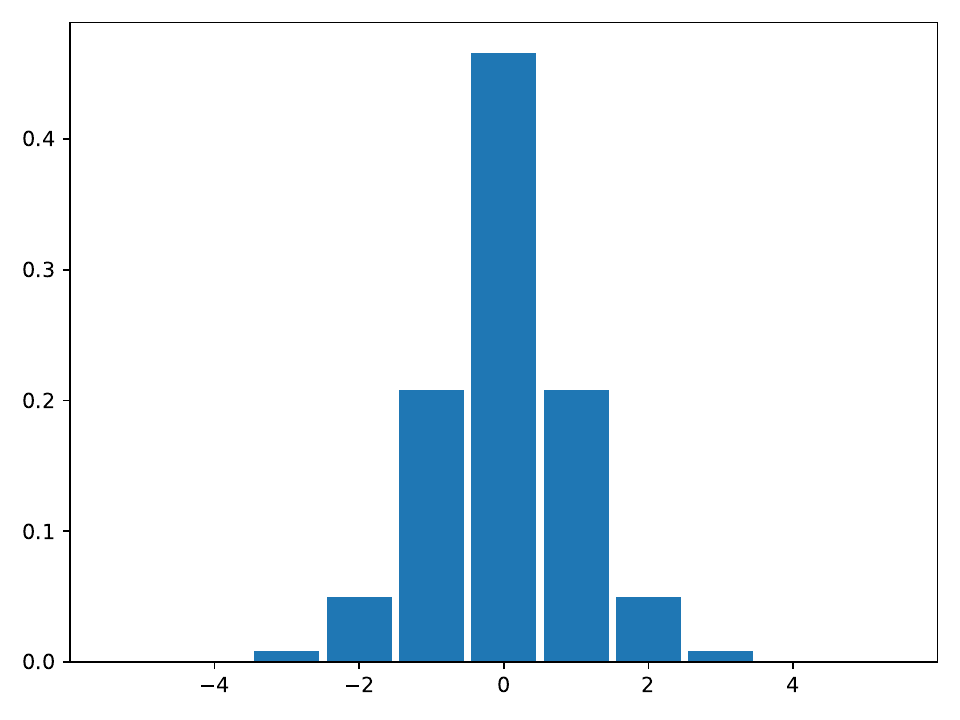} \\
      \includegraphics[width=0.22\textwidth]{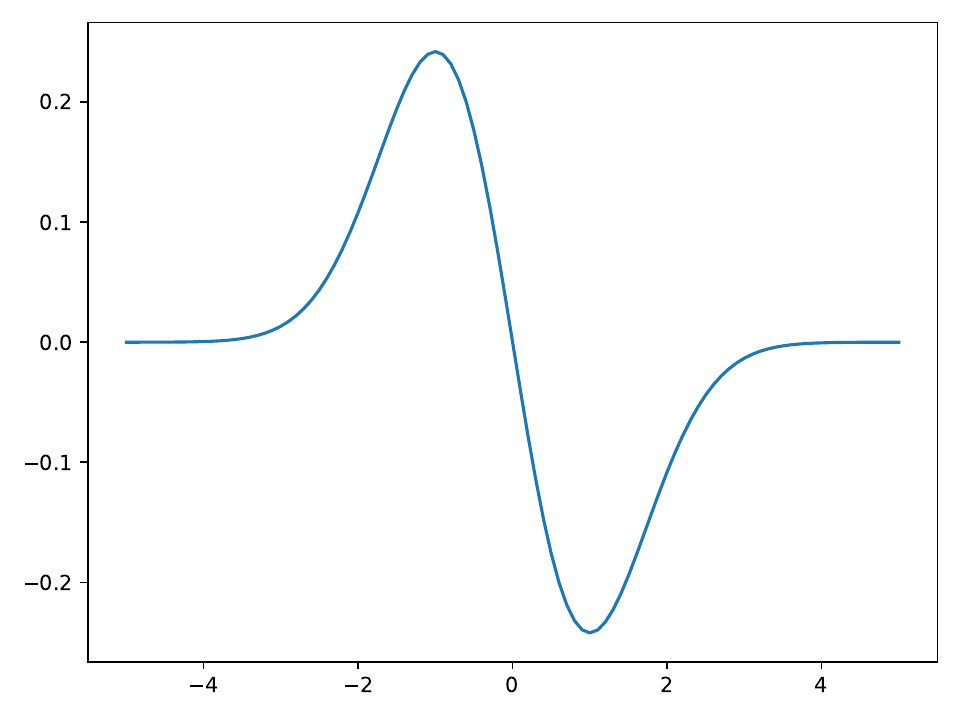}
      & \includegraphics[width=0.22\textwidth]{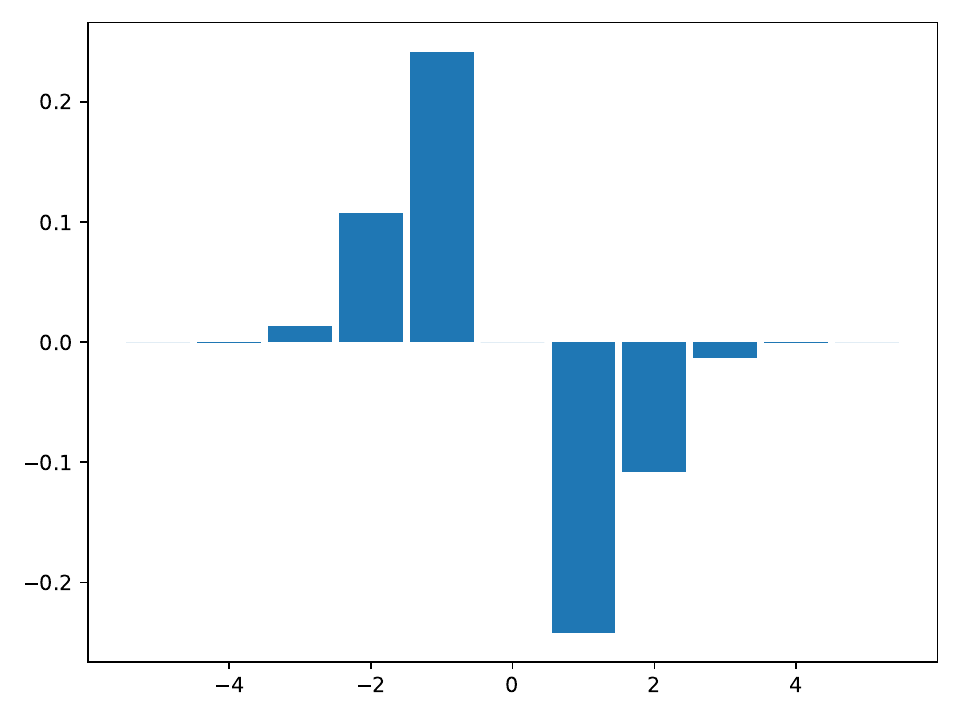}
      & \includegraphics[width=0.22\textwidth]{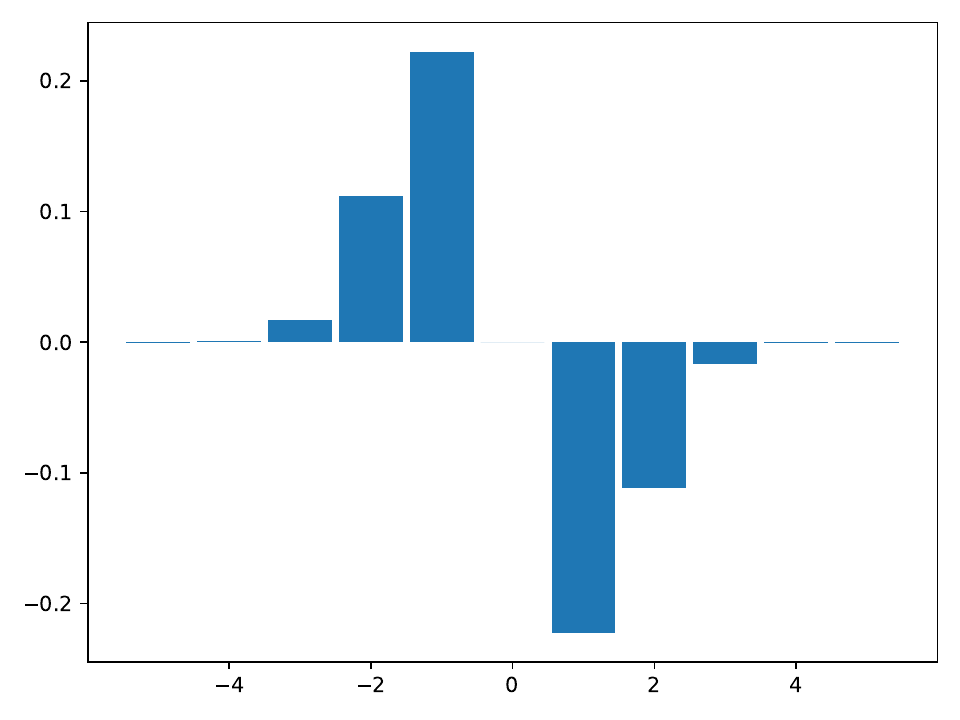}
      & \includegraphics[width=0.22\textwidth]{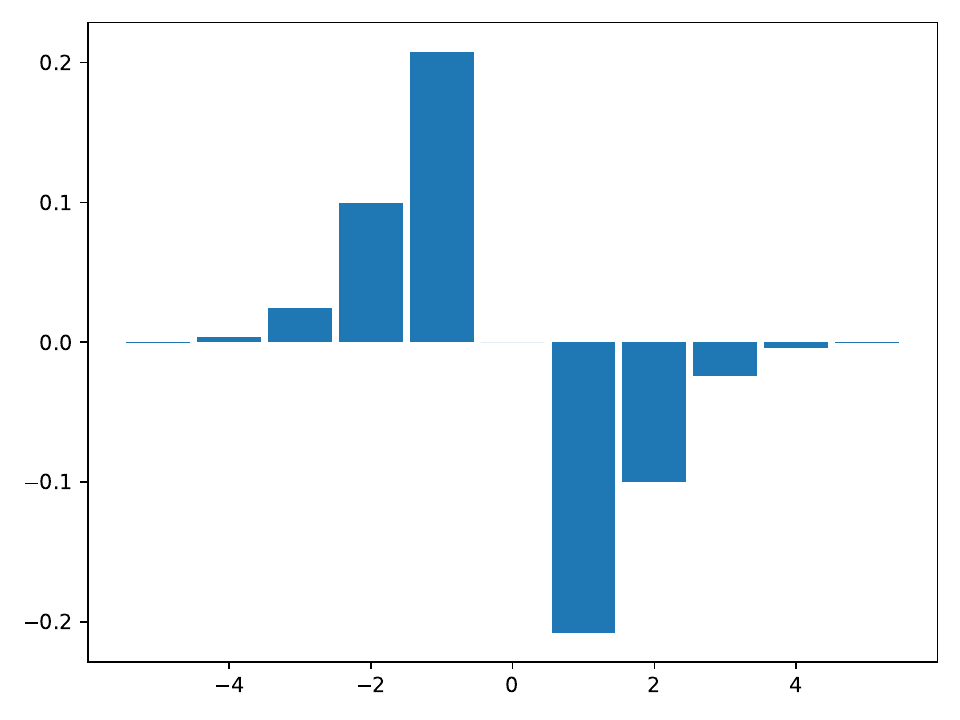} \\
      \includegraphics[width=0.22\textwidth]{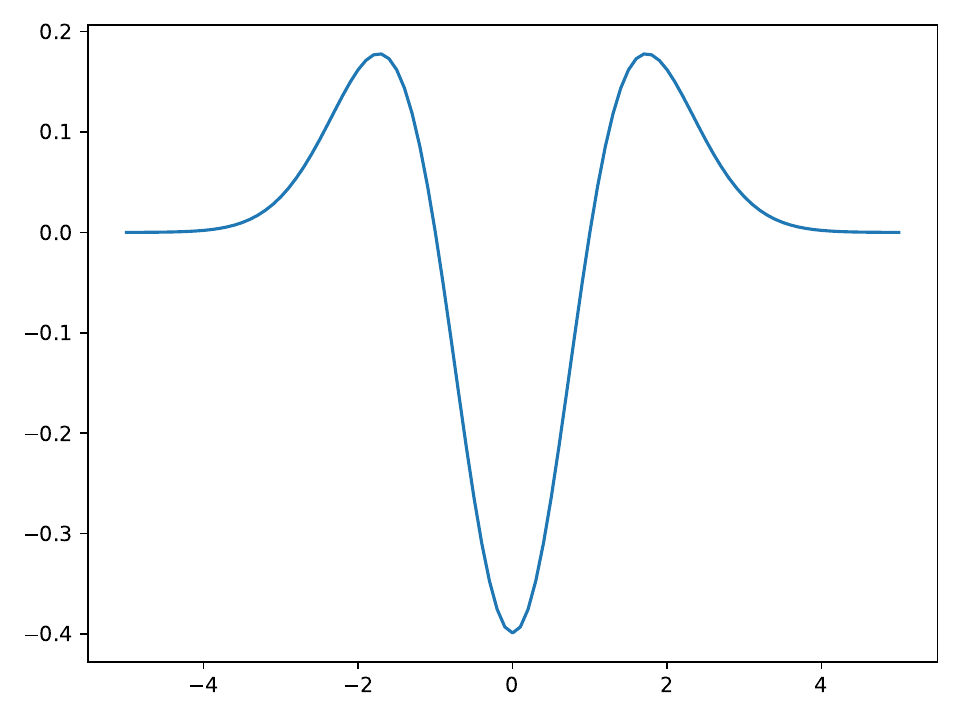}
      & \includegraphics[width=0.22\textwidth]{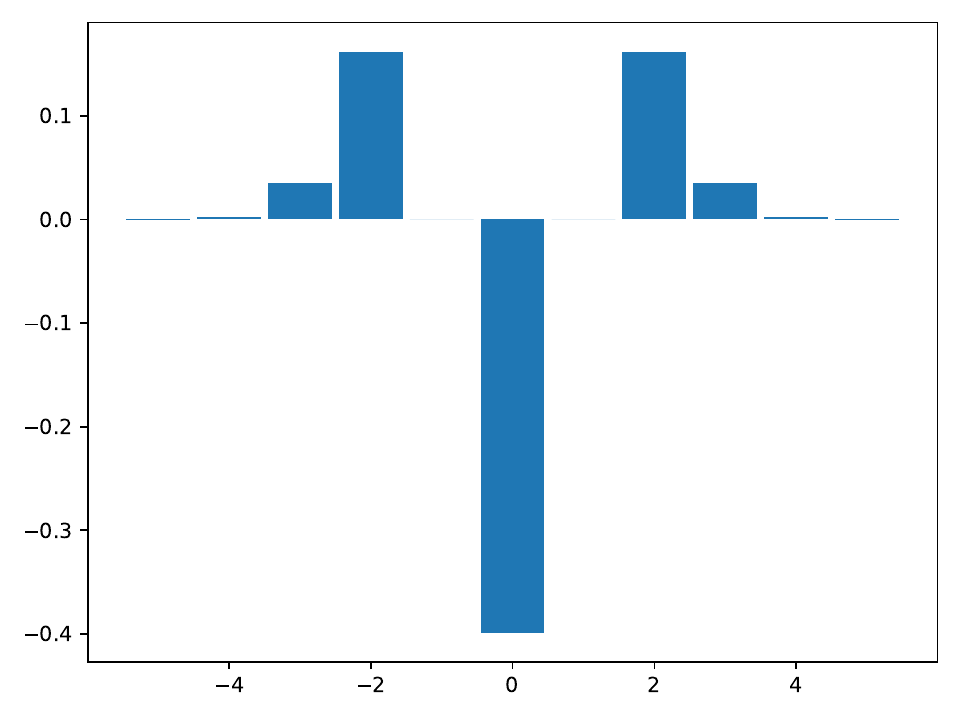}
      & \includegraphics[width=0.22\textwidth]{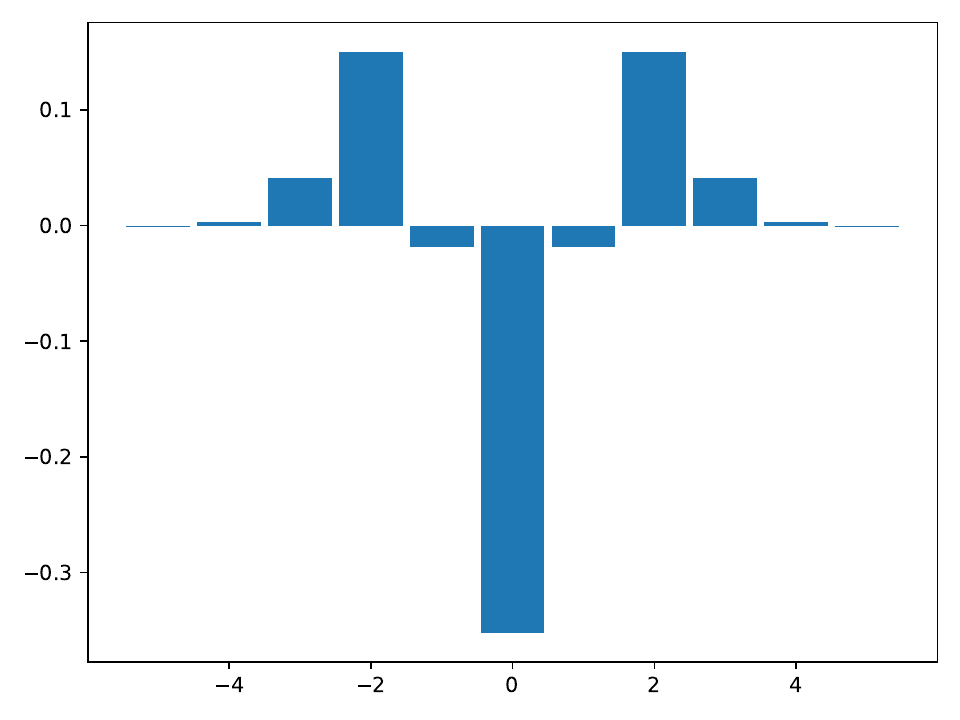}
      & \includegraphics[width=0.22\textwidth]{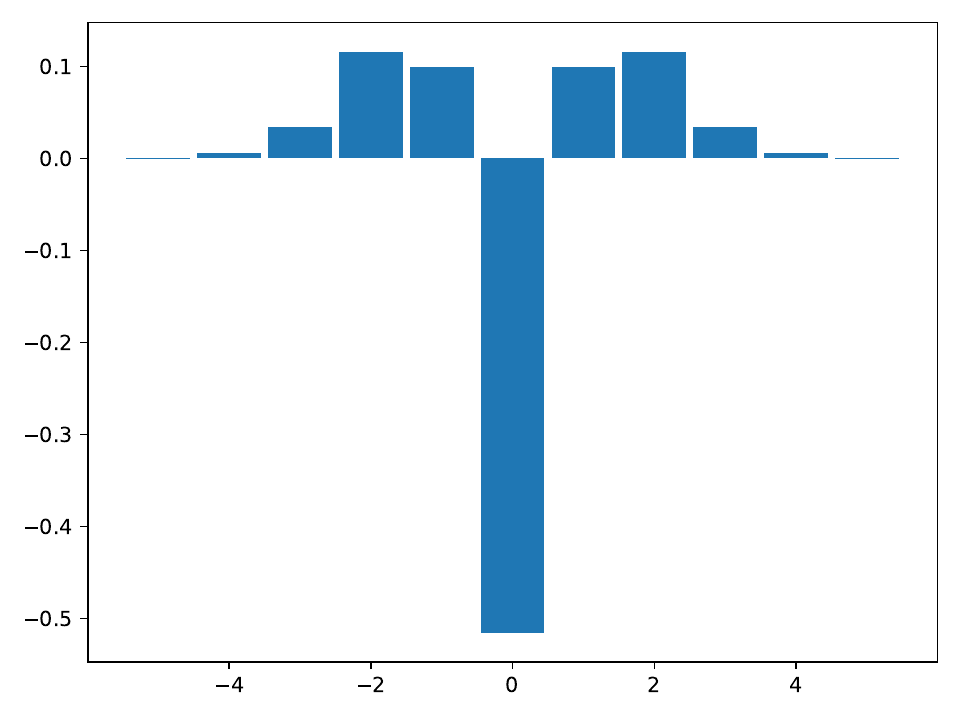} \\
      \includegraphics[width=0.22\textwidth]{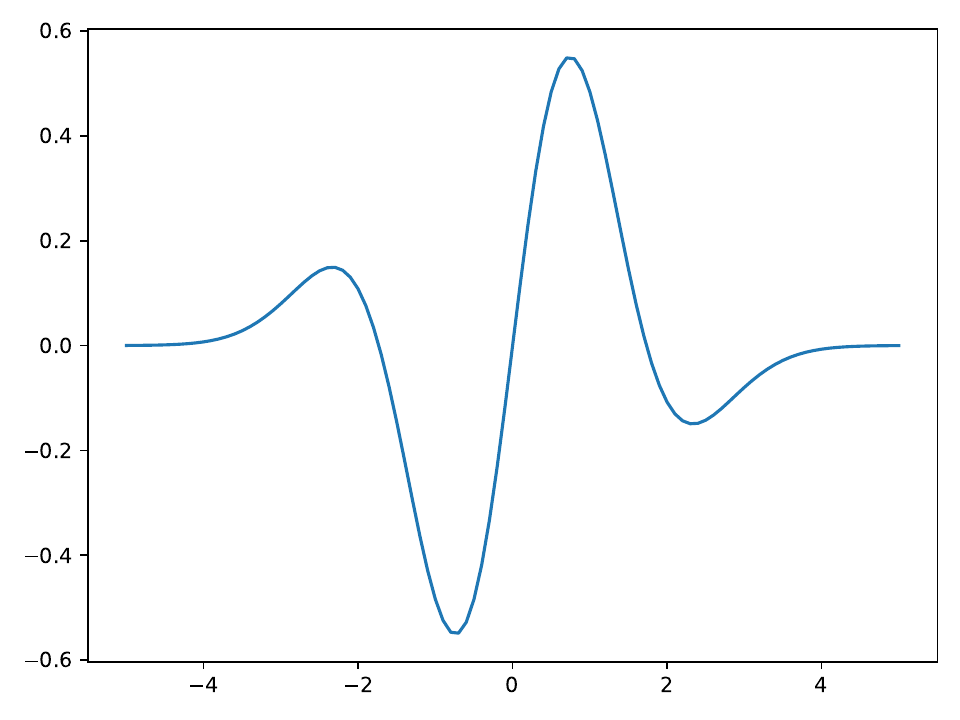}
      & \includegraphics[width=0.22\textwidth]{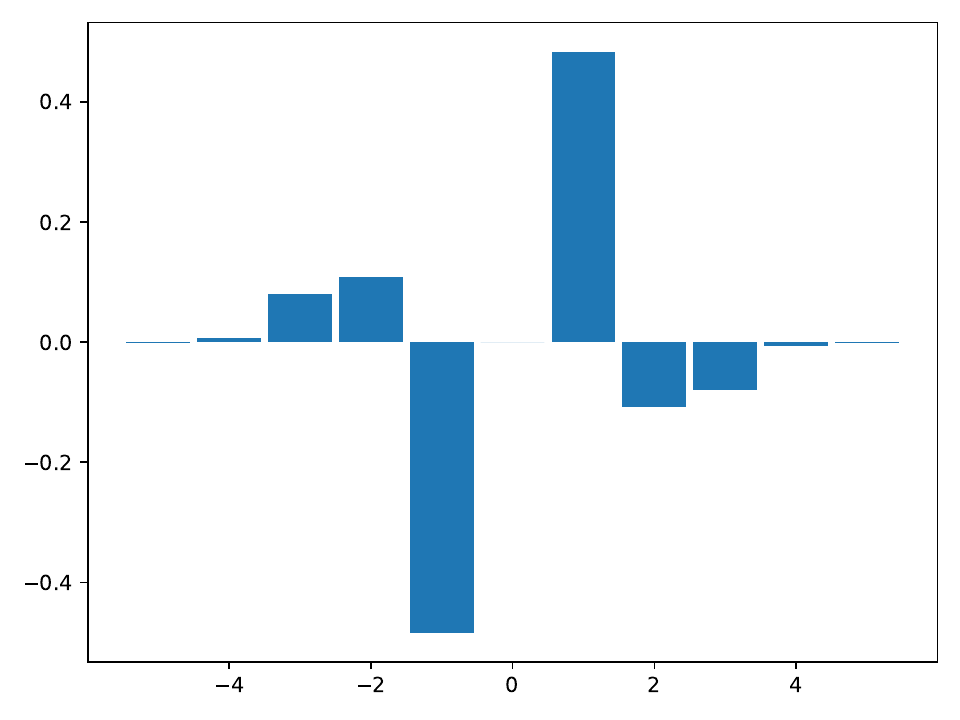}
      & \includegraphics[width=0.22\textwidth]{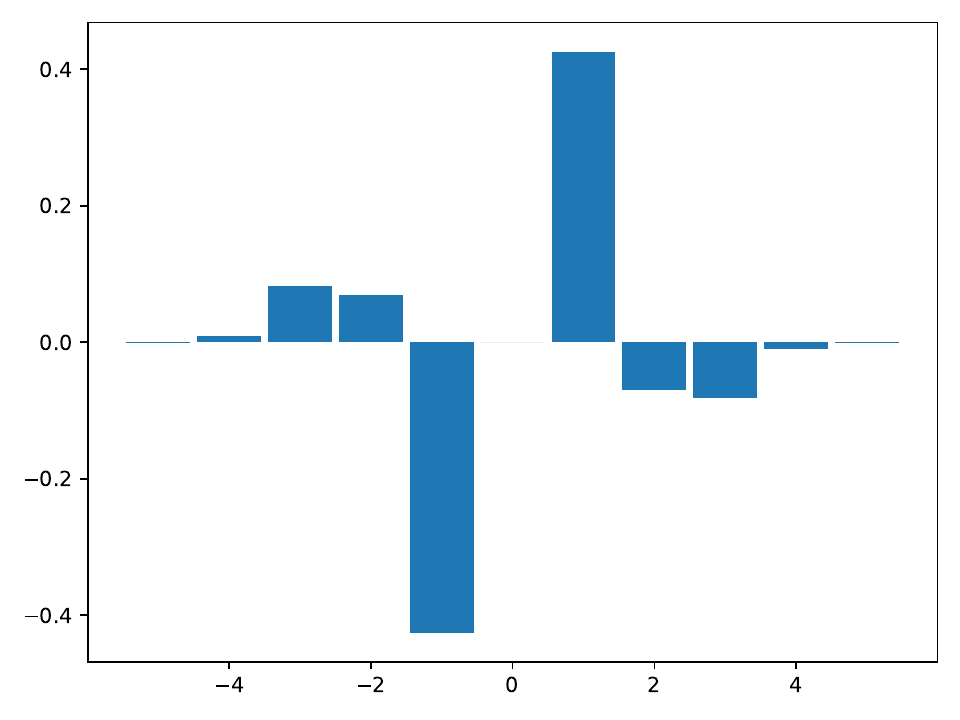}
      & \includegraphics[width=0.22\textwidth]{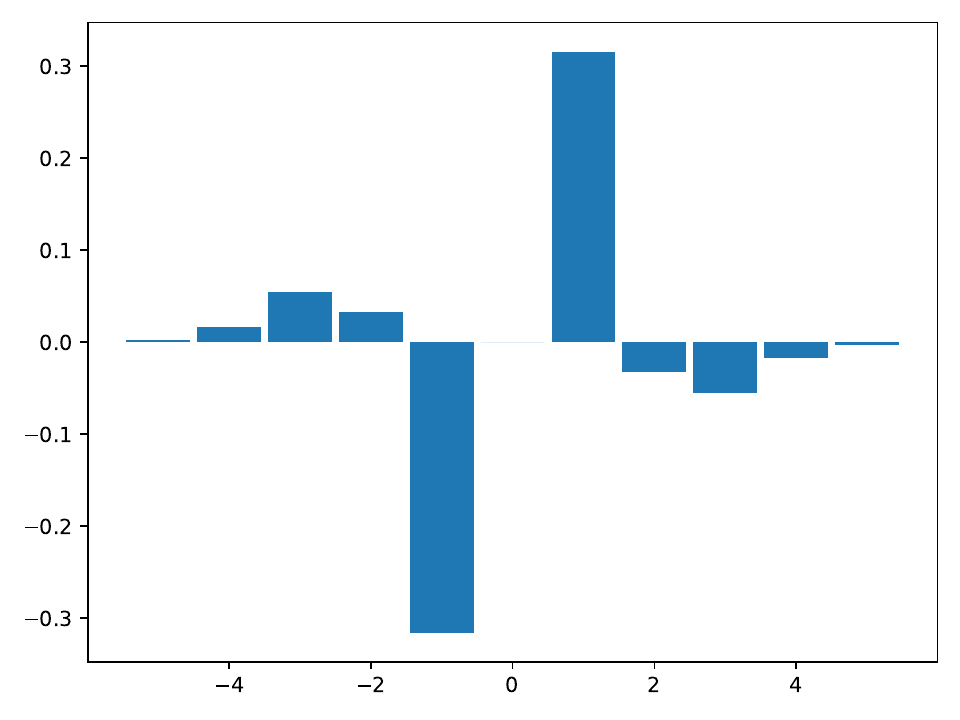} \\
      \includegraphics[width=0.22\textwidth]{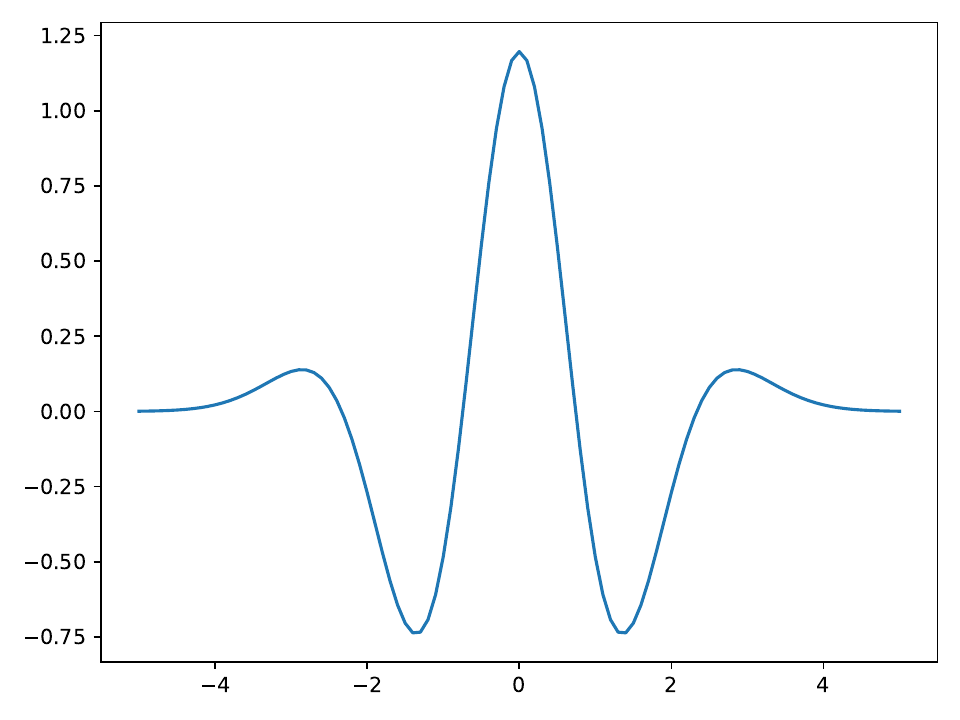}
      & \includegraphics[width=0.22\textwidth]{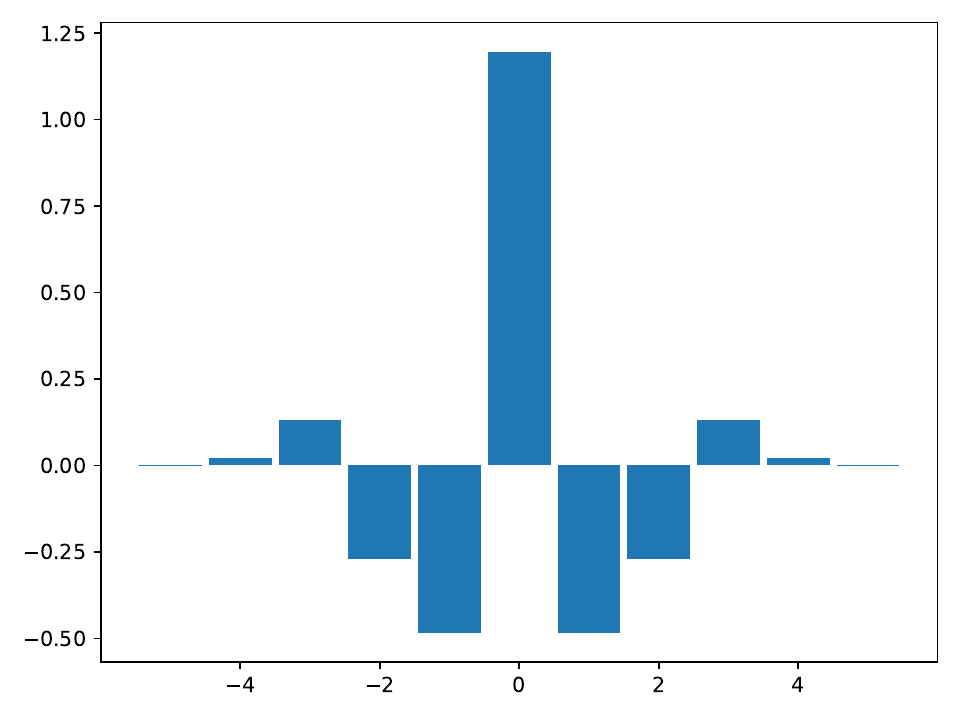}
      & \includegraphics[width=0.22\textwidth]{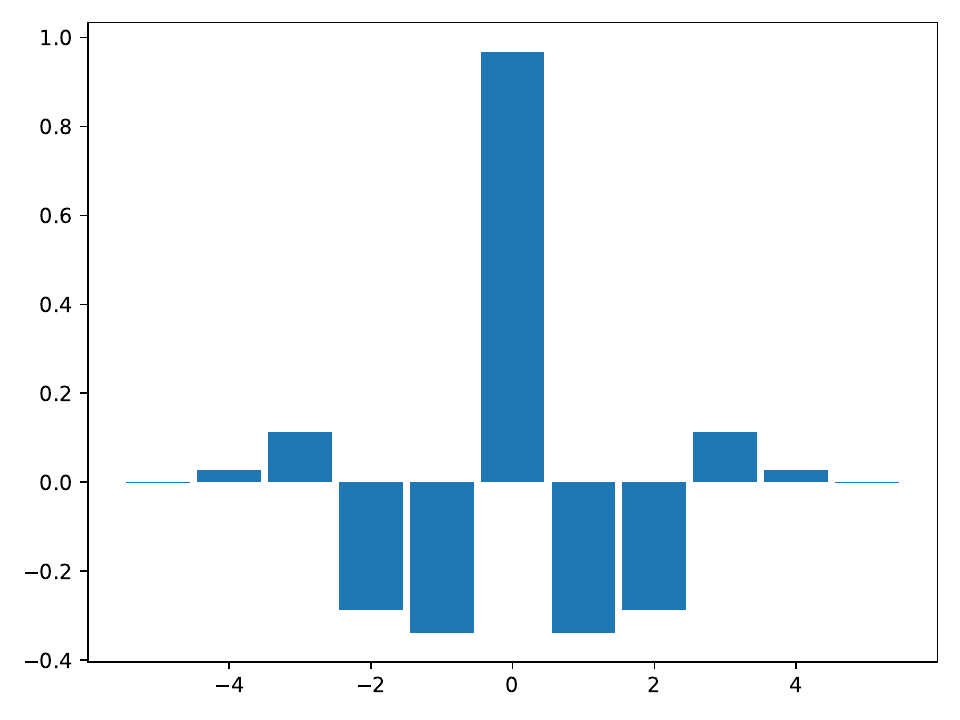}
      & \includegraphics[width=0.22\textwidth]{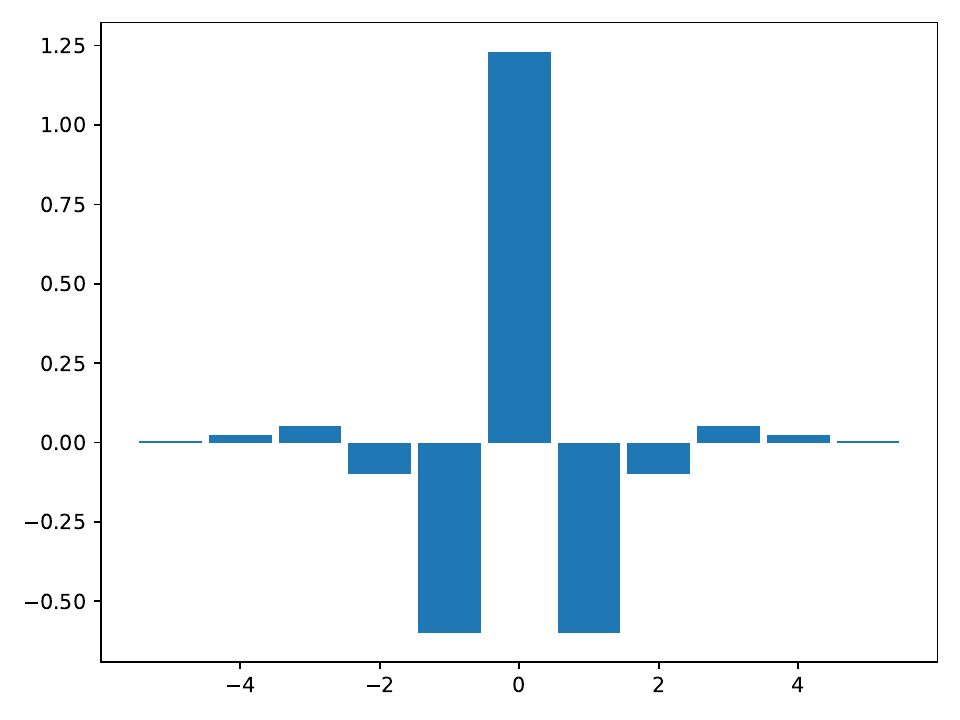} \\
    \end{tabular}
  \end{center}
  \caption{Graphs of the main types of Gaussian smoothing kernels and
    Gaussian derivative kernels considered in this paper, here at the
    scale $\sigma = 1$, with the raw
    smoothing kernels in the top row and the order of spatial differentiation
    increasing downwards up to order 4:
    (left column) continuous Gaussian kernels and continuous Gaussian derivatives,
    (middle left column) sampled Gaussian kernels and sampled Gaussian derivatives,
    (middle right column) integrated Gaussian kernels and integrated Gaussian derivatives, and
    (right column) discrete Gaussian kernels and discrete analogues of
    Gaussian derivatives.
    Note that the scaling of the vertical axis may vary between the
    different subfigures.
   (Horizontal axis: the 1-D spatial coordinate $x \in [-5, 5]$.)}
  \label{fig-kernel-graphs}
\end{figure*}

\subsection{The sampled Gaussian kernel}
\label{sec-sampl-gauss}

The presumably simplest approach for discretizing the 1-D Gaussian
convolution integral (\ref{eq-1D-cont-gauss-conv}) in terms of a
discrete convolution of the form (\ref{eq-1D-disc-gauss-conv}), is by
choosing the discrete kernel $T(n;\; s)$ as the sampled Gaussian kernel
\begin{equation}
  \label{eq-sampl-gauss}
  T_{\sampl}(n;\; s) = g(n;\; s).
\end{equation}
While this choice is easy to implement in practice, there are, however, three
major conceptual problems with using such a discretization at very
fine scales:
\begin{itemize}
\item
  the filter coefficients may not be limited to the interval $[0, 1]$,
\item
  the sum of the filter coefficients may become substantially greater than
  1, and
\item
  the resulting filter kernel may have too narrow shape, in the sense
  that the spatial variance of the discrete kernel
  $V(T_{\sampl}(\cdot;\; s))$ is substantially smaller than the spatial variance
  $V(g(\cdot;\; s))$ of the continuous Gaussian kernel.
\end{itemize}
The first two problems imply that the resulting discrete spatial
smoothing kernel is no longer a spatial weighted averaging kernel in
the sense of Section~\ref{sec-spat-avg}, which implies problems, if
attempting to interpret the result of convolutions with the sampled
Gaussian kernels as reflecting different spatial scales.
The third problem implies that there will not be a direct match
between the value of the scale parameter provided as argument to the
sampled Gaussian kernel and the scales that the discrete kernel would
reflect in the image data.

Figures~\ref{fig-gaussL1normerror} and~\ref{fig-gaussstddev}
show numerical characterizations of
these entities for a range of small values of the scale parameter.

More fundamentally, it can be shown
(see Section~VII.A in Lindeberg \citeyear{Lin90-PAMI})
that convolution with the sampled Gaussian kernel is guaranteed to
not increase the number of local
extrema (or zero-crossings) in the signal from the input signal to any
coarser level of scale. The transformation from an arbitrary scale
level to some other arbitrary coarser scale level is, however, not
guaranteed to obey such a simplification property between any pair of
scale levels.
In this sense, convolutions with sampled Gaussian kernels do not
truly obey non-creation of local extrema from finer to coarser levels
of scale, in the sense described in Section~\ref{sec-non-creat-struct}.

\subsection{The normalized sampled Gaussian kernel}
\label{sec-norm-sampl-gauss}

A straightforward, but ad hoc, way of avoiding the problems that the
discrete filter coefficients may, for small values 
of the scale parameter have their sum exceed 1, is by
normalizing the sampled Gaussian kernel with its discrete $l_1$-norm:
\begin{equation}
  \label{eq-def-norm-sampl-gauss}
  T_{\normsampl}(n;\; s) = \frac{g(n;\; s)}{\sum_{m \in \bbbz} g(m;\; s)}.
\end{equation}
By definition, we in this way avoid this problems that the regular
sampled Gaussian kernel is not spatial weighted averaging kernel in
the sense of Section~\ref{sec-spat-avg}.

The problem that the spatial variance of the discrete kernel
$V(T_{\normsampl}(\cdot;\; s))$ is substantially smaller that the spatial variance
$V(g(\cdot;\; s))$ of the continuous Gaussian kernel, will, however, persist,
since the variance of a kernel is not affected by a uniform scaling of
its amplitude values. In this sense, the resulting discrete kernels
will not for small scale values accurately reflect the spatial scale corresponding to the
scale argument, as specified by the scale parameter $s$.

\subsection{The integrated Gaussian kernel}
\label{sec-int-gauss}

A possibly better way of enforcing the weights of the filter kernels
to sum up to 1, is by instead
letting the discrete kernel be determined by the integral of the
continuous Gaussian kernel over each pixel support region
(Lindeberg \citeyear{Lin93-Dis} Equation~(3.89))
\begin{equation}
  \label{eq-def-int-gauss-kern}
   T_{\intdisc}(n;\; s) = \int_{x = n - 1/2}^{n + 1/2} g(x;\; s) \, dx,
 \end{equation}
 which in terms of the scaled error function
 $\operatorname{erg}(x;\; s)$ can be expressed as
 \begin{equation}
   \label{eq-def-int-gauss-erg}
   T_{\intdisc}(n;\; s)
   = \operatorname{erg}(n + \tfrac{1}{2};\; s) - \operatorname{erg}(n - \tfrac{1}{2};\; s)
 \end{equation}
 with
 \begin{equation}
   \operatorname{erg}(x;\; s)
   = \frac{1}{2}
       \left( 1 +
        \operatorname{erf} \left( \frac{x}{\sqrt{2 s}} \right)
      \right),
\end{equation}
where $\operatorname{erf}(x)$ denotes the regular error function
according to
\begin{equation}
  \operatorname{erf}(x)
  = \frac{2}{\sqrt{\pi}}
      \int_{t = 0}^x e^{-t^2} dt.
\end{equation}
A conceptual argument for defining the integrated Gaussian kernel
model is that, we may, given a discrete
signal $f(n)$, define a continuous signal $\tilde{f}(x)$, by letting
the values of the signal in each pixel support region be equal to the
value of the corresponding discrete signal,
see Appendix~\ref{app-deriv-int-gauss} for an explicit derivation.
In this sense, there is a
possible physical motivation for using this form of scale-space
discretization.

By the continuous Gaussian kernel having its integral equal to 1, it
follows that the sum of the discrete filter coefficients will over an
infinite spatial domain also be exactly equal to 1. Furthermore, the
discrete filter coefficients are also guaranteed to be in the
interval $[0, 1]$.
In these respects, the resulting discrete kernels
will represent a true spatial weighting process, in the sense of
Section~\ref{sec-spat-avg}.

Concerning the spatial variances $V(T_{\intdisc}(\cdot;\; s))$
of the resulting discrete kernels, they will also for smaller scale values be
closer to the spatial variances
$V(g(\cdot;\; s))$ of the continuous Gaussian kernel, than for the
sampled Gaussian kernel or the normalized sampled Gaussian kernel,
as shown in Figures~\ref{fig-gaussstddev} and~\ref{fig-gaussscalediff}.
For larger scale values, the box integration
over each pixel support region, will, on the other hand, however,
introduce a scale offset, which for larger values of the scale
parameter $s$ approaches
\begin{equation}
  \Delta s_{\intdisc} = \frac{1}{12} \approx 0.0833,
\end{equation}
which, in turn, corresponds to the spatial variance of a continuous box filter over
each pixel support region, defined by
\begin{equation}
  w_{\boxfilt}
  = \left\{
        \begin{array}{ll}
            1 & \mbox{if $|x| \leq \frac{1}{2}$,} \\
            0 & \mbox{otherwise,}
        \end{array}
      \right.
\end{equation}
and which is used for defining the integrated Gaussian kernel from the
continuous Gaussian kernel in (\ref{eq-def-int-gauss-kern}).

Figure~\ref{fig-gaussstddev} shows a numerical characterization of the difference in
scale values between the variance $V(T_{\intdisc}(n;\; s))$ of the discrete
integrated Gaussian kernel and the scale parameter $s$ provided as
argument to this function.

In terms of theoretical scale-space properties, it can be shown that
the transformation from the input signal to any coarse scale always
implies a simplification, in the sense that the number of local extrema
(or zero-crossings) at any coarser level of scale is guaranteed to not
exceed the number of local extrema (or zero-crossings) in the input
signal (see Section~3.6.3 in Lindeberg \citeyear{Lin93-Dis}).
The transformation from any finer scale level to any coarser
scale level will, however, not be guaranteed to obey such a
simplification property. In this respect, the integrated Gaussian
kernel does not fully represent a discrete scale-space transformation,
in the sense of Section~\ref{sec-non-creat-struct}.

\subsection{The discrete analogue of the Gaussian kernel}
\label{sec-disc-anal-gauss}

According to a genuinely discrete theory for spatial scale-space
representation in Lindeberg (\citeyear{Lin90-PAMI}),
the discrete scale space is defined from discrete
kernels of the form
\begin{equation}
  \label{eq-disc-gauss}
   T_{\disc}(n;\; s) = e^{-s} I_n(s),
 \end{equation}
 where $I_n(s)$ denote the modified Bessel functions of integer order
 (see Abramowitz and Stegun \citeyear{AS64}), which are related to the
 regular Bessel functions $J_n(z)$ of the first kind according to
\begin{equation}
  I_n(x) = i^{-n} \, J_n(i \, x) = e^{-\frac{n \pi i}{2}} \,
  J_n(e^{\frac{i \pi}{2}} x),
\end{equation}
and which for integer values of $n$, as we will restrict ourselves to here, can be
expressed as
\begin{equation}
  I_n(x)
  = \frac{1}{\pi}
      \int_{\theta = 0}^{\pi} e^{x \cos \theta}
         \cos(n \, \theta) \, d\theta.
\end{equation}
The discrete analogue of the Gaussian kernel $T_{\disc}(n;\; s)$
does specifically have the practically useful properties that:
\begin{itemize}
\item
  the filter coefficients are guaranteed to be in the interval $[0, 1]$,
\item
  the filter coefficients sum up to 1
  (see Equation~(3.43) in Lindeberg \citeyear{Lin93-Dis})
  \begin{equation}
     \sum_{n \in \bbbz} T_{\disc}(n;\; s) = 1,
   \end{equation}
\item
  the spatial variance of the discrete kernel is exactly equal to
  the scale parameter
  (see Equation~(3.53) in Lindeberg \citeyear{Lin93-Dis})
  \begin{equation}
     V(T_{\disc}(\cdot;\; s)) = s.
   \end{equation}
\end{itemize}
These kernels do also exactly obey a semi-group property over spatial
scales (see Equation~(3.41) in Lindeberg \citeyear{Lin93-Dis})
\begin{equation}
  T_{\disc}(\cdot;\; s_1) * T_{\disc}(\cdot;\; s_2) = T_{\disc}(\cdot;\; s_1 + s_2),
\end{equation}
which implies that the resulting discrete scale-space representation
also obeys an exact cascade smoothing property
\begin{equation}
  L_{\disc}(\cdot;\; s_2) = T_{\disc}(\cdot;\; s_2 - s_1) * L_{\disc}(\cdot;\; s_1).
\end{equation}
More fundamentally, these discrete kernels do furthermore preserve
scale-space properties to the discrete domain, in the sense that:
\begin{itemize}
\item
  the number of local extrema (or zero-crossings) at a coarser scale
  is guaranteed to not exceed the number of local extrema (or
  zero-crossings) at any finer scale,
\item
  the resulting discrete scale-space representation is guaranteed to
  obey non-enhancement of local extrema, in the sense that the value at
  any local maximum is guaranteed to not increase with increasing
  scale, and that the value at any local minimum is guaranteed to not
  decrease with increasing scale.
\end{itemize}
In these respects, the discrete analogue of the Gaussian kernel obeys
all the desirable theoretical properties of a discrete scale-space
representation, corresponding to discrete analogues of the theoretical
properties of the Gaussian scale-space representation stated in
Section~\ref{sec-theor-props-gauss-scsp}.

Specifically, the theoretical properties of the discrete analogue of
the Gaussian kernel are better than the theoretical properties of
the sampled Gaussian kernel, the normalized sampled Gaussian kernel
or the integrated Gaussian kernel.

\subsubsection{Diffusion equation interpretation of the genuinely
  discrete scale-space representation concept}

In terms of diffusion equations, the discrete scale-space
representation generated by convolving a 1-D discrete signal $f$ by the
discrete analogue of the Gaussian kernel according to
(\ref{eq-disc-gauss})
\begin{equation}
  \label{eq-1D-disc-scsp}
   L(x;\; s) = \sum_{n \in \bbbz} T_{\disc}(n;\, s) \, f(x - n)
\end{equation}
satisfies the semi-discrete 1-D diffusion equation
(Lindeberg \citeyear{Lin93-Dis} Theorem~3.28)
\begin{equation}
  \partial_s L = \frac{1}{2} \, \delta_{xx} L
\end{equation}
with initial condition $L(x;\; 0) = f(x)$, where $\delta_{xx}$
denotes the second-order discrete difference operator
\begin{equation}
  \delta_{xx} = (+1, -2, +1).
\end{equation}
Over a 2-D discrete spatial domain, the discrete scale-space
representation of an image $f(x, y)$, generated by separable
convolution with the discrete analogue of the Gaussian kernel
\begin{equation}
  \label{eq-2D-disc-scsp}  
  L(x, y;\; s) =
  \sum_{m \in \bbbz} T(m;\; s) \sum_{n \in \bbbz} \, T(n;\; s) \, f(x-m, y-n),
\end{equation}
satisfies the semi-discrete 2-D diffusion equation
(Lindeberg \citeyear{Lin93-Dis} Proposition~4.14)
\begin{equation}
  \partial_s L = \frac{1}{2} \, \nabla_5^2 L
\end{equation}
with initial condition $L(x, y;\; 0) = f(x, y)$, where $\nabla_5^2$
denotes the following discrete approximation of the Laplacian operator
\begin{equation}
  \nabla_5^2
  = \left(
        \begin{array}{ccc}
             0 & +1 &    0 \\
          +1 & -4  & +1 \\
            0 & -1  &    0
         \end{array}
    \right).
\end{equation}
In this respect, the discrete scale-space representation generated by
convolution with the discrete analogue of the Gaussian kernel can be seen
as a purely spatial discretization of the continuous diffusion
equation (\ref{eq-cont-2d-diffusion-eq}),
which can serve as an equivalent way of defining the
continuous scale-space representation. 

\subsection{Performance measures for quantifying deviations from theoretical
  properties of discretizations of Gaussian kernels}
\label{sec-perf-meas-gauss-smooth}

To quantify the deviations between properties of the discrete kernels,
and desirable properties of the discrete kernels that are to transfer
the desirable properties of a continuous scale-space representation to
a corresponding discrete implementation, we will in this section
quantity such deviations in terms of the following error measures:
\begin{itemize}
\item
  {\bf Normalization error:}
  The difference between the $l_1$-norm of the discrete kernels and
  the desirable unit $l_1$-norm normalization will be measured by%
\footnote{When implementing this operation in practice, the infinite sum is
  replaced by a finite sum
  $E_{\norm}(T(\cdot;\; s)) = \sum_{n = - N}^{N} T(n;\; s) - 1$,
  with the truncation bound $N$ chosen such that
  $2 \int_{x = N}^{\infty} g(x;\; s) \, dx \leq \epsilon$ for a small
  $\epsilon$ chosen as $10^{-8}$, according to Footnote~\ref{footnote-trunction-error}.}
  \begin{equation}
    \label{eq-gauss-L1norm-error}
    E_{\norm}(T(\cdot;\; s)) = \sum_{n \in \bbbz} T(n;\; s) - 1.
  \end{equation}
\item
  {\bf Absolute scale difference:}
  The difference between the variance of the discrete kernel and the
  argument of the scale parameter will be measured by
  \begin{equation}
    \label{eq-scale-diff-gauss}
    E_{\Delta s}(T(\cdot;\; s)) = V(T(\cdot;\; s)) - s.
  \end{equation}
  This error measure is expressed in absolute units of the scale parameter.
  The reason, why we express this measure in units of the variance of
  the discretizations of the Gaussian kernel, is that variances are
  additive under convolutions of non-negative kernels.
\item
  {\bf Relative scale difference:}
  The relative scale difference, between the actual standard deviation
  of the discrete kernel and the argument of the scale parameter, will
  be measured by
  \begin{equation}
    \label{eq-rel-scale-error-gauss}    
    E_{\relscale}(T(\cdot;\; s)) = \sqrt{\frac{V(T(\cdot;\; s))}{s}} - 1.
  \end{equation}
  This error measure is expressed in relative units of the scale
  parameter.%
\footnote{The motivation for using both an absolute and a
  relative error measure in this context, for the purpose of quantifying
  the differences between the estimated and
  the specified scale values for the discrete approximations of the
  Gaussian kernels, is that the definition of the integrated
  Gaussian kernel is associated with a specific type of scale
  offset, that can be seen as genuinely additive at coarse levels
  of scale, whereas for the purpose of comparing the scale differences
  computed at different scales, it may be more appropriate to quantify
  how large the scale differences are in relation to the scales at
  which they are actually computed.}
  The reason, why we express this entity in units of the standard
  deviations of the discretizations of the Gaussian kernels, is that
  these standard deviations correspond to interpretations of the scale
  parameter in units of $[\mbox{length}]$, in a way that is thus
  proportional to the scale level.

\item
  {\bf Cascade smoothing error:}
  The deviation from the cascade smoothing property of a scale-space
  kernel according to (\ref{eq-casc-prop-raw-scsp}) and the actual
  result of convolving a discrete approximation of the scale-space representation at a
  given scale $s$, with its corresponding discretization of the
  Gaussian kernel, will be measured by
  \begin{equation}
    \label{eq-casc-error-gauss}
    E_{\cascade}(T(\cdot;\; s))
    = \frac{\| T(\cdot;\; s) * T(\cdot;\; s)  - T(\cdot;\; 2s)  \|_1}
               {\| T(\cdot;\; 2s) \|_1}.
  \end{equation}
  While this measure of cascade smoothing error could in principle
  instead be formulated for arbitrary relations between the scale
  level of the discrete approximation of the scale-space
  representation and the amount of additive spatial smoothing, we fix
  these scale levels to be equal for the purpose of conceptual
  simplicity.%
  \footnote{It could, indeed, be interesting, to also investigate the
    cascade smoothing error for incremental smoothing for smaller
    values of $\Delta s$.}
\end{itemize}
In the ideal theoretical case, all of these error measures should be
equal to zero (up to numerical errors in the discrete computations).
Any deviations from zero of these error measures do therefore represent a
quantification of deviations from desirable theoretical properties in
a discrete approximation of the Gaussian smoothing operation.

\begin{figure}[hbtp]
  \begin{center}
    \begin{tabular}{c}
    {\em $l_1$-norm-based normalization error of the discrete kernels\/} \\
    \includegraphics[width=0.45\textwidth]{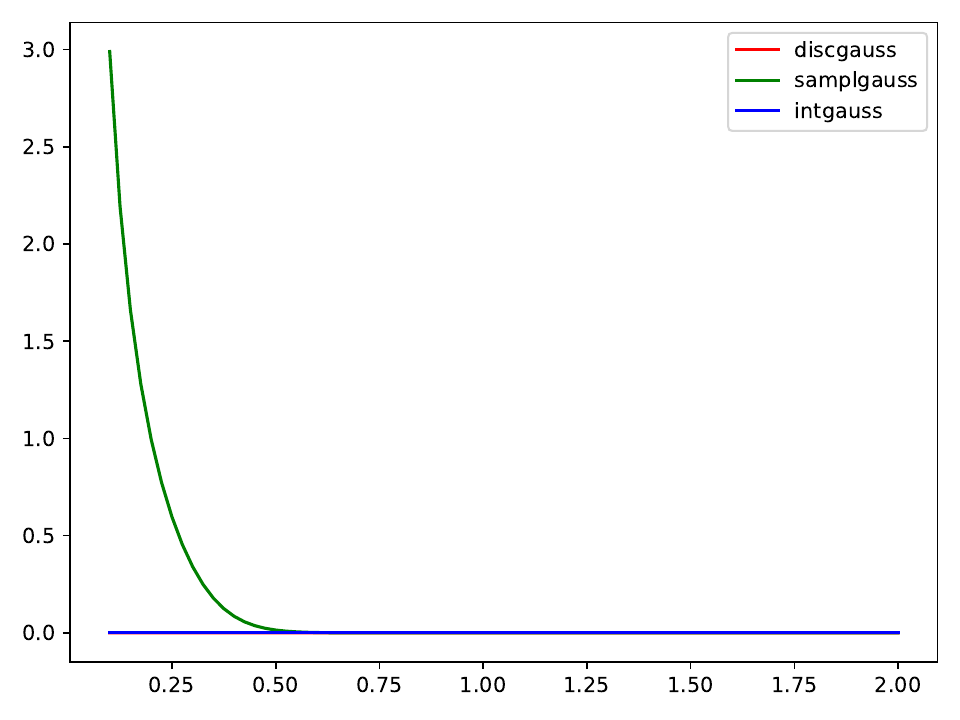}
    \end{tabular}
  \end{center}
  \caption{Graphs of the $l_1$-norm-based normalization error
    $E_{\norm}(T(\cdot;\; s))$, according to (\ref{eq-gauss-L1norm-error}),
    for the discrete analogue of the Gaussian kernel, the sampled
    Gaussian kernel and the integrated Gaussian kernel. Note that this
    error measure is equal to zero for both the discrete analogue of
    the Gaussian kernel, the normalized sampled Gaussian kernel
    and the integrated Gaussian kernel.
    (Horizontal axis: Scale parameter in units
    of $\sigma = \sqrt{s} \in [0.1, 2]$.)}
  \label{fig-gaussL1normerror}
\end{figure}

\begin{figure}[hbtp]
  \begin{center}
    \begin{tabular}{c}
    {\em Spatial standard deviations of the discrete kernels\/} \\
    \includegraphics[width=0.45\textwidth]{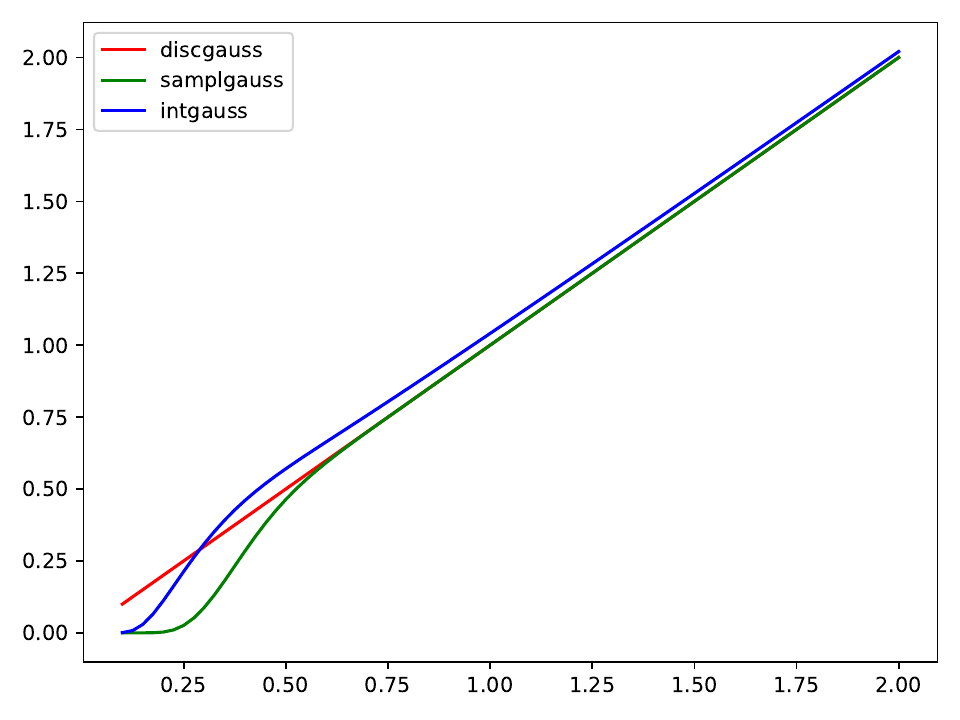}
    \end{tabular}
  \end{center}
  \caption{Graphs of the spatial standard deviations $\sqrt{V(T(\cdot;\; s))}$
    for the discrete analogue of the Gaussian kernel, the sampled
    Gaussian kernel and the integrated Gaussian kernel. The standard
    deviation is exactly equal to the scale parameter $\sigma = \sqrt{s}$
    for the discrete analogue of the Gaussian kernel. The standard
    deviation of the normalized sampled Gaussian kernel is equal to the
    standard deviation of the regular sampled Gaussian kernel.
    (Horizontal axis: Scale parameter in units of
    $\sigma = \sqrt{s} \in [0.1, 2]$.)}
  \label{fig-gaussstddev}
\end{figure}

\begin{figure}[hbpt]
  \begin{center}
    \begin{tabular}{c}
    {\em Spatial variance offset of the discrete kernels\/} \\
    \includegraphics[width=0.45\textwidth]{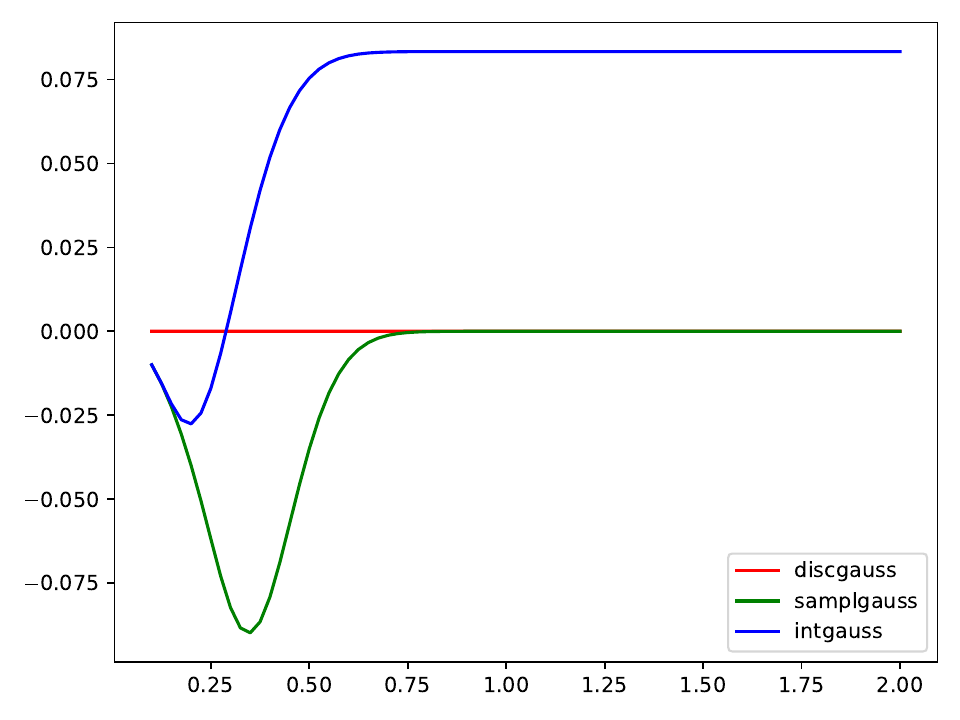}
    \end{tabular}
  \end{center}
  \caption{Graphs of the absolute scale difference $E_{\Delta s}(T(\cdot;\; s))$,
    according to (\ref{eq-scale-diff-gauss}) and
    in units of the spatial variance $V(T(\cdot;\; s))$,
    for the discrete analogue of the Gaussian kernel, the sampled
    Gaussian kernel and the integrated Gaussian kernel. This scale
    difference is exactly equal to zero for the discrete analogue of
    the Gaussian kernel. For scale values $\sigma < 0.75$, the
    absolute scale difference is substantial for the sampled Gaussian kernel, and
    then rapidly tends to zero for larger scales. For the integrated
    Gaussian kernel, the absolute scale difference does, however, not approach
    zero with increasing scale. Instead, it approaches the numerical
    value $\Delta s \approx 0.0833$, close to the spatial variance
    $1/12$ of a box filter over each pixel support region.
    The spatial variance-based absolute scale difference for the normalized sampled Gaussian
    kernel is equal to the spatial variance-based absolute scale difference
    for the regular sampled Gaussian kernel.
    (Horizontal axis: Scale parameter in units of
    $\sigma = \sqrt{s} \in [0.1, 2]$.)}
  \label{fig-gaussscalediff}
\end{figure}

\begin{figure}[hbpt]
  \begin{center}
    \begin{tabular}{c}
    {\em Relative scale difference of the discrete kernels\/} \\
    \includegraphics[width=0.45\textwidth]{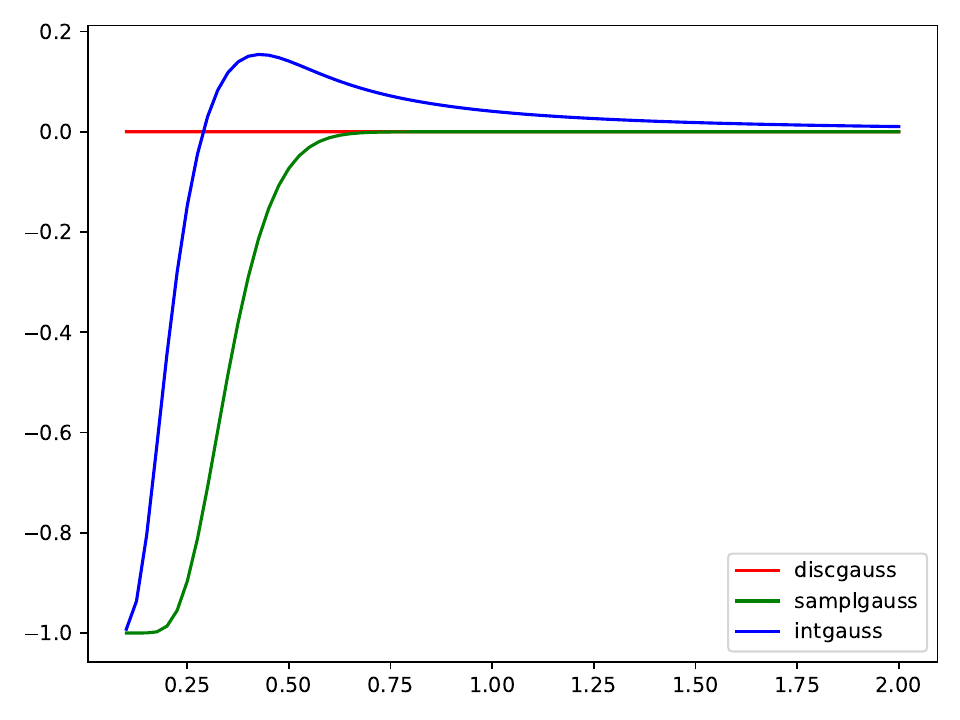}
    \end{tabular}
  \end{center}
  \caption{Graphs of the relative scale difference $E_{\relscale}(T(\cdot;\; s))$,
    according to (\ref{eq-rel-scale-error-gauss}) and
    in units of the spatial standard deviation of the discrete kernels,
    for the discrete analogue of the Gaussian kernel, the sampled
    Gaussian kernel and the integrated Gaussian kernel. This relative
    scale error is exactly equal to zero for the discrete analogue of
    the Gaussian kernel. For scale values $\sigma < 0.75$, the
    relative scale difference is substantial for sampled Gaussian kernel, and
    then rapidly tends to zero for larger scales. For the integrated
    Gaussian kernel, the relative scale difference is significantly larger,
    while approaching zero with increasing scale.
    The relative scale difference for the normalized sampled Gaussian
    kernel is equal to the relative scale difference for the regular
    sampled Gaussian kernel.
    (Horizontal axis: Scale parameter in units of
    $\sigma = \sqrt{s} \in [0.1, 2]$.)}
  \label{fig-gaussrelscaleerror}
\end{figure}

\begin{figure}[hbpt]
  \begin{center}
    \begin{tabular}{c}
    {\em Cascade smoothing error\/} \\
    \includegraphics[width=0.45\textwidth]{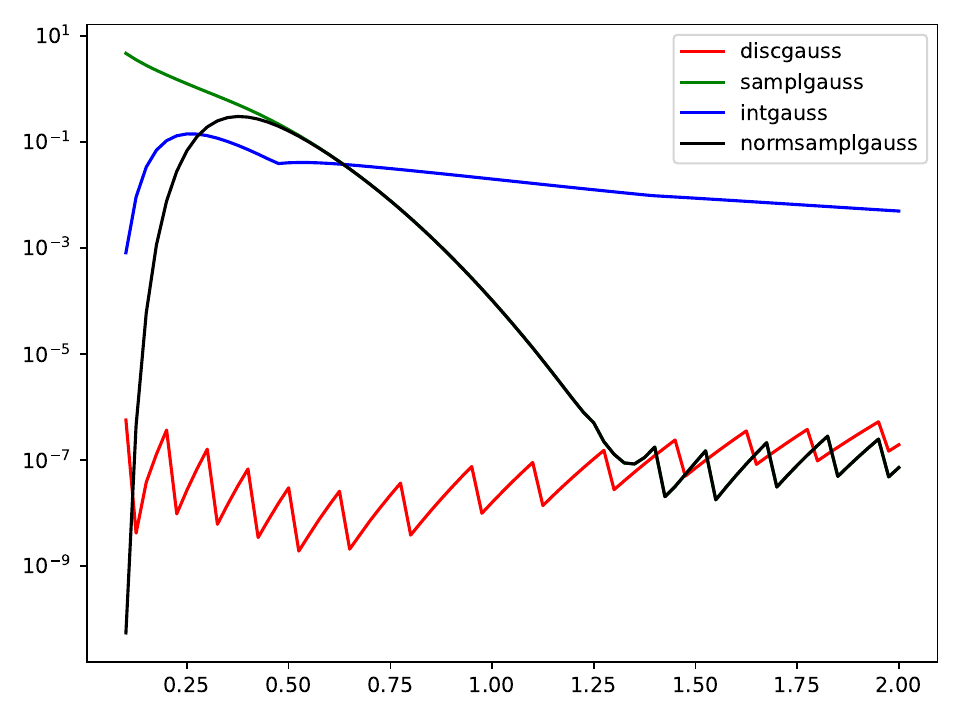}
    \end{tabular}
  \end{center}
  \caption{Graphs of the cascade smoothing error $E_{\cascade}(T(\cdot;\; s))$,
    according to (\ref{eq-casc-error-gauss}),
    for the discrete analogue of the Gaussian kernel, the sampled
    Gaussian kernel, the integrated Gaussian kernel, as well as the
    normalized sampled Gaussian kernel.
    For exact numerical computations, this cascade smoothing error would be
    identically equal to zero for the discrete analogue of
    the Gaussian kernel. In the numerical implementation underlying
    these computations, there are, however, numerical errors of a low
    amplitude. For the sampled Gaussian kernel, the cascade smoothing
    error is very large for $\sigma \leq 0.5$, notable for $\sigma < 0.75$,
    and then rapidly
    decreases with increasing scale. For the normalized sampled
    Gaussian kernel, the cascade smoothing error is for
    $\sigma \leq 0.5$ significantly lower than for the regular sampled
    Gaussian kernel. For the integrated Gaussian
    kernel, the cascade smoothing error is lower than for the sampled
    Gaussian kernel for  $\sigma \leq 0.5$, while then decreasing
    substantially slower to zero than for the sampled Gaussian kernel.
    (Horizontal axis: Scale parameter in units of
    $\sigma = \sqrt{s} \in [0.1, 2]$.)}
  \label{fig-gausscascerror}
\end{figure}

\subsection{Numerical quantifications of performance measures}

In the following, we will show results of computing the above measures
concerning desirable properties of discretizations of scale-space
kernels for the cases of
(i)~the sampled Gaussian kernel, (ii)~the integrated Gaussian kernel
and (iii)~the discrete analogue of the Gaussian kernel.
Since the discretization effects are largest for small scale values,
we will focus on the scale interval $\sigma \in [0.1, 2.0]$, however, in a few
cases extended to the scale interval $\sigma \in [0.1, 4.0]$.
(The reason for delimiting the scale parameter to the lower bound of
$\sigma \geq 0.1$ is to avoid the singularity at $\sigma = 0$.)

\subsubsection{Normalization error}

Figure~\ref{fig-gaussL1normerror} shows graphs of the $l_1$-norm-based
normalization error $E_{\norm}(T(\cdot;\; s))$ according to
(\ref{eq-gauss-L1norm-error}) for the
main classes of discretizations of Gaussian kernels. For the
integrated Gaussian kernel, the discrete analogue of the Gaussian
kernel and the normalized sampled Gaussian kernel, the normalization
error is identically equal to zero. For $\sigma \leq 0.5$, the
normalization error is, however, substantial for the regular sampled Gaussian kernel.

\subsubsection{Standard deviations of the discrete kernels}

Figure~\ref{fig-gaussstddev} shows graphs of the standard deviations
$\sqrt{V(T(\cdot;\; s))}$ for the different main types of
discretizations of the Gaussian kernels, which constitutes a natural
measure of their spatial extent. For the discrete analogue of
the Gaussian kernel, the standard deviation of the discrete kernel is exactly
equal to the value of the scale parameter in units of $\sigma =
\sqrt{s}$.
For the sampled Gaussian kernel, the standard deviation is
substantially lower than the value of the scale parameter in units of
$\sigma = \sqrt{s}$ for $\sigma \leq 0.5$. For the integrated Gaussian
kernel, the standard deviation is for smaller values of the scale
parameter closer to a the desirable linear trend. For larger values of
the scale parameter, the standard deviation of the discrete kernel is,
however, notably higher than $\sigma$.

\subsubsection{Spatial variance offset of the discrete kernels}

To quantify in a more detailed manner how the scale offset of the
discrete approximations of Gaussian kernels depends upon the scale
parameter, Figure~\ref{fig-gaussscalediff} shows graphs of the spatial variance-based
scale difference measure $E_{\Delta s}(T(\cdot;\; s))$
according to (\ref{eq-scale-diff-gauss}) for the different
discretization methods. For the discrete analogue of the Gaussian
kernel, the scale difference is exactly equal to zero. For the sampled
Gaussian kernel, the scale difference measure differs significantly
from zero for $\sigma < 0.75$, while then rapidly approaching zero
for larger scales. For the integrated Gaussian kernel, the
variance-based scale difference measure does, however, not approach
zero for larger scales. Instead, it approaches the numerical
value $\Delta s \approx 0.0833$, close to the spatial variance
$1/12$ of a box filter over each pixel support region.
The spatial variance-based scale difference for the normalized sampled Gaussian
kernel is equal to the spatial variance-based scale difference
for the regular sampled Gaussian kernel.

\subsubsection{Spatial standard-deviation-based relative scale difference}

Figure~\ref{fig-gaussrelscaleerror}  shows the spatial
standard-deviation-based relative scale difference $E_{\relscale}(T(\cdot;\; s))$
according to (\ref{eq-rel-scale-error-gauss}) for the main classes of
discretizations of Gaussian kernels.
This relative scale difference is exactly equal to zero for the discrete analogue of
the Gaussian kernel. For scale values $\sigma < 0.75$, the
relative scale difference is substantial for sampled Gaussian kernel, and
then rapidly tends to zero for larger scales. For the integrated
Gaussian kernel, the relative scale difference is significantly larger,
while approaching zero with increasing scale.
The relative scale difference for the normalized sampled Gaussian
kernel is equal to the relative scale difference for the regular
sampled Gaussian kernel.

\subsubsection{Cascade smoothing error}

Figure~\ref{fig-gausscascerror} shows the
cascade smoothing error $E_{\cascade}(T(\cdot;\; s))$
according to (\ref{eq-casc-error-gauss}) for the main classes of
discretizations of Gaussian kernels, while here complemented also with
results for the normalized sampled Gaussian kernel, since the results
for the latter kernel are different than for the regular sampled
Gaussian kernel.

For exact numerical computations, this cascade smoothing error should be
identically equal to zero for the discrete analogue of
the Gaussian kernel. In the numerical implementation underlying
these computations, there are, however, numerical errors of a low
amplitude. For the sampled Gaussian kernel, the cascade smoothing
error is very large for $\sigma \leq 0.5$, notable for $\sigma < 0.75$,
and then rapidly
decreases with increasing scale. For the normalized sampled
Gaussian kernel, the cascade smoothing error is for
$\sigma \leq 0.5$ significantly lower than for the regular sampled
Gaussian kernel. For the integrated Gaussian
kernel, the cascade smoothing error is lower than for the sampled
Gaussian kernel for  $\sigma \leq 0.5$, while then decreasing much
slower than for the sampled Gaussian kernel.

\subsection{Summary of the characterization results from the theoretical
  analysis and the quantitative performance measures}

To summarize the theoretical and the experimental results presented in
this section, the discrete analogue of the
Gaussian kernel stands out as having the best theoretical properties
in the stated respects, out of the set of treated discretization
methods for the Gaussian smoothing operation.

The choice, concerning which method is preferable out of the choice
between either the sampled Gaussian kernel or the integrated kernel,
depends on whether one would prioritize the behaviour at either very fine
scales or at coarse scales. The integrated Gaussian kernel has
significantly better approximation of theoretical properties at fine
scales, whereas its variance-based scale offset at coarser scales
implies significantly larger deviations from the desirable theoretical
properties at coarser scales, compared to either the sampled Gaussian
kernel or the normalized sampled Gaussian kernel. The normalized
sampled Gaussian kernel has properties closer to the desirable
properties than the regular sampled Gaussian kernel. If one would
introduce complementary mechanisms to compensate for the scale offset
of the integrated Gaussian kernel, that kernel could, however, also
constitute a viable solution at coarser scales.

\section{Discrete approximations of Gaussian derivative operators}
\label{disc-approx-gauss-ders}

According to the theory by Koenderink and van Doorn
(\citeyear{KoeDoo87-BC,KoeDoo92-PAMI}), Gaussian derivatives constitute a canonical
family of operators to derive from a Gaussian scale-space
representation. Such Gaussian derivative operators can be equivalently
defined by, either differentiating the Gaussian scale-space
representation
\begin{equation}
  L_{x^{\alpha} y^{\beta}}(x, y;\; s)
  = \partial_{x^{\alpha}  y^{\beta}} L(x, y;\; s),
\end{equation}
or by convolving the input image by Gaussian derivative kernels
\begin{multline}
  L_{x^{\alpha} y^{\beta}}(x, y;\; s) = \\
  = \int_{\xi \in \bbbr} \int_{\eta \in \bbbr}
        g_{2D,x^{\alpha} y^{\beta}}(\xi, \eta;\; s) \, f(x - \xi, y - \eta) \, d\xi \, d\eta,
\end{multline}
where
\begin{equation}
  g_{2D,x^{\alpha} y^{\beta}}(x, y;\; s)
  = \partial_{x^{\alpha}  y^{\beta}} g_{2D}(x, y;\; s)
\end{equation}
and $\alpha$ and $\beta$ are non-negative integers.

\subsection{Theoretical properties of Gaussian derivatives}


Due to the cascade smoothing property of the Gaussian smoothing
operation, in combination with the commutative property of
differentiation under convolution operations, it follows that the
Gaussian derivative operators also satisfy a cascade smoothing
property over scales:
\begin{equation}
   \label{eq-casc-prop-gauss-der}
   L_{x^{\alpha} y^{\beta}}(\cdot, \cdot;\; s_2)
   = g(\cdot, \cdot;\; s_2 - s_1) *
      L_{x^{\alpha} y^{\beta}}(\cdot, \cdot;\; s_1).
    \end{equation}
Combined with the simplification property of the Gaussian kernel under
increasing values of the scale parameter, it follows that the Gaussian
derivative responses also obey such a simplifying property from finer
to coarser levels of scale, in terms of (i)~non-creation of new local
extrema from finer to coarser levels of scale for 1-D signals, or
(ii)~non-enhancement of local extrema for image data over any number
of spatial dimensions.

\subsection{Separable Gaussian derivative operators}

By the separability of the Gaussian derivative kernels
\begin{equation}
  g_{2D, x^{\alpha} y^{\beta}}(x, y;\; s) = g_{x^{\alpha}}(x;\; s) \, g_{y^{\beta}}(y;\; s),
\end{equation}
the 2-D Gaussian derivative response can also be written as a
separable convolution of the form
\begin{align}
   \begin{split}
      & L_{x^{\alpha} y^{\beta}}(x, y;\; s) = \\
   \end{split}\nonumber\\
   \begin{split}
      & = \int_{\xi \in \bbbr} g_{x^{\alpha}}(\xi;\, s) \times
   \end{split}\nonumber\\
   \begin{split}
       & \quad\quad\quad\quad
      \left( \int_{\eta \in \bbbr}
               g_{y^{\beta}}(\eta;\; s) \, f(x - \xi, y - \eta) \, d\eta
             \right) \, d\xi.
    \end{split}
\end{align}
In analogy with the previous treatment of purely Gaussian convolution
operations, we will henceforth, for simplicity, consider the case with
1-D Gaussian derivative convolutions of the form
\begin{equation}
  \label{eq-1D-cont-gauss-der-conv}
   L_{x^{\alpha}}(x;\; s) = \int_{\xi \in \bbbr} g_{x^{\alpha}}(\xi;\, s) \, f(x - \xi) \, d\xi,
 \end{equation}
which are to be implemented in terms of discrete convolutions of the form
\begin{equation}
  \label{eq-1D-disc-gauss-der-conv}
   L_{x^{\alpha}}(x;\; s) = \sum_{n \in \bbbz} \, T_{x^{\alpha}}(n;\, s) \, f(x - n)
\end{equation}
for some family of discrete filter kernels $T_{x^{\alpha}}(n;\; s)$.

\subsubsection{Measures of the spatial extent of Gaussian derivative
  or derivative approximation kernels}

The spatial extent (spread) of a Gaussian derivative operator
$g_{x^{\alpha}}(\xi;\, s)$ of the form
(\ref{eq-1D-cont-gauss-der-conv}) will be measured by the variance of
its absolute value
\begin{equation}
  S_{\alpha} = {\cal S}(g_{x^{\alpha}}(\cdot;\, s)) = \sqrt{V(|g_{x^{\alpha}}(\cdot;\, s)|)}.
\end{equation}
Explicit expressions for these spread measures computed for continuous
Gaussian derivative kernels up to order 4 are given in
Appendix~\ref{app-spread-measures}.

Correspondingly, the spatial extent of a discrete kernel
$T_{x^{\alpha}}(n;\; s)$ designed to approximate a Gaussian derivative
operator will be measured by the entity
\begin{equation}
  {\cal S}(T_{x^{\alpha}}(\cdot;\, s)) = \sqrt{V(|T_{x^{\alpha}}(\cdot;\, s)|)}.
\end{equation}

\subsection{Sampled Gaussian derivative kernels}

In analogy with the previous treatment for the sampled Gaussian
kernel in Section~\ref{sec-sampl-gauss}, the presumably simplest
way to discretize the Gaussian derivative convolution integral
(\ref{eq-1D-cont-gauss-der-conv}), is by letting the discrete
filter coefficients in the discrete convolution operation
(\ref{eq-1D-disc-gauss-der-conv}) be determined as sampled
Gaussian derivatives
\begin{equation}
  \label{eq-sampl-gauss-der}
  T_{\sampl,x^{\alpha}}(n;\, s) = g_{x^{\alpha}}(n;\, s).
\end{equation}
Appendix~\ref{sec-app-expl-gauss-ders} describes how the Gaussian
derivative kernels are related to the probabilistic Hermite
polynomials, and does also give explicit expressions for the 1-D
Gaussian derivative kernels up to order 4.

For small values of the scale parameter, the resulting discrete
kernels may, however, suffer from the following problems:
\begin{itemize}
\item
  the $l_1$-norms of the discrete kernels may deviate substantially
  from the $L_1$-norms of the corresponding continuous Gaussian
  derivative kernels (with explicit expressions for the $L_1$-norms
  of the continuous Gaussian derivative kernels up to order 4
  given in Appendix~\ref{app-L1-norms-gauss-der-kernels}),
\item
  the resulting filters may have too narrow shape, in the sense that
  the spatial variance of the absolute value of the discrete kernel
  $V(|T_{\sampl,x^{\alpha}}(\cdot;\, s)|)$ may differ substantially from the
  spatial variance of the absolute value of the corresponding
  continuous Gaussian derivative kernel $V(|g_{x^{\alpha}}(\cdot;\,
  s)|)$ (see Appendix~\ref{app-spread-measures} for explicit
  expressions for these spatial spread measures for the continuous Gaussian
  derivatives up to order 4).
\end{itemize}
Figures~\ref{fig-gauss-der1-L1-norms}--\ref{fig-gauss-der4-L1-norms}
and Figures~\ref{fig-gauss-der1-relscaleerr}--\ref{fig-gauss-der4-relscaleerr}
show how the $l_1$-norms as well as the spatial spread measures vary 
as function of the scale parameter, with comparisons to the
scale dependencies for the corresponding fully continuous measures.

\subsection{Integrated Gaussian derivative kernels}

In analogy with the treatment of the integrated Gaussian kernel in
Section~\ref{sec-int-gauss}, a possible way of making
the $l_1$-norm of the discrete
approximation of a Gaussian derivative kernel closer to the
$L_1$-norm of its continuous counterpart, is by defining the discrete
kernel as the integral of the continuous Gaussian derivative kernel
over each pixel support region
\begin{equation}
  \label{eq-def-int-gauss-der}
  T_{\intdisc,x^{\alpha}}(n;\, s) = \int_{x = n - 1/2}^{n + 1/2} g_{x^{\alpha}}(x;\; s) \, dx,
\end{equation}
again with a physical motivation of extending the discrete input
signal $f(n)$ to a continuous input signal $f_c(x)$, defined to be equal
to the discrete value within each pixel support region, and then
integrating that continuous input signal with a continuous Gaussian
kernel, which does then correspond to convolving the discrete input
signal with the corresponding integrated Gaussian derivative kernel
(see Appendix~\ref{app-deriv-int-gauss} for an explicit derivation).

Given that $g_{x^{\alpha - 1}}(x;\; s)$ is a primitive function of
$g_{x^{\alpha}}(x;\; s)$, we can furthermore for $\alpha \geq 1$,
write the relationship (\ref{eq-def-int-gauss-der}) as
\begin{equation}
  T_{\intdisc,x^{\alpha}}(n;\; s)
  = g_{x^{\alpha-1}}(n + \tfrac{1}{2};\; s) - g_{x^{\alpha-1}}(n - \tfrac{1}{2};\; s).
\end{equation}
With this definition, it
follows immediately that the contributions to the $l_1$-norm of the discrete kernel
$T_{\intdisc,x^{\alpha}}(n;\; s)$ will be equal to the contributions to the $L_1$-norm of
$g_{x^{\alpha}}(n;\; s)$ over those pixels where the continuous kernel
has the same sign over the entire pixel support region. For those
pixels where the continuous kernel changes its sign within the support
region of the pixel, however, the
contributions will be different, thus implying that the contributions
to the $l_1$-norm of the discrete kernel may be lower
than the contributions to the $L_1$-norm of the
corresponding continuous Gaussian derivative kernel
(see Figure~\ref{fig-kernel-graphs} for an illustration of such graphs
of integrated Gaussian derivative kernels).

Similarly to the previously treated case with the integrated Gaussian
kernel, the integrated Gaussian derivative kernels will also imply a
certain scale offset, as shown in
Figures~\ref{fig-gauss-der1-relscaleerr}--\ref{fig-gauss-der4-relscaleerr} and
Figures~\ref{fig-gauss-der1-relscaleerr-diff}--\ref{fig-gauss-der4-relscaleerr-diff}. 

\subsection{Discrete analogues of Gaussian derivative kernels}

A common characteristics of the approximation methods for computing
discrete Gaussian derivative responses considered so far, is that the computation
of each Gaussian derivative operator of a given order will imply a
spatial convolution with a large-support kernel.
Thus, the amount of necessary computational work will increase by the
number of Gaussian derivative responses, that are to be used when
constructing visual operations that base their processing steps on
using Gaussian derivative responses as input.

A characteristic property of the theory for discrete derivative
approximations with scale-space properties in Lindeberg
(\citeyear{Lin93-JMIV,Lin93-Dis}), however, is that discrete derivative approximations
can instead be computed by applying small-support central difference operators to the
discrete scale-space representation, and with preserved scale-space
properties in terms of either (i)~non-creation of local extrema with
increasing scale for 1-D signals, or (ii)~non-enhancement of local extrema towards
increasing scales in arbitrary dimensions. With regard to the amount of computational work,
this property specifically means that the amount of additive
computational work needed, to add more Gaussian derivative responses as
input to a visual module, will be substantially lower than for the
previously treated discrete approximations, based on computing each
Gaussian derivative response using convolutions with large-support
spatial filters.

According to the genuinely discrete theory for defining discrete
analogues of Gaussian derivative operators, discrete derivative
approximations are from the discrete scale-space representation,
generated by convolution with the discrete
analogue of the Gaussian kernel according to (\ref{eq-disc-gauss})
\begin{equation}
  L(\cdot;\; s) = T_{\disc}(\cdot;\; s) * f(\cdot),
\end{equation}
computed as
\begin{equation}
  \label{eq-disc-der-scsp-1D}
  L_{x^{\alpha}}(x;\; s) = (\delta _{x^{\alpha}} L)(x;\; s),
\end{equation}
where $\delta _{x^{\alpha}}$ are small-support difference operators of
the following forms in the special cases when $\alpha = 1$ or $\alpha = 2$
\begin{align}
  \begin{split}
    \delta_x & = (-\tfrac{1}{2}, 0, +\tfrac{1}{2}),
   \end{split}\\
 \begin{split}
    \delta_{xx} & = (+1, -2, +1),
   \end{split}
\end{align}
to ensure that the estimates of the first- and second-order
derivatives are located at the pixel values, and not in between,
and of the following forms for higher values of $\alpha$:
\begin{equation}
  \label{eq-def-cent-diff-op-arb-order}
  \delta_{x^{\alpha}}
  = \left\{
         \begin{array}{ll}
           \delta_x (\delta_{xx})^i & \mbox{if $\alpha = 1 + 2 i$,} \\
            (\delta_{xx})^i & \mbox{if $\alpha = 2 i$,}
          \end{array}
        \right.
\end{equation}
for integer $i$, where the special cases $\alpha = 3$ and $\alpha = 4$
then correspond to the difference operators
\begin{align}
  \begin{split}
    \delta_{xxx} & = (-\tfrac{1}{2}, +1, 0, -1, +\tfrac{1}{2}),
   \end{split}\\
 \begin{split}
    \delta_{xxxx} & = (+1, -4, +6, -4, +1).
   \end{split}
\end{align}
For 2-D images, corresponding discrete derivative
approximations are then computed as straightforward extensions
of the 1-D discrete derivative approximation operators
\begin{equation}
  \label{eq-disc-der-scsp-2D}
  L_{x^{\alpha} y^{\beta}}(x, y;\; s)
  = (\delta _{x^{\alpha} y^{\beta}} L)(x, y;\; s)
  = (\delta _{x^{\alpha}} \delta_{y^{\beta}} L)(x, y;\; s),  
\end{equation}
where $L(x, y;\; s)$ here denotes the discrete scale-space
representation (\ref{eq-2D-disc-scsp}) computed using
separable convolution with the discrete analogue of the
Gaussian kernel (\ref{eq-disc-gauss}) along each dimension.

In terms of explicit convolution kernels, computation of these types
of discrete derivative approximations correspond to applying discrete
derivative approximation kernels of the form
\begin{equation}
  \label{eq-disc-der-gauss}
  T_{\disc,x^{\alpha}}(n;\; s) = (\delta_{x^{\alpha}} T_{\disc})(n;\; s)
\end{equation}
to the input data. 
In practice, such explicit derivative approximation
kernels should not, however, never be applied for actual computations
of discrete Gaussian derivative responses, since those operations can
be carried out much more efficiently by computations of the forms
(\ref{eq-disc-der-scsp-1D}) or (\ref{eq-disc-der-scsp-2D}), provided
that the computations are carried out with sufficiently high numerical
accuracy, so that the numerical errors do not grow too much because of
cancellation of digits. 

\subsubsection{Cascade smoothing property}

A theoretically attractive property of these types of discrete
approximations of Gaussian derivative operators, is that they exactly
obey a cascade smoothing property over scales, in 1-D of the form
\begin{equation}
  L_{x^{\alpha}}(x;\; s_2)
  = T_{\disc}(\cdot;\; s_2 - s_1) * L_{x^{\alpha}}(\cdot;\; s_1),
\end{equation}
and in 2-D of the form
\begin{equation}
  L_{x^{\alpha} y^{\beta}}(\cdot, \cdot;\; s_2)
  = T_{\disc}(\cdot, \cdot;\; s_2 - s_1) * L_{x^{\alpha} y^{\beta}}(\cdot, \cdot;\; s_1),
\end{equation}
where $T_{\disc}(\cdot, \cdot;\; s)$ here denotes the 2-D extension of the 
1-D discrete analogue of the Gaussian kernel by separable convolution
\begin{equation}
  T_{\disc}(m, n;\; s) = T_{\disc}(m;\; s) \, T_{\disc}(n;\; s).
\end{equation}
In practice, this cascade smoothing property implies that the
transformation from any finer level of scale to any coarser level of
scale is always a simplifying transformation, implying that this
transformation always ensures: (i)~non-creation of new local extrema
(or zero-crossings) from finer to coarser levels of scale for 1-D
signals, and (ii)~non-enhancement of local extrema, in the sense that
the derivative of the scale-space representation with respect to the
scale parameter, always satisfies $\partial_{s} L \leq 0$ at any local
spatial maximum point and $\partial_{s} L \geq 0$ at any local
spatial minimum point.

\begin{figure*}[hbpt]
  \begin{center}
    
    \begin{subfigure}[t]{0.5\textwidth}
       \begin{tabular}{c}
          {\em 1st-order derivative approximation applied to $f(x) = x$\/} \\
          \includegraphics[width=0.90\textwidth]{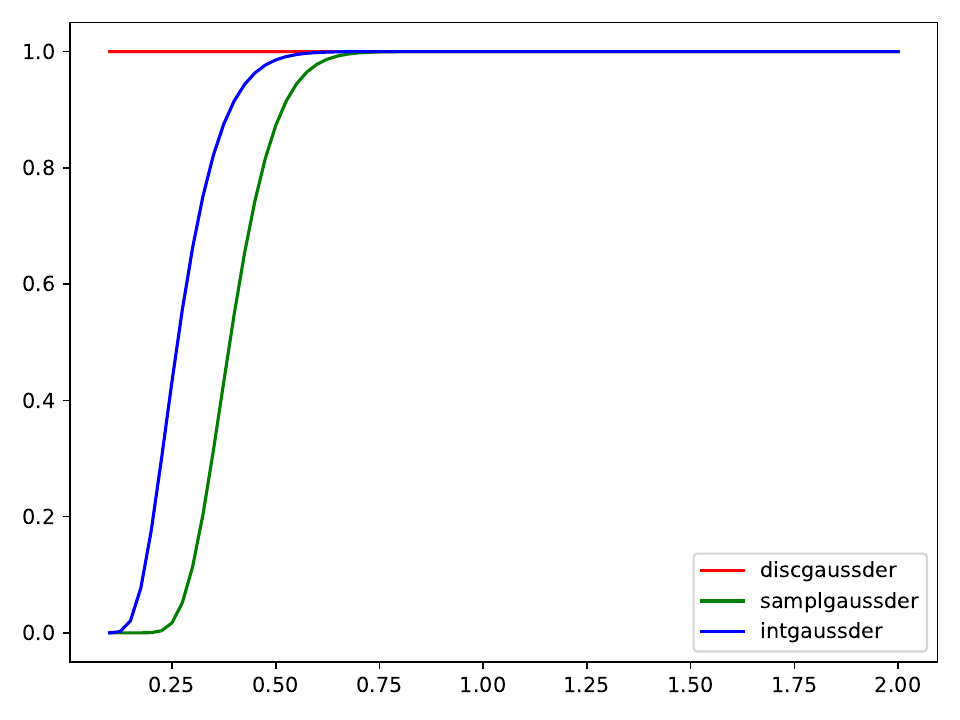}
       \end{tabular}
       \caption{Case: $M = 1$, Ideal value: $1! = 1$.}
     \label{fig-mon1-der1-resp}
     \end{subfigure}%
     ~
    \begin{subfigure}[t]{0.5\textwidth}
       \begin{tabular}{c}
         {\em 2nd-order derivative approximation applied to $f(x) = x^2$\/} \\
         \includegraphics[width=0.90\textwidth]{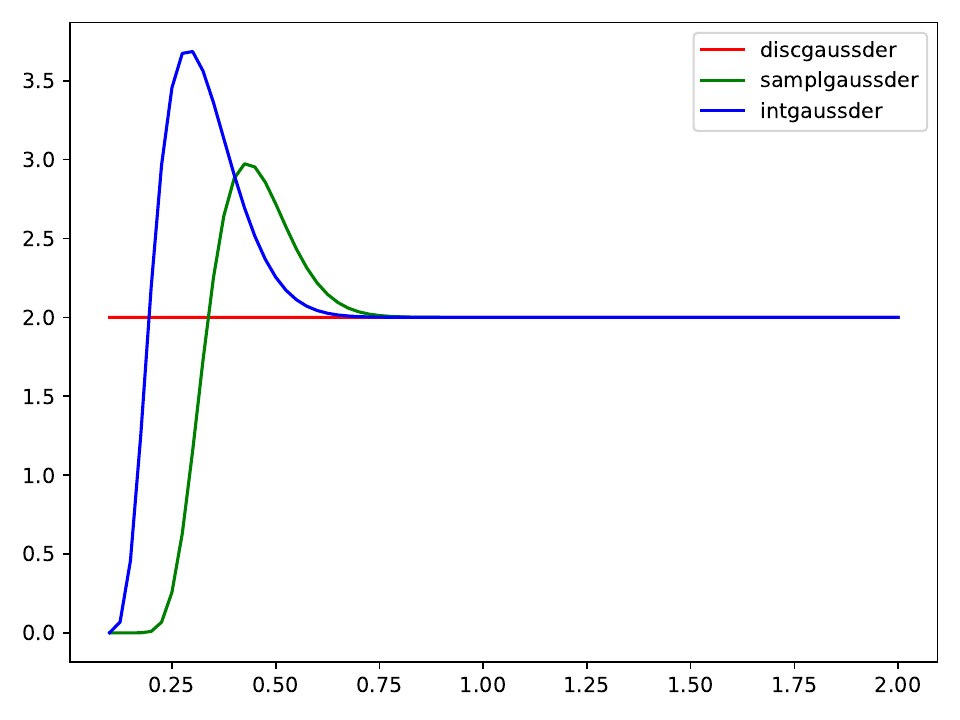}
       \end{tabular}
       \caption{Case: $M = 2$, Ideal value: $2! = 2$.}
    \label{fig-mon2-der2-resp}
    \end{subfigure}

    \bigskip
    \bigskip    

    \begin{subfigure}[t]{0.5\textwidth}
       \begin{tabular}{c}
          {\em 3rd-order derivative approximation applied to $f(x) = x^3$\/} \\
          \includegraphics[width=0.90\textwidth]{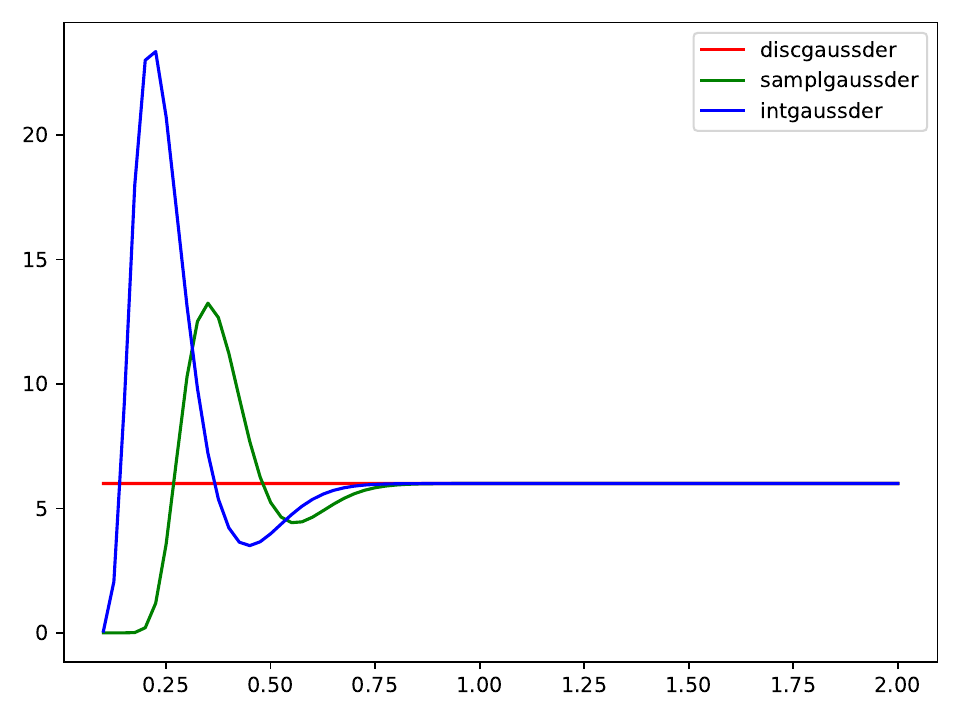}
       \end{tabular}
      \caption{Case: $M = 3$, Ideal value: $3! = 6$.}
    \label{fig-mon3-der3-resp}
    \end{subfigure}%
~
    \begin{subfigure}[t]{0.5\textwidth}
       \begin{tabular}{c}
         {\em 4th-order derivative approximation applied to $f(x) = x^4$\/} \\
         \includegraphics[width=0.90\textwidth]{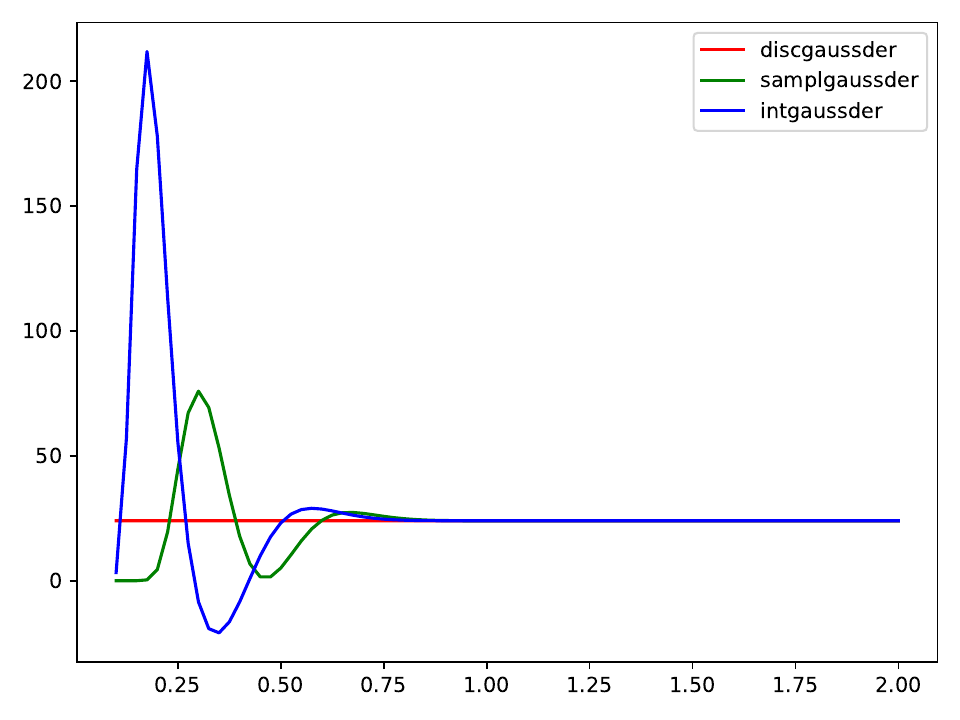}
      \end{tabular}
      \caption{Case: $M = 4$, Ideal value: $4! = 24$.}
   \label{fig-mon4-der4-resp}
    \end{subfigure}
 \end{center}
 
  \caption{The responses to $M$:th-order monomials $f(x) = x^M$ for
    different discrete approximations of $M$:th-order Gaussian
    derivative kernels, for orders up to $M = 4$,
    for either discrete analogues of Gaussian
    derivative kernels $T_{\disc,x^{\alpha}}(n;\; s)$
    according to (\ref{eq-disc-der-gauss}), sampled Gaussian
    derivative kernels $T_{\sampl,x^{\alpha}}(n;\, s)$ according to
    (\ref{eq-sampl-gauss-der}) or integrated Gaussian derivative
    kernels $T_{\intdisc,x^{\alpha}}(n;\, s)$ according to
    (\ref{eq-def-int-gauss-der}).
    In the ideal continuous case, the resulting value should be equal
    to $M!$.
    (Horizontal axis: Scale parameter in units of
    $\sigma = \sqrt{s} \in [0.1, 2]$.)}
\end{figure*}

\begin{figure*}[hbpt]
  \begin{center}

     \begin{subfigure}[t]{0.5\textwidth}
        \begin{tabular}{c}
           {\em 3rd-order derivative approximation applied to $f(x) = x$\/} \\
          \includegraphics[width=0.90\textwidth]{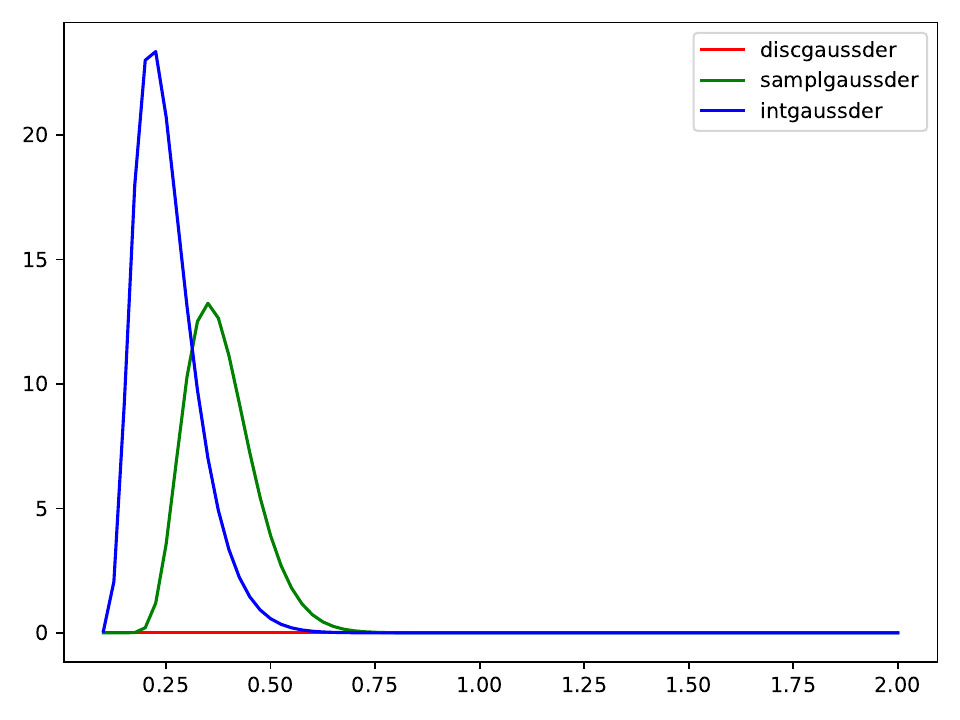}
        \end{tabular}
        \caption{Case $N = 1$, $M = 3$, Ideal value: 0.}
       \label{fig-mon1-der3-resp}
     \end{subfigure}%
     ~
     \begin{subfigure}[t]{0.5\textwidth}     
        \begin{tabular}{c}
            {\em 4th-order derivative approximation applied to $f(x) = x^2$\/} \\
            \includegraphics[width=0.90\textwidth]{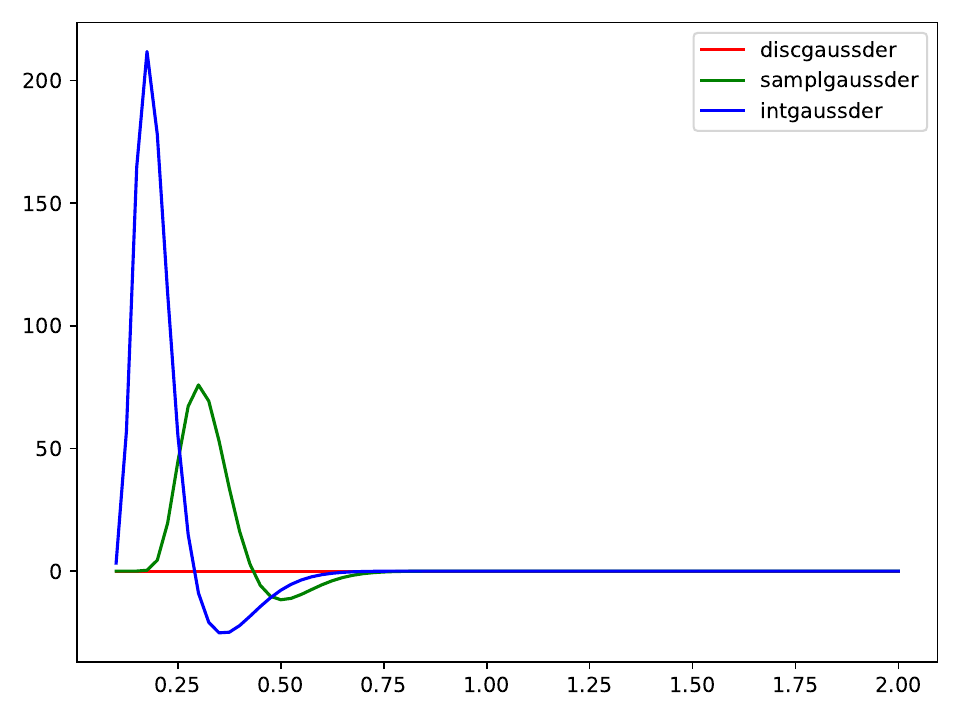}
         \end{tabular}
  \caption{Case $N = 2$, $M = 4$, Ideal value: 0.}
    \label{fig-mon2-der4-resp}
  \end{subfigure}
  \end{center}
  \caption{The responses to different $N$:th-order monomials $f(x) = x^N$ for
    different discrete approximations of $M$:th-order Gaussian
    derivative kernels, for $M > N$, for either discrete analogues of Gaussian
    derivative kernels $T_{\disc,x^{\alpha}}(n;\; s)$
    according to (\ref{eq-disc-der-gauss}), sampled Gaussian
    derivative kernels $T_{\sampl,x^{\alpha}}(n;\, s)$ according to
    (\ref{eq-sampl-gauss-der}) or integrated Gaussian derivative
    kernels $T_{\intdisc,x^{\alpha}}(n;\, s)$ according to
    (\ref{eq-def-int-gauss-der}).
   In the ideal continuous case, the resulting value should be equal
    to 0, whenever the order $M$ of differentiation is higher than the
    order $N$ of the monomial.
    (Horizontal axis: Scale parameter in units of
    $\sigma = \sqrt{s} \in [0.1, 2]$.)}
\end{figure*}

\subsection{Numerical correctness of the derivative estimates}
\label{sec-num-corr-ders}

To measure how well a discrete approximation of a Gaussian derivative
operator reflects a differentiation operator, one can study the
response properties to polynomials.%
\footnote{More generally, the influence of the Gaussian smoothing
  operation on polynomial input can be described by diffusion
  polynomials, as described in
  Appendix~\ref{sec-app-cont-diff-poly}. Given a monomial
  $p_k(x) = x^k$ as input, the result of convolving this input signal
  with a Gaussian kernel is described by a diffusion polynomial
  $q_k(x;\, s)$, with explicit expressions for these up to order 4
  in Equations~(\ref{eq-diff-pol-0})--(\ref{eq-diff-pol-4}). By combining such diffusion
  polynomials with spatial differentiation operations, we can obtain
  closed-form expressions for the results of applying Gaussian
  derivative operators to any polynomial.}
Specifically, in the 1-D case, the
$M$:th-order derivative of an $M$-order monomial should be:
\begin{equation}
  \partial_{x^M} ( x^M ) = M!.
\end{equation}
Additionally, the derivative of any lower-order polynomial should be zero:
\begin{equation}
  \partial_{x^M} ( x^N ) = 0 \quad\quad \mbox{if $M > N$}.
\end{equation}
With respect to Gaussian derivative responses to monomials of the form
\begin{equation}
  p_k(x) = x^k,
\end{equation}
the commutative property between continuous Gaussian smoothing and the
computation of continuous derivatives then specifically implies that
\begin{equation}
   g_{x^M}(\cdot;\; s) * p_M(\cdot) = M!
\end{equation}
and
\begin{equation}
   g_{x^M}(\cdot;\; s) * p_N(\cdot) = 0 \quad\quad \mbox{if $M > N$}.
\end{equation}
If these relationships are not sufficiently well satisfied for the corresponding
result of replacing a continuous Gaussian derivative operator by a
numerical approximations of a Gaussian derivative, then the
corresponding discrete approximation cannot be regarded as a valid
approximation of the Gaussian derivative operator, that in turn is
intended to reflect the differential structures in the image data.

It is therefore of interest to consider entities of the following type
\begin{equation}
  P_{\alpha,k}(s) =
  \left.
    (T_{x^{\alpha}}(\cdot;\; s) * p_k(\cdot))(x;\; s) \, 
   \right|_{x = 0},
\end{equation}
to characterize how well a discrete approximation
$T_{x^{\alpha}}(n;\; s)$ of a Gaussian
derivative operator of order $\alpha$ serves as a differentiation
operator on a monomial of order $k$.

Figures~\ref{fig-mon1-der1-resp}--\ref{fig-mon4-der4-resp}
and Figures~\ref{fig-mon1-der3-resp}--\ref{fig-mon2-der4-resp}
show the results of computing the responses of
the discrete approximations of Gaussian derivative operators to
different monomials in this way, up to order 4.
Specifically, Figure~\ref{fig-mon1-der1-resp} shows the entity
$P_{1,1}(s)$, which in the continuous case should be equal to 1.
Figure~\ref{fig-mon2-der2-resp} shows the entity
$P_{2,2}(s)$, which in the continuous case should be equal to 2.
Figure~\ref{fig-mon3-der3-resp} shows the entity
$P_{3,3}(s)$, which in the continuous case should be equal to 6.
Figure~\ref{fig-mon4-der4-resp} shows the entity
$P_{4,4}(s)$, which in the continuous case should be equal to 24.
Figure~\ref{fig-mon1-der3-resp} and Figure~\ref{fig-mon2-der4-resp}
show the entities $P_{3,1}(s)$ and $P_{4,2}(s)$, respectively, which
in the continuous case should be equal to zero.

As can be seen from the graphs, the responses of the derivative
approximation kernels to monomials of the same order as the order of
differentiation do for the sampled Gaussian derivative kernels
deviate notably from the corresponding ideal results obtained for continuous Gaussian
derivatives, when the scale parameter is a bit below 0.75.
For the integrated Gaussian derivative kernels, the responses of the
derivative approximation kernels do also deviate when the scale
parameter is a bit below 0.75. Within a narrow range of scale values
in intervals of the order of $[0.5, 0.75]$, the integrated Gaussian derivative
kernels do, however, lead to somewhat lower deviations in the derivative estimates than
for the sampled Gaussian derivative kernels.
Also the responses of the third-order sampled Gaussian and integrated
Gaussian derivative approximation kernels to a first-order monomial as
well as the response of the fourth-order sampled Gaussian and integrated
Gaussian derivative approximation kernels to a second-order monomial
differ substantially from the ideal continuous values when the scale
parameter is a bit below 0.75.

For the discrete analogues of the Gaussian derivative kernels, the 
results are, on the other hand, equal to the corresponding continuous
counterparts, in fact, in the case of exact computations, exactly equal.
This property can be shown by studying the responses of the central difference
operators to the monomials, which are given by
\begin{equation}
  \delta_{x^M} ( x^M ) = M!
\end{equation}
and
\begin{equation}
  \delta_{x^M} ( x^N ) = 0 \quad\quad \mbox{if $M > N$}.
\end{equation}
Since the central difference operators commute with the spatial
smoothing step with the discrete analogue of the Gaussian kernel, the
responses of the discrete analogues of the Gaussian derivatives to the
monomials are then obtained as
\begin{multline}
  P_{\alpha,k}(s)
  = \left.
         T_{\disc,x^{\alpha}}(\cdot; s) * p_k(\cdot) \,
      \right|_{x = 0}
    = \\
    =  \left.
      T_{\disc} (\cdot; s) * (\delta_{x^{\alpha}} \, p_k(\cdot)) \,
      \right|_{x = 0},
\end{multline}
implying that
\begin{equation}
  \label{eq-monom-resp-disc-gauss-equal-diff-order}
   T_{\disc,x^M}(\cdot;\; s) * p_M(\cdot) = M!
\end{equation}
and
\begin{equation}
   \label{eq-monom-resp-disc-gauss-larger-diff-order}
   T_{\disc,x^M}(\cdot;\; s) * p_N(\cdot) = 0 \quad\quad \mbox{if $M > N$}.
\end{equation}
In this respect, there is a fundamental difference between the
discrete approximations of Gaussian derivatives obtained from the
discrete analogues of Gaussian derivatives, compared to the sampled or
the integrated Gaussian derivatives. At very fine scales, the discrete
analogues of Gaussian derivatives produce much better estimates of
differentiation operators, than the sampled or the integrated Gaussian derivatives.

The requirement that the Gaussian derivative operators and their
discrete approximations should lead to numerically accurate derivative
estimates for monomials of the same order as the order of
differentiation is a natural consistency requirement for
non-infinitesimal derivative approximation operators.
The use of monomials as test functions, as used here, is particularly suitable in a
multi-scale context, since the monomials are essentially
scale-free, and are not associated with any particular intrinsic scales.

\subsection{Additional performance measures for quantifying deviations from theoretical
  properties of discretizations of Gaussian derivative kernels}
\label{sec-perf-meas-gauss-ders}

To additionally quantify the deviations between the properties of the discrete
kernels, designed to approximate Gaussian derivative operators, and
desirable properties of discrete kernels, that are to transfer the
desirable properties of the continuous Gaussian derivatives to a
corresponding discrete implementation, we will in this section
quantify these deviations in terms of the following complementary error measures:
\begin{itemize}
\item
  {\bf Normalization error:}
  The difference between the $l_1$-norm of the discrete kernel and
  the desirable $l_1$-normal\-ization to a similar $L_1$-norm as for the
  continuous Gaussian derivative kernel will be measured by
   \begin{equation}
     E_{\norm}(T_{x^{\alpha}}(\cdot;\; s))
     = \frac{\| T_{x^{\alpha}}(\cdot;\; s) \|_1}{\| g_{x^{\alpha}}(\cdot;\; s) \|_1}
         - 1.
   \end{equation}
\item 
  {\bf Spatial spread measure:}
  The spatial extent of the discrete derivative approximation kernel
  will be measured by the entity
  \begin{equation}
    \label{eq-def-rel-scale-err-gauss-ders}
     \sqrt{V(|T_{x^{\alpha}}(\cdot;\; s)|)}
  \end{equation}
  and will be graphically compared to the spread measure
  $S_{\alpha}(s) = V(|g_{x^{\alpha}}(\cdot;\; s)|)$ for a
  corresponding continuous Gaussian derivative kernel.
  Explicit expressions for the latter spread measures $S_{\alpha}(s)$
  computed from continuous Gaussian derivative kernels are given
  in Appendix~\ref{app-spread-measures}.
\item 
  {\bf Spatial spread measure offset:}
  To quantify the absolute deviation between the above measured spatial spread
  measure $\sqrt{V(|T_{x^{\alpha}}(\cdot;\; s)|)}$ with the
  corresponding ideal value $V(|g_{x^{\alpha}}(\cdot;\; s)|)$
  for a continuous Gaussian derivative kernel, we will measure this
  offset in terms of the entity
  \begin{equation}
    \label{eq-spat-spread-meas-offset}
    O_{\alpha}(s)
    = \sqrt{V(|T_{x^{\alpha}}(\cdot;\; s)|)} - \sqrt{V(|g_{x^{\alpha}}(\cdot;\; s)|)}.
  \end{equation}
\item
  {\bf Cascade smoothing error:}
  The deviation, from the cascade smoothing property of continuous
  Gaussian derivatives according to (\ref{eq-casc-prop-gauss-der})
  and the actual result of convolving a discrete approximation of a
  Gaussian derivative response at a given scale with its corresponding
  discretization of the Gaussian kernel, will be
  measured by
  \begin{multline}
    \label{eq-def-casc-smooth-err}
    E_{\cascade}(T_{x^{\alpha}}(\cdot;\; s)) = \\
    = \frac{\| T_{x^{\alpha}}(\cdot;\; 2s) - T(\cdot;\; s) * T_{x^{\alpha}}(\cdot;\; s) \|_1}
               {\| T_{x^{\alpha}}(\cdot;\; 2s) \|_1}.
  \end{multline}
   For simplicity, we here restrict ourselves to the special case, when
   the scale parameter for the amount of incremental smoothing with a
   discrete approximation of the Gaussian kernel is equal to the scale
   parameter for the finer scale approximation of the Gaussian
   derivative response.%
\footnote{Notably, it could, however, also be of interest to
   study these effects deeper for the case when the scale of the
   incremental smoothing is significantly lower than the scale of the
   discrete approximation of the Gaussian derivative response, which
   we, however, leave to future work.}
\end{itemize}
Similarly to the previous treatment about error measures in
Section~\ref{sec-perf-meas-gauss-smooth}, the normalization error and
the cascade smoothing error should also be equal to zero in the ideal theoretical case.
Any deviations from zero of these error measures do therefore
represent a quantification of deviations from desirable theoretical
properties in a discrete approximation of Gaussian derivative
computations.

\begin{figure*}[hbpt]
  \begin{center}

     \begin{subfigure}[t]{0.5\textwidth}
       \begin{tabular}{c}
          {\em $l_1$-norms of 1st-order derivative approximation kernels\/} \\
          \includegraphics[width=0.90\textwidth]{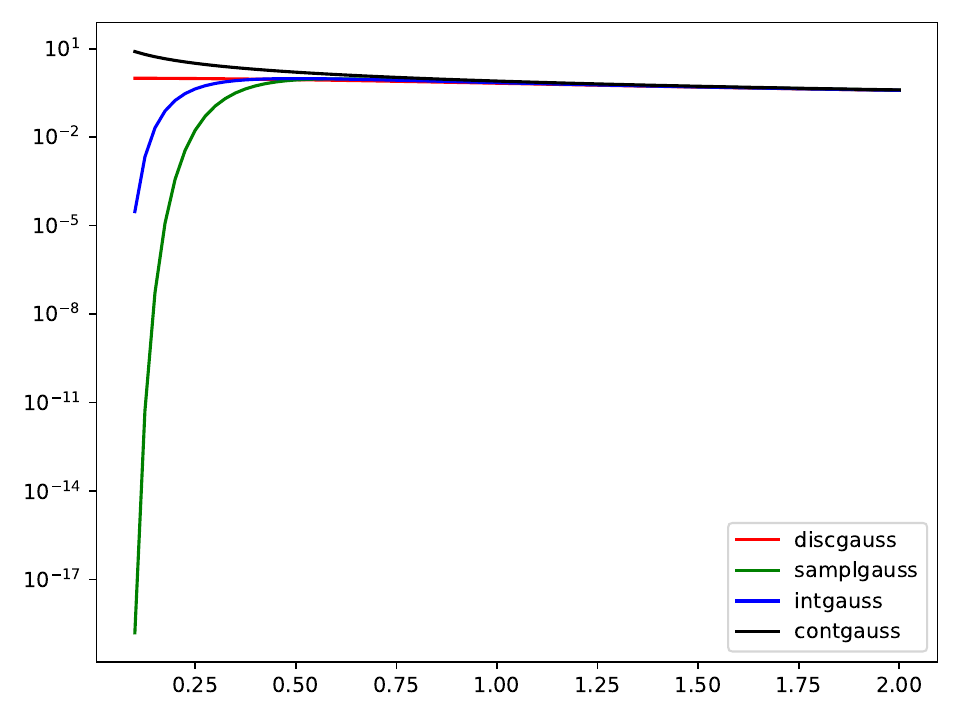}
       \end{tabular}
       \caption{Case: $\alpha = 1$.}
  \label{fig-gauss-der1-L1-norms}
     \end{subfigure}%
     ~
    \begin{subfigure}[t]{0.5\textwidth}
       \begin{tabular}{c}
           {\em $l_1$-norms of 2nd-order derivative approximation kernels\/} \\
           \includegraphics[width=0.90\textwidth]{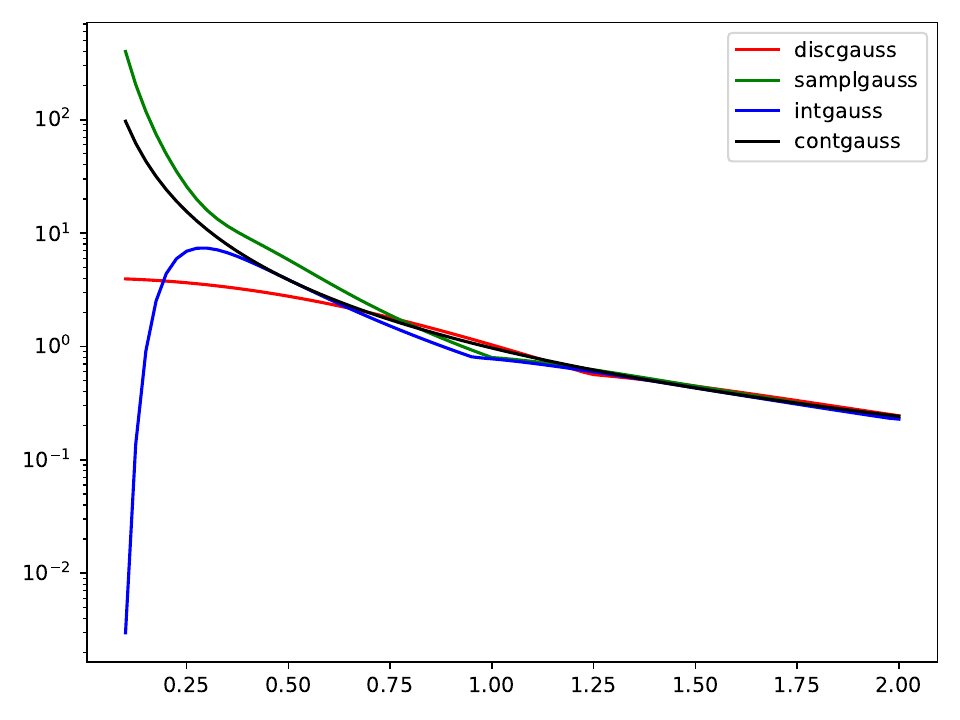}
        \end{tabular}
        \caption{Case: $\alpha = 2$.}
       \label{fig-gauss-der2-L1-norms}
    \end{subfigure}

    \bigskip
    \bigskip    

    \begin{subfigure}[t]{0.5\textwidth}
       \begin{tabular}{c}
          {\em $l_1$-norms of 3rd-order derivative approximation kernels\/} \\
          \includegraphics[width=0.90\textwidth]{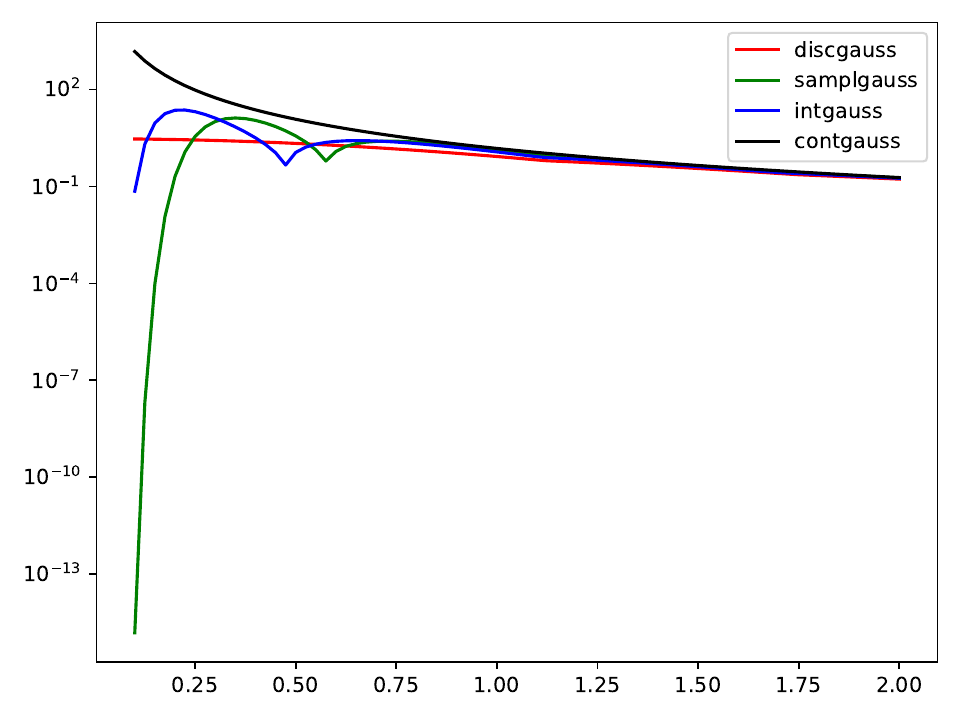}
       \end{tabular}
       \caption{Case: $\alpha = 3$.}
       \label{fig-gauss-der3-L1-norms}
    \end{subfigure}%
~
    \begin{subfigure}[t]{0.5\textwidth}
        \begin{tabular}{c}
           {\em $l_1$-norms of 4th-order derivative approximation kernels\/} \\
           \includegraphics[width=0.90\textwidth]{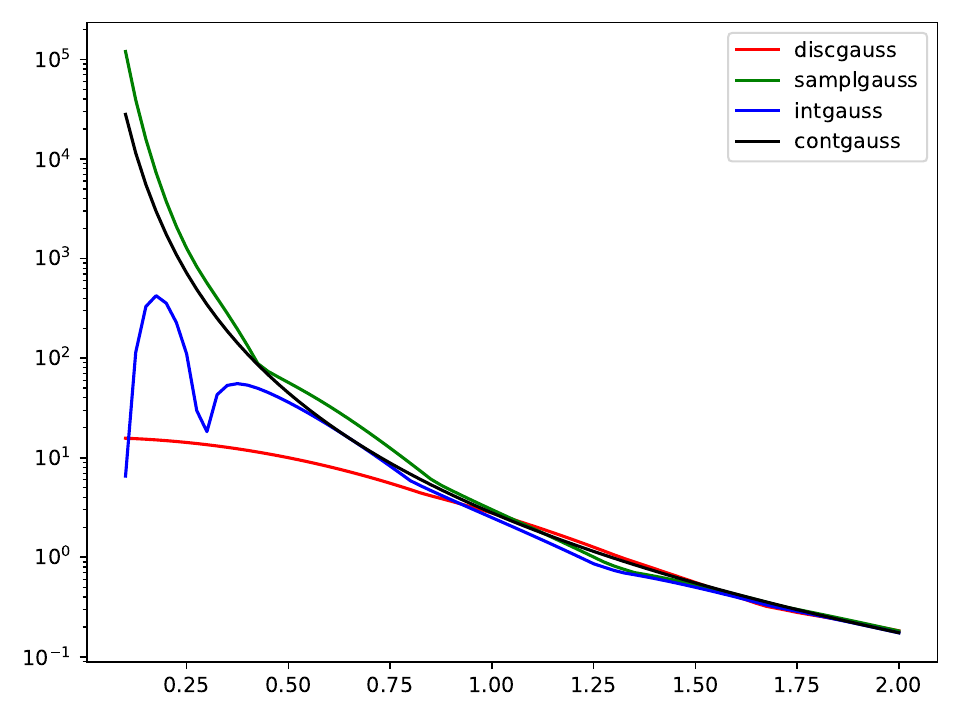}
        \end{tabular}
        \caption{Case: $\alpha = 4$.}
     \label{fig-gauss-der4-L1-norms}
    \end{subfigure}
  \end{center}
  \caption{Graphs of the $l_1$-norms $\| T_{x^{\alpha}}(\cdot;\, s) \|_1$
    of different discrete approximations of Gaussian
    derivative kernels of order $\alpha$, for either discrete analogues of Gaussian
    derivative kernels $T_{\disc,x^{\alpha}}(n;\; s)$
    according to (\ref{eq-disc-der-gauss}), sampled Gaussian
    derivative kernels $T_{\sampl,x^{\alpha}}(n;\, s)$ according to
    (\ref{eq-sampl-gauss-der}) or integrated Gaussian derivative
    kernels $T_{\intdisc,x^{\alpha}}(n;\, s)$ according to
    (\ref{eq-def-int-gauss-der}), together with the graph of the
    $L_1$-norms $\| g_{x^{\alpha}}(\cdot;\, s) \|_1$ of the
    corresponding fourth-order Gaussian derivative kernels.
    (Horizontal axis: Scale parameter in units of
    $\sigma = \sqrt{s} \in [0.1, 2]$.)}
\end{figure*}

\begin{figure*}[hbpt]
  \begin{center}

     \begin{subfigure}[t]{0.45\textwidth}
        \begin{tabular}{c}
            {\em Spatial spread measures for 1st-order derivative kernels\/} \\
            \includegraphics[width=0.97\textwidth]{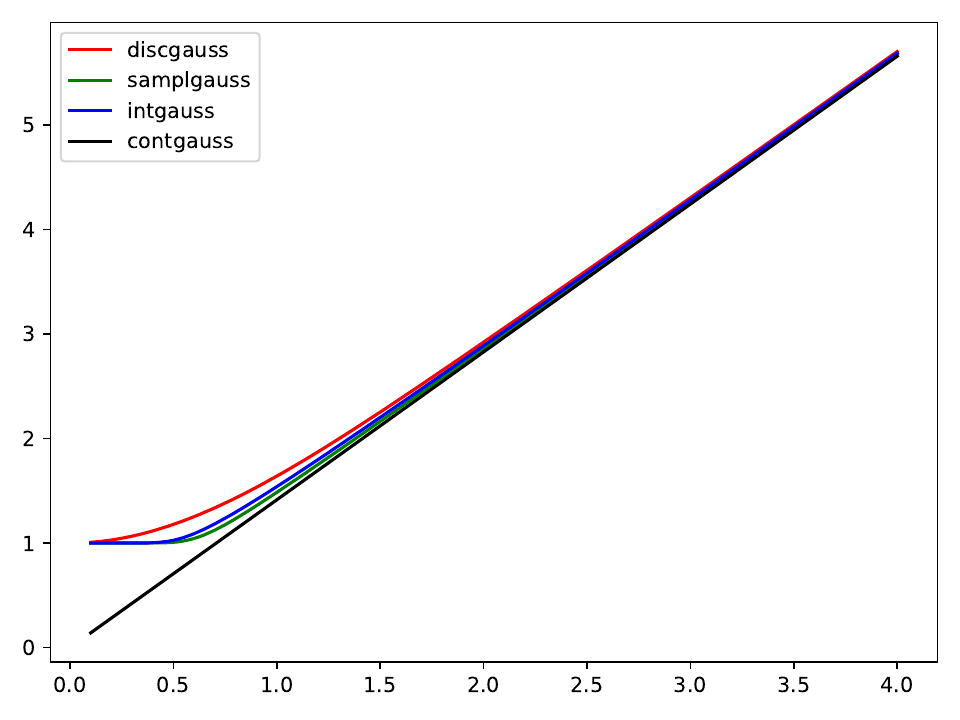}
         \end{tabular}
         \caption{Case: $\alpha = 1$.
    Note that the spatial spread measure of the discrete analogue of
    the first-order Gaussian derivative kernel is delimited from below by the
    spatial variance of the absolute value of the first-order
    central difference operator $|\delta_x|$, which is $\sqrt{V(|\delta_x|)} = 1$.}
      \label{fig-gauss-der1-relscaleerr}
     \end{subfigure}%
     ~\quad\quad~
    \begin{subfigure}[t]{0.45\textwidth}
       \begin{tabular}{c}
          {\em Spatial spread measures for 2nd-order derivative kernels\/} \\
          \includegraphics[width=0.97\textwidth]{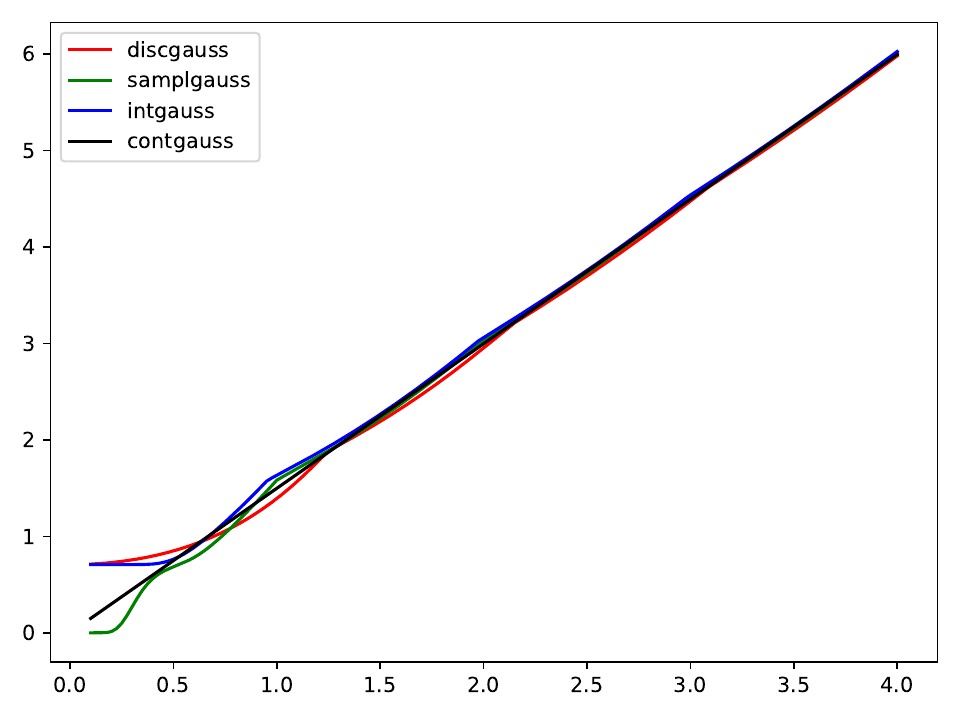}
        \end{tabular}
        \caption{Case: $\alpha = 2$. 
    Note that the spatial spread measure of the discrete analogue of
    the second-order Gaussian derivative kernel is delimited from below by the
    spatial variance of the absolute value of the second-order
    central difference operator $|\delta_{xx}|$,
    which is $\sqrt{V(|\delta_{xx}|)} = 1/\sqrt{2}$.}
      \label{fig-gauss-der2-relscaleerr}
    \end{subfigure}

    \bigskip
    \bigskip    

    \begin{subfigure}[t]{0.45\textwidth}
       \begin{tabular}{c}
          {\em Spatial spread measures for 3rd-order derivative kernels\/} \\
          \includegraphics[width=0.97\textwidth]{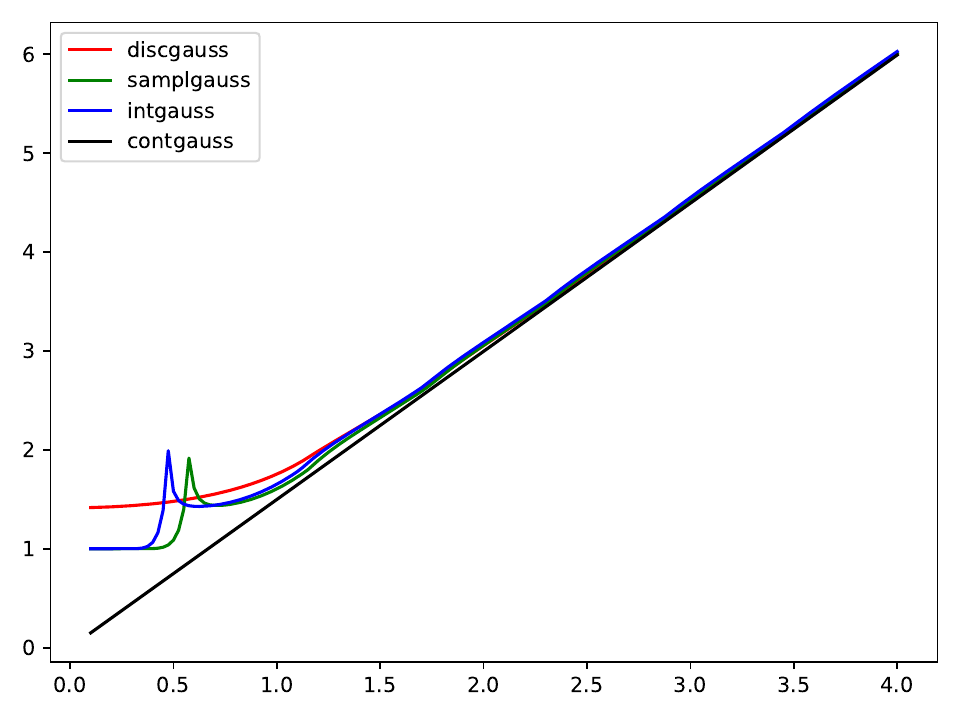}
       \end{tabular}
       \caption{Case: $\alpha = 3$.  
    Note that the spatial spread measure of the discrete analogue of
    the third-order Gaussian derivative kernel is delimited from below by the
    spatial variance of the absolute value of the third-order
    central difference operator $|\delta_{xxx}|$,
    which is $\sqrt{V(|\delta_{xxx}|)} = \sqrt{2}$.}
     \label{fig-gauss-der3-relscaleerr}
    \end{subfigure}%
     ~\quad\quad~
    \begin{subfigure}[t]{0.45\textwidth}
       \begin{tabular}{c}
           {\em Spatial spread measures for 4th-order derivative kernels\/} \\
           \includegraphics[width=0.97\textwidth]{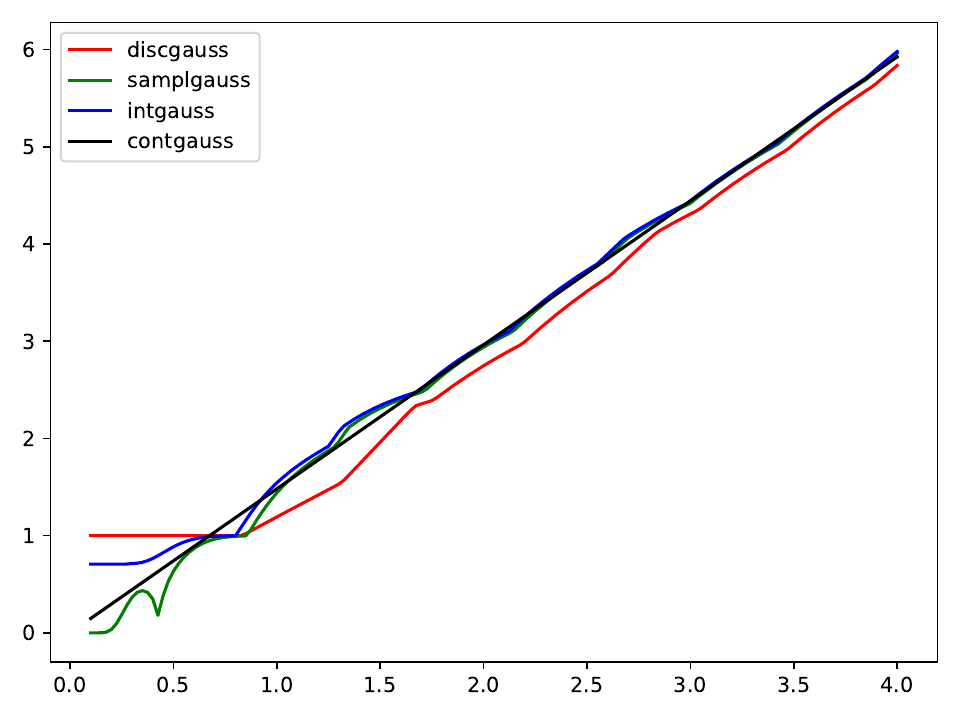}
       \end{tabular}
       \caption{Case: $\alpha = 4$. 
    Note that the spatial spread measure of the discrete analogue of
    the fourth-order Gaussian derivative kernel is delimited from below by the
    spatial variance of the absolute value of the fourth-order
    central difference operator $|\delta_{xxxx}|$,
    which is $\sqrt{V(|\delta_{xxxx}|)} = 1$.}
     \label{fig-gauss-der4-relscaleerr}
     \end{subfigure}
   \end{center}
   \caption{Graphs of the {\em spatial spread measure\/}
    $\sqrt{V(|T_{x^{\alpha}}(\cdot;\; s)|)}$, according to
    (\ref{eq-def-rel-scale-err-gauss-ders}), for
    different discrete approximations of Gaussian
    derivative kernels of order $\alpha$, for either discrete analogues of Gaussian
    derivative kernels $T_{\disc,x^{\alpha}}(n;\; s)$
    according to (\ref{eq-disc-der-gauss}), sampled Gaussian
    derivative kernels $T_{\sampl,x^{\alpha}}(n;\, s)$ according to
    (\ref{eq-sampl-gauss-der}) or integrated Gaussian derivative
    kernels $T_{\intdisc,x^{\alpha}}(n;\, s)$ according to
    (\ref{eq-def-int-gauss-der}). (Horizontal axis: Scale parameter in units of
    $\sigma = \sqrt{s} \in [0.1, 4]$.)}
\end{figure*}

\begin{figure*}[hbpt]
  \begin{center}

     \begin{subfigure}[t]{0.45\textwidth}
        \begin{tabular}{c}
            {\em Spread measure offsets for 1st-order derivative kernels\/} \\
            \includegraphics[width=0.97\textwidth]{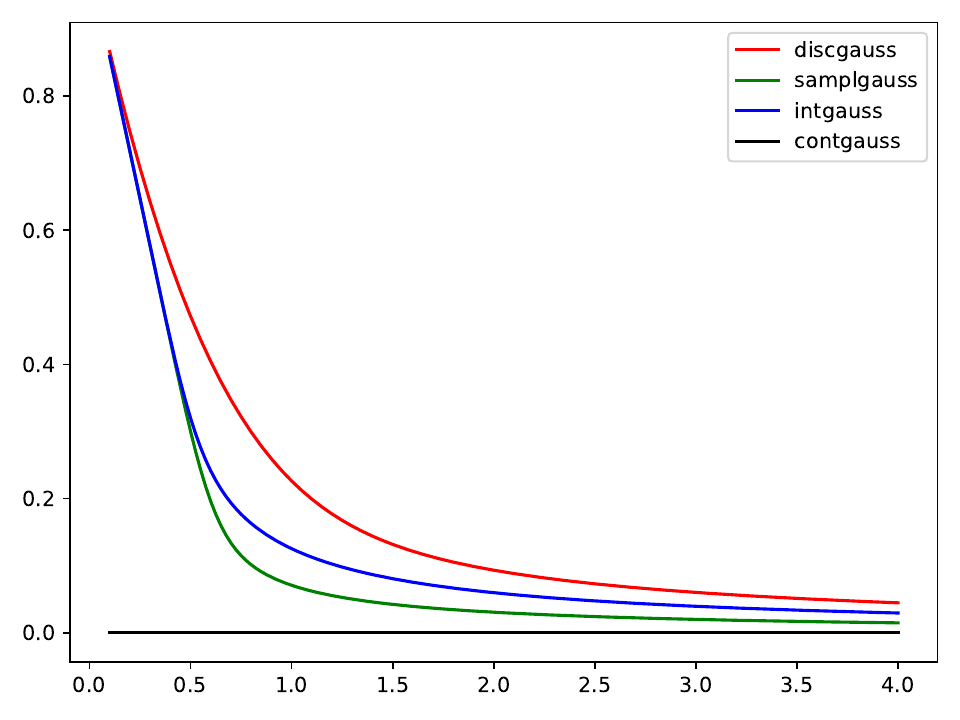}
         \end{tabular}
         \caption{Case: $\alpha = 1$.}
      \label{fig-gauss-der1-relscaleerr-diff}
     \end{subfigure}%
     ~\quad\quad~
    \begin{subfigure}[t]{0.45\textwidth}
       \begin{tabular}{c}
          {\em Spread measure offsets for 2nd-order derivative kernels\/} \\
          \includegraphics[width=0.97\textwidth]{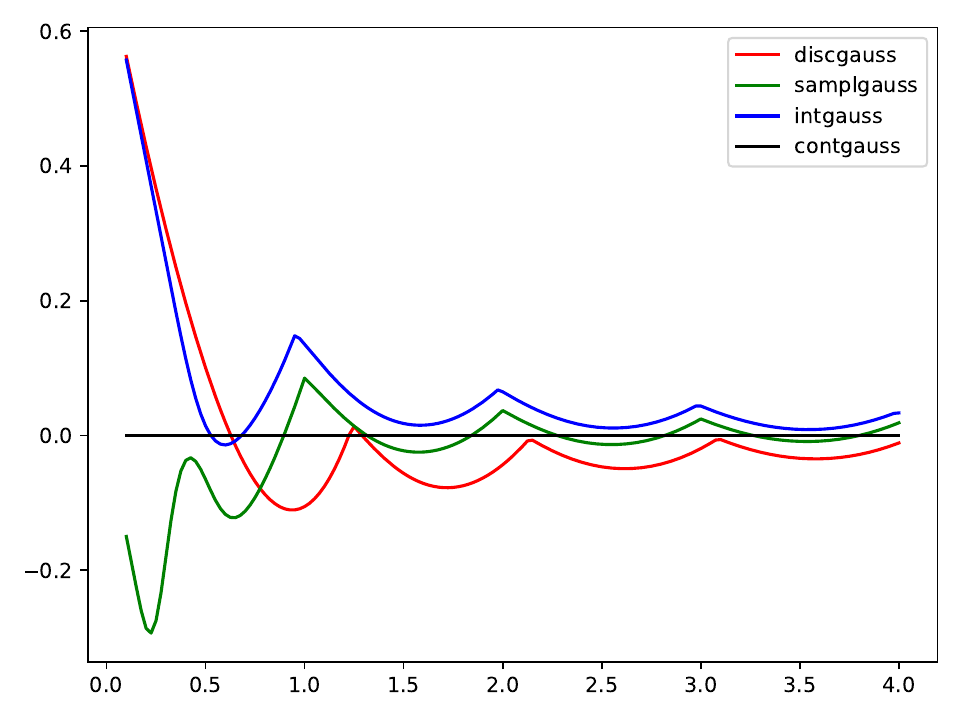}
        \end{tabular}
        \caption{Case: $\alpha = 2$.}
      \label{fig-gauss-der2-relscaleerr-diff}
    \end{subfigure}

    \bigskip
    \bigskip    

    \begin{subfigure}[t]{0.45\textwidth}
       \begin{tabular}{c}
          {\em Spread measure offsets for 3rd-order derivative kernels\/} \\
          \includegraphics[width=0.97\textwidth]{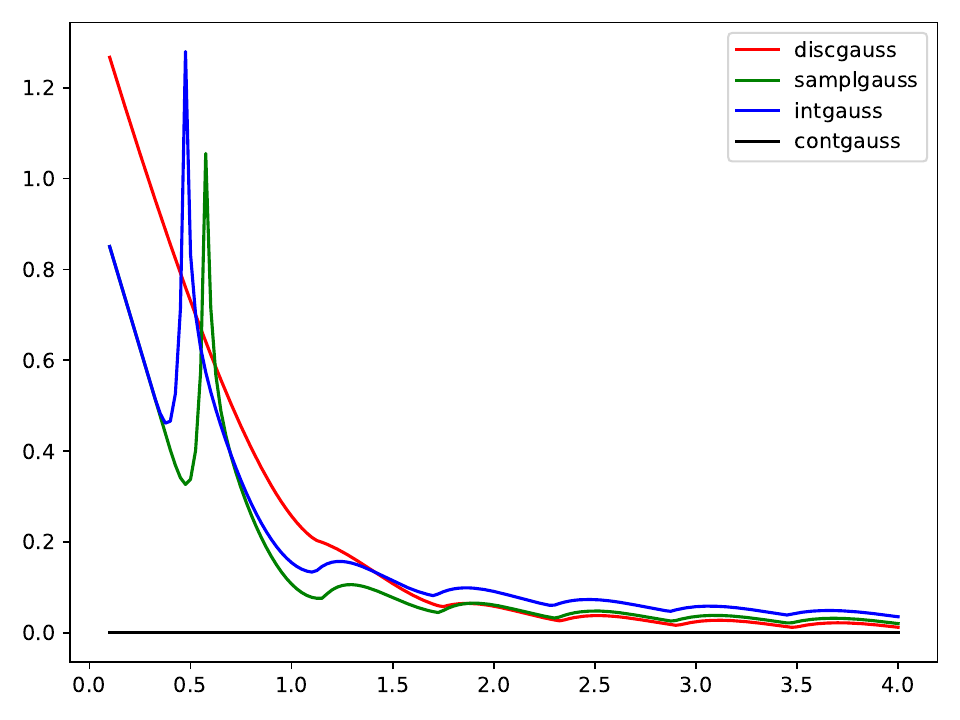}
       \end{tabular}
       \caption{Case: $\alpha = 3$.}
     \label{fig-gauss-der3-relscaleerr-diff}
    \end{subfigure}%
     ~\quad\quad~
    \begin{subfigure}[t]{0.45\textwidth}
       \begin{tabular}{c}
           {\em Spread measure offsets for 4th-order derivative kernels\/} \\
           \includegraphics[width=0.97\textwidth]{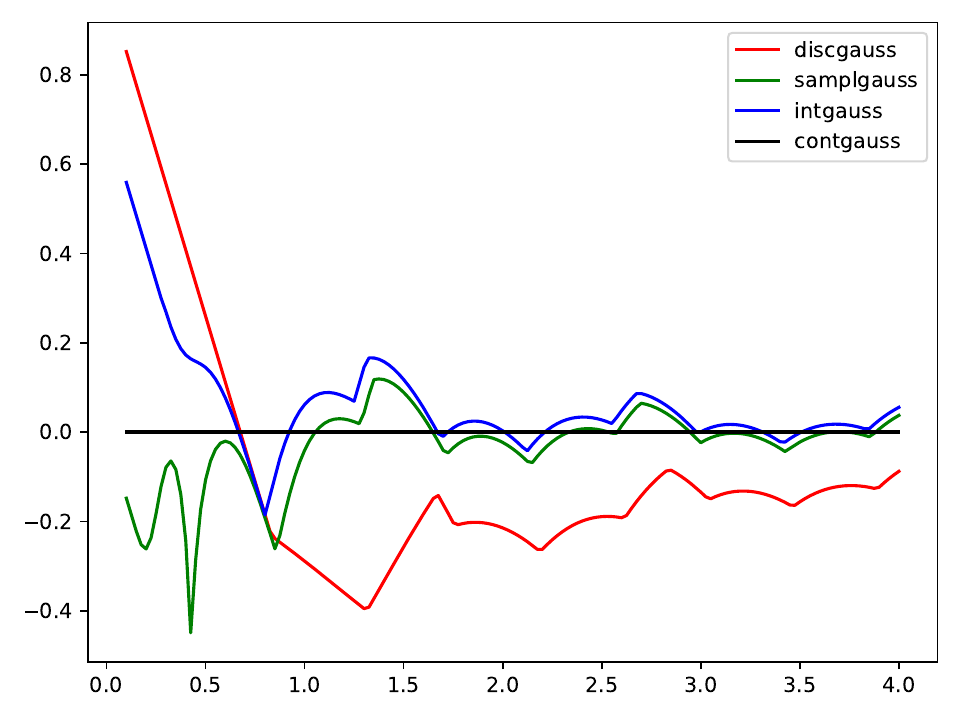}
       \end{tabular}
       \caption{Case: $\alpha = 4$.}
     \label{fig-gauss-der4-relscaleerr-diff}
     \end{subfigure}
   \end{center}
   \caption{Graphs of the {\em spatial spread measure offset\/}
     $O_{\alpha}(s)$, relative to the spatial spread of a continuous
     Gaussian kernel, according to (\ref{eq-spat-spread-meas-offset}), for
    different discrete approximations of Gaussian
    derivative kernels of order $\alpha$, for either discrete analogues of Gaussian
    derivative kernels $T_{\disc,x^{\alpha}}(n;\; s)$
    according to (\ref{eq-disc-der-gauss}), sampled Gaussian
    derivative kernels $T_{\sampl,x^{\alpha}}(n;\, s)$ according to
    (\ref{eq-sampl-gauss-der}) or integrated Gaussian derivative
    kernels $T_{\intdisc,x^{\alpha}}(n;\, s)$ according to
    (\ref{eq-def-int-gauss-der}). (Horizontal axis: Scale parameter in units of
    $\sigma = \sqrt{s} \in [0.1, 4]$.)}
\end{figure*}

\begin{figure*}[hbpt]
  \begin{center}
    
     \begin{subfigure}[t]{0.5\textwidth}    
        \begin{tabular}{c}
           {\em Cascade smoothing error for 1st-order derivative kernels\/} \\
           \includegraphics[width=0.90\textwidth]{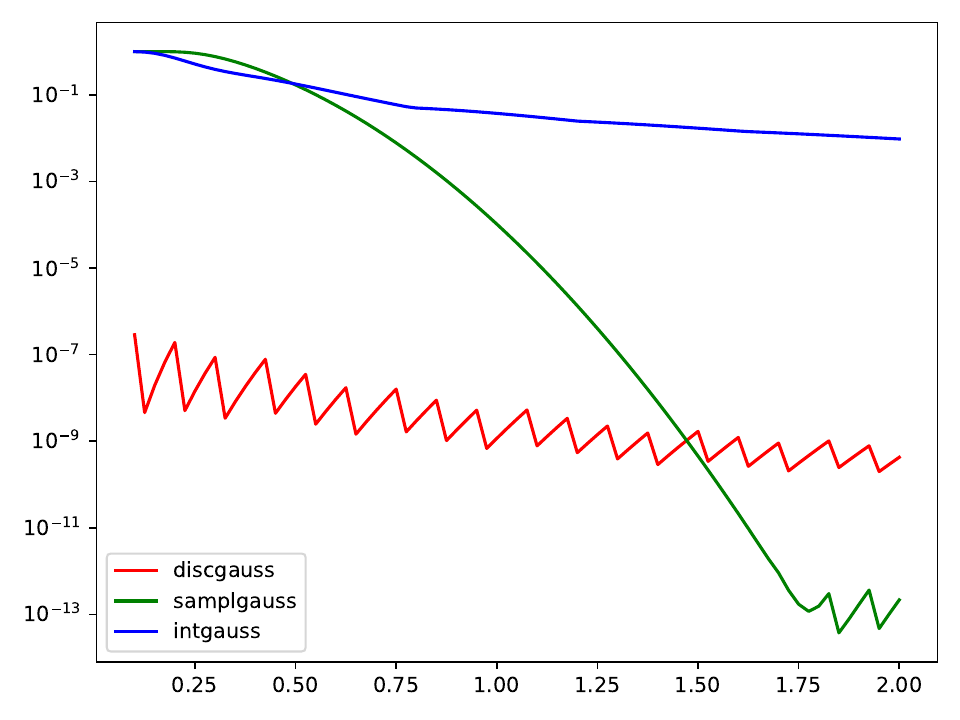}
         \end{tabular}
         \caption{Case: $\alpha = 1$.}
       \label{fig-gauss-der1-cascerr}
     \end{subfigure}%
     ~
    \begin{subfigure}[t]{0.5\textwidth}
       \begin{tabular}{c}
          {\em Cascade smoothing error for 2nd-order derivative kernels\/} \\
          \includegraphics[width=0.90\textwidth]{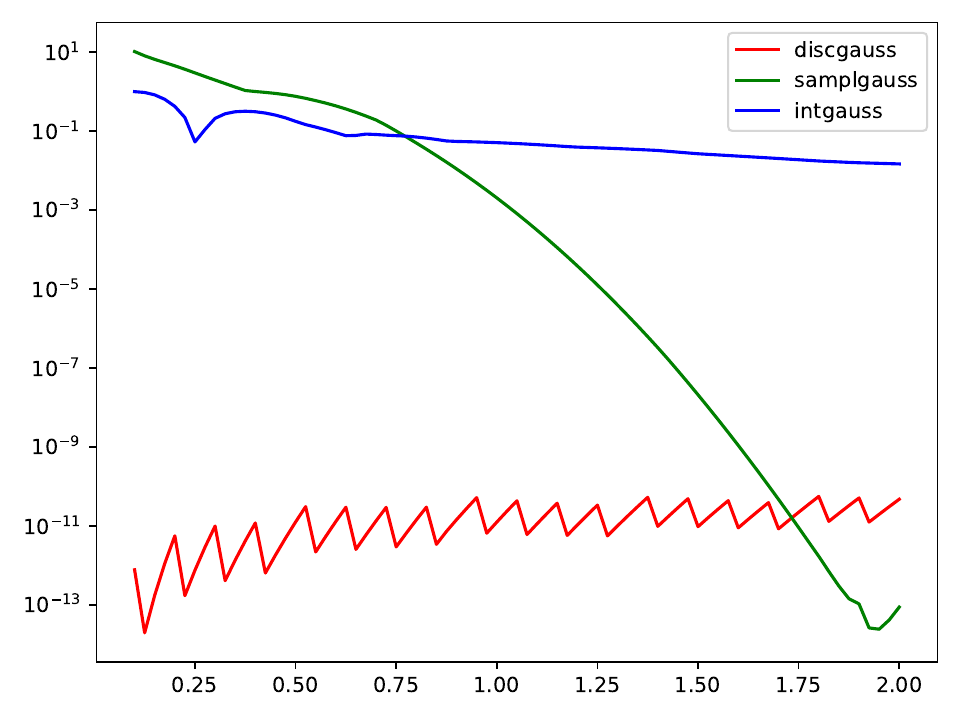}
        \end{tabular}
        \caption{Case: $\alpha = 2$.}
       \label{fig-gauss-der2-cascerr}
    \end{subfigure}

    \bigskip
    \bigskip    

    \begin{subfigure}[t]{0.5\textwidth}
       \begin{tabular}{c}
          {\em Cascade smoothing error for 3rd-order derivative kernels\/} \\
          \includegraphics[width=0.90\textwidth]{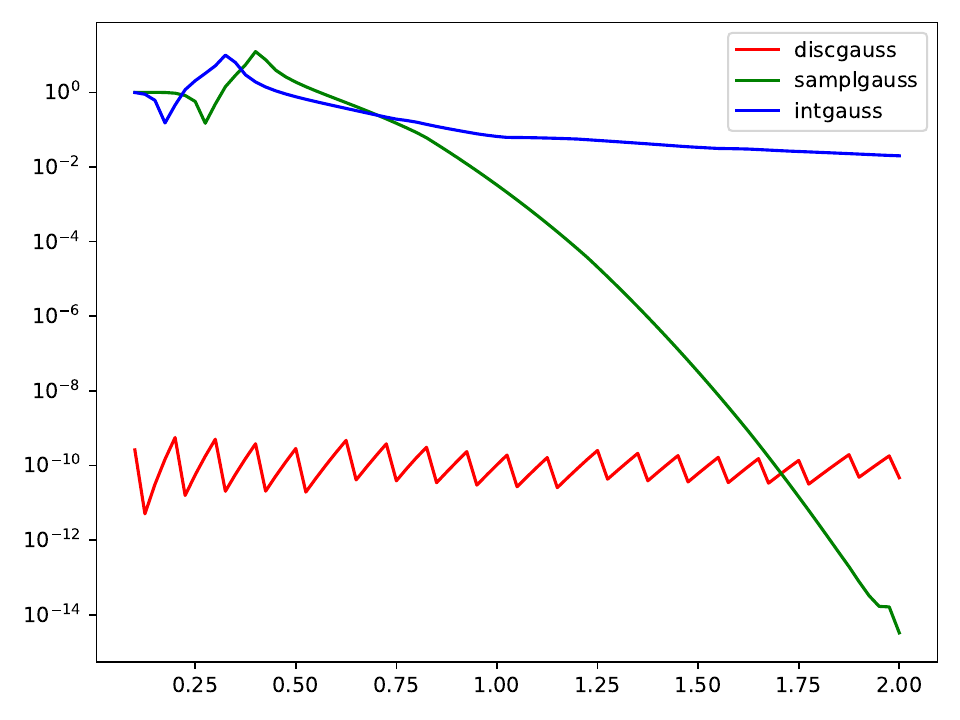}
       \end{tabular}
       \caption{Case: $\alpha = 3$.}
      \label{fig-gauss-der3-cascerr}
    \end{subfigure}%
~
    \begin{subfigure}[t]{0.5\textwidth}
        \begin{tabular}{c}
            {\em Cascade smoothing error for 4th-order derivative kernels\/} \\
            \includegraphics[width=0.90\textwidth]{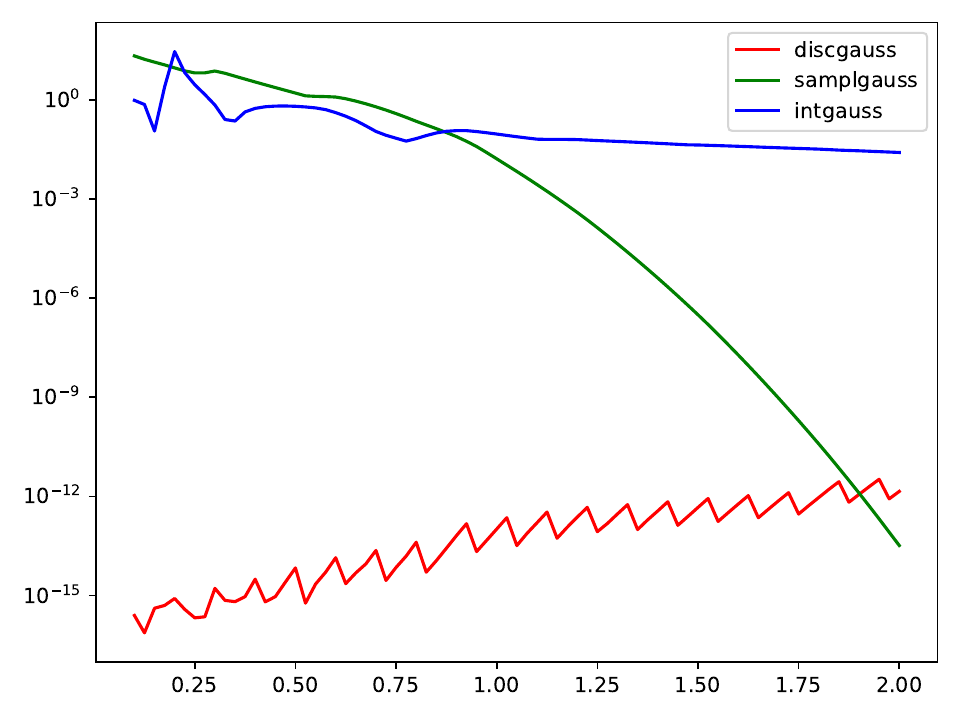}
         \end{tabular}
         \caption{Case: $\alpha = 4$.}
      \label{fig-gauss-der4-cascerr}
    \end{subfigure}
  \end{center}
\caption{Graphs of the {\em cascade smoothing error\/}
    $E_{\cascade}(T_{x^{\alpha}}(\cdot;\; s))$, according to
    (\ref{eq-def-casc-smooth-err}), for
    different discrete approximations of Gaussian
    derivative kernels of order $\alpha$, for either discrete analogues of Gaussian
    derivative kernels $T_{\disc,x^{\alpha}}(n;\; s)$
    according to (\ref{eq-disc-der-gauss}), sampled Gaussian
    derivative kernels $T_{\sampl,x^{\alpha}}(n;\, s)$ according to
    (\ref{eq-sampl-gauss-der}) or integrated Gaussian derivative
    kernels $T_{\intdisc,x^{\alpha}}(n;\, s)$ according to
    (\ref{eq-def-int-gauss-der}).
    (Horizontal axis: Scale parameter in units of
    $\sigma = \sqrt{s} \in [0.1, 2]$.)}
\end{figure*}

\subsection{Numerical quantification of deviations from theoretical
  properties of discretizations of Gaussian derivative kernels}
\label{sec-num-perf-meas-gauss-ders}

\subsubsection{$l_1$-norms of discrete approximations of Gaussian
  derivative approximation kernels}

Figures~\ref{fig-gauss-der1-L1-norms}--\ref{fig-gauss-der4-L1-norms}
show the $l_1$-norms $\| T_{x^{\alpha}}(\cdot;\, s) \|_1$ for the different
methods for approximating Gaussian derivative kernels
with corresponding discrete approximations for differentiation orders
up to~4,
together with graphs of the $L_1$-norms $\| g_{x^{\alpha}}(\cdot;\, s) \|_1$
of the corresponding continuous Gaussian derivative kernels.

From these graphs, we can first observe that the behaviour between the
different methods differ significantly for values of the scale
parameter $\sigma$ up to about 0.75, 1.25 or 1.5, depending on the order of
differentiation. 

For the sampled Gaussian derivatives, the $l_1$-norms tend to zero as
the scale parameter $\sigma$ approaches zero for the kernels of odd
order, whereas the $l_1$-norms tend to infinity for the kernels of
even order. For the kernels of even order, the behaviour of the
sampled Gaussian derivative kernels has the closest similarity to the
behaviour of the corresponding continuous Gaussian derivatives.
For the kernels of odd order, the behaviour is, on the other hand, worst.

For the integrated Gaussian derivatives, the behaviour is
for the kernels of odd order markedly less singular as the
scale parameter $\sigma$ tends to zero, than for the
sampled Gaussian derivatives. 
For the kernels of even order, the behaviour does, on the other hand
differ more. There is also some jaggedness behaviour at fine scales for
the third- and fourth-order derivatives,  caused by positive and
negative values of the kernels cancelling their contribution within
the support regions of single pixels.

For the discrete analogues of Gaussian derivatives, the behaviour is,
qualitatively different for finer scales, in that the discrete analogues of the
Gaussian derivatives tend to the basic central difference operators,
as the scale parameter $\sigma$ tends to zero,
and do therefore show a much more smooth behaviour as $\sigma \rightarrow 0$.

\subsubsection{Spatial spread measures}

Figures~\ref{fig-gauss-der1-relscaleerr}--\ref{fig-gauss-der4-relscaleerr}
show graphs of the standard-deviation-based
spatial spread measure $\sqrt{V(|T_{x^{\alpha}}(\cdot;\; s)|)}$
according to (\ref{eq-def-rel-scale-err-gauss-ders}),
for the main classes of discretizations of Gaussian derivative
kernels, together with graphs of the corresponding spatial spread
measures computed for continuous Gaussian derivative kernels.

As can be seen from these graphs, the spatial spread measures differ
significantly from the corresponding continuous measures for smaller
values of the the scale parameters; for $\sigma$ less than about 1
or 1.5, depending on the order of differentiation, and caused by the
fact that too fine scales in the data cannot be appropriately resolved after a
spatial discretization.
For the sampled and the integrated Gaussian kernels, there is a
certain jaggedness in some of the curves at fine scales, caused by
interactions between the grid and the lobes of the continuous Gaussian kernels, that
the discrete kernels are defined from.
For the discrete analogues of the Gaussian kernels, these spatial
spread measures are notably bounded from below by the corresponding
measures for the central difference operators, that they approach with
decreasing scale parameter.

Figures~\ref{fig-gauss-der1-relscaleerr-diff}--\ref{fig-gauss-der4-relscaleerr-diff}
show more detailed visualizations of the deviations between these
spatial spread measures and their corresponding ideal values for
continuous Gaussian derivative kernels in terms of the spatial spread measure offset
$O_{\alpha}(s)$ according to (\ref{eq-spat-spread-meas-offset}), for
the different orders of spatial differentiation. The jaggedness of
these curves for orders of differentiation greater than one, is due to
interactions between the lobes in the derivative approximation kernels
and the grid. As can be seen from these
graphs, the relative properties of the spatial spread measure offsets
for the different discrete approximations to the Gaussian derivative
operators differ somewhat, depending on the order of spatial
differentiation. We can, however, note that the spatial spread measure
offset for the integrated Gaussian derivative kernels is mostly
somewhat higher than the spatial spread measure offset for the sampled
Gaussian derivative kernels, consistent with the previous observation
that the spatial box integration used for defining the integrated
Gaussian derivative kernel introduces an additional amount of spatial
smoothing in the spatial discretization.

\subsubsection{Cascade smoothing errors}

Figures~\ref{fig-gauss-der1-cascerr}--\ref{fig-gauss-der4-cascerr}
show graphs of the cascade smoothing error
$E_{\cascade}(T_{x^{\alpha}}(\cdot;\; s))$ according to
(\ref{eq-def-casc-smooth-err}) for the main classes of
methods for discretizing Gaussian derivative operators.

For the sampled Gaussian kernels, the cascade smoothing error is
substantial for $\sigma < 0.75$ or $\sigma < 1.0$, depending on the
order of differentiation. Then, for larger scale values, this
error measure decreases rapidly.

For the integrated Gaussian kernels, the cascade smoothing error is
lower than the cascade smoothing error for the sampled Gaussian
kernels for $\sigma < 0.5$, $\sigma < 0.75$ or $\sigma < 1.0$,
depending on the order of differentiation. For larger scale values,
the cascade smoothing error for the integrated Gaussian kernels
do, on the other hand, decrease much less rapidly with increasing
scale than for the sampled Gaussian kernels, due to the additional
spatial variance in the filters caused by the box integration,
underlying the definition of the integrated Gaussian derivative kernels.

For the discrete analogues of the Gaussian derivatives, the cascade
smoothing error should in the ideal case of exact 
computations lead to a zero error. In the graphs of these errors, we
do, however, see a jaggedness at a very low level, caused by
numerical errors.

\subsection{Summary of the characterization results from the theoretical
  analysis and the quantitative performance measures}

To summarize the theoretical and the experimental results presented in
this section, there is a substantial difference in the quality of the
discrete approximations of Gaussian derivative kernels at fine scales:

For values of the scale parameter $\sigma$ less than about a bit below
0.75,
the sampled Gaussian kernels and the integrated Gaussian kernels do
not produce numerically accurate or consistent estimates of the derivatives of
monomials. In this respects, these discrete approximations of Gaussian
derivatives do not serve as good approximations of derivative
operations at very fine scales. Within a narrow scale interval below
about 0.75, the integrated Gaussian derivative kernels do,
however, degenerate in a somewhat less serious manner than the sampled
Gaussian derivative kernels.

For the discrete analogues of Gaussian derivatives, obtained
by convolution with the discrete analogue of the Gaussian kernel
followed by central difference operators, the corresponding derivative
approximation are, on the other hand, exactly equal to their
continuous counterparts. This property does, furthermore, hold over the
entire scale range.

For larger values of the scale parameter, the sampled Gaussian kernel
and the integrated Gaussian kernel do, on the other hand, lead to
successively better numerical approximations of the corresponding
continuous counterparts. In fact, when the value of the scale
parameter is above about 1, the sampled Gaussian kernel leads to the
numerically most accurate approximations of the corresponding
continuous results, out of the studied three methods.

Hence, the choice between what discrete
approximation to use for approximating the Gaussian derivatives,
depends upon what scale ranges are important for the analysis, in
which the Gaussian derivatives should be used.

In next section, we will build upon these results, and extend them further,
by studying the effects
of different discretization methods for the purpose of
performing automatic scale selection.
The motivation for studying that problem, as a benchmark proxy task for
evaluating the quality of different discrete approximations of
Gaussian derivatives, is that it involves explicit comparisons of
feature responses at different scales.

\section{Application to scale selection from local extrema over scale of scale-normalized derivatives}
\label{disc-approx-sc-norm-ders}

When performing scale-space operations at multiple scales jointly, a
critical problem concerns how to compare the responses of an image
operator between different scales. Due to the scale-space smoothing operation,
the amplitude of both Gaussian smoothed image data and of Gaussian
derivative responses can be expected to decrease with scale. A
practical problem then concerns how to compare a response of the same
image operator at some coarser scale to a corresponding response at a
finer scale. This problem is particularly important regarding the
topic of scale selection (Lindeberg \citeyear{Lin21-EncCompVis}),
where the goal is to determine locally
appropriate scale levels, to process and analyze particular image
structures in a given image.

\subsection{Scale-normalized derivative operators}

A theoretically well-founded way of performing scale normalization, to enable
comparison between the responses of scale-space operations at
different scales, is by defining scale-normalized derivative operators
according to (Lindeberg \citeyear{Lin97-IJCV,Lin98-IJCV})
\begin{equation}
  \label{eq-def-sc-norm-ders}
  \partial_{\xi} = s^{\gamma/2} \, \partial_x, \quad\quad
  \partial_{\eta} = s^{\gamma/2} \, \partial_y, 
\end{equation}
where $\gamma > 0$ is a scale normalization power, to be chosen
for the feature detection task at hand,
and then basically replacing the regular Gaussian derivative responses by
corresponding scale-normalized Gaussian derivative responses in the
modules that implement visual operations on image data.

\subsection{Scale covariance property of scale-normalized derivative
  responses}

It can be shown that, given two images $f(x, y)$ and $f'(x', y')$ that
are related according to a uniform scaling transformation
\begin{equation}
  x' = S \, x, \quad\quad y' = S \, y,
\end{equation}
for some spatial scaling factor $S > 0$,
and with corresponding Gaussian derivative responses defined over the
two respective image domains according to
\begin{align}
  \begin{split}
    L_{\xi^{\alpha} \eta^{\beta}}(\cdot, \cdot;\; s)
    & = \partial_{\xi^{\alpha} \eta^{\beta}} (g_{2D}(\cdot, \cdot;\; s) * f(\cdot, \cdot)),
  \end{split}\\
  \begin{split}
      L'_{{\xi'}^{\alpha} {\eta'}^{\beta}}(\cdot, \cdot;\; s')
    & = \partial_{{\xi'}^{\alpha} {\eta'}^{\beta}} (g_{2D}(\cdot, \cdot;\; s') * f'(\cdot, \cdot)),
  \end{split}
\end{align}
these Gaussian derivative responses in the two domains
will then be related according to (Lindeberg \citeyear{Lin97-IJCV}, Equation~(25))
\begin{equation}
  L_{\xi^{\alpha} \eta^{\beta}}(x, y;\; s)
  = S^{(\alpha + \beta)(1 - \gamma)} \, L'_{{\xi'}^{\alpha} {\eta'}^{\beta}}(x', y';\; s'),
\end{equation}
provided that the values of the scale parameters are matched according
to (Lindeberg \citeyear{Lin97-IJCV}, Equation~(15))
\begin{equation}
  s' = S^2 \, s.
\end{equation}
Specifically, in the special case when $\gamma = 1$, the corresponding
scale-normalized Gaussian derivative responses will then be equal
\begin{equation}
  L_{\xi^{\alpha} \eta^{\beta}}(x, y;\; s) = L'_{{\xi'}^{\alpha} {\eta'}^{\beta}}(x', y';\; s').
\end{equation}

\subsection{Scale selection from local extrema over scales of
  scale-normalized derivative responses}

A both theoretically well-founded and experimentally extensively
verified methodology to perform automatic scale selection, is by
choosing hypotheses for locally appropriate scale levels from local
extrema over scales of scale-normalized derivative responses
(Lindeberg \citeyear{Lin97-IJCV,Lin98-IJCV}). In the following, we
will apply this methodology to four basic tasks in feature detection.

\subsubsection{Interest point detection}

With regard to the topic of interest point detection, consider the
scale-normalized Laplacian operator
(Lindeberg \citeyear{Lin97-IJCV} Equation~(30))
\begin{equation}
  \label{eq-sc-norm-lapl}
  \nabla_{norm}^2 L = s \, (L_{xx} + L_{yy}),
\end{equation}
or the scale-normalized determinant of the Hessian
(Lindeberg \citeyear{Lin97-IJCV} Equation~(31))
\begin{equation}
  \label{eq-sc-norm-det-hess}  
  \det {\cal H}_{norm} L = s^2 \, (L_{xx} \, L_{yy} - L_{xy}^2),
\end{equation}
where we have here chosen $\gamma = 1$ for simplicity.
It can then be shown that, if we consider the responses of these
operators to a Gaussian blob of size $s_0$
\begin{equation}
  \label{eq-def-2D-gauss-blob-s0}
  f_{\blob,s_0}(x, y) = g_{2D}(x, y;\; s_0),
\end{equation}
for which the scale-space representation by the semi-group property of
the Gaussian kernel (\ref{eq-semi-group-gauss}) will be of the form
\begin{equation}
  \label{eq-scsprepr-2D-gauss-blob-s0}
  L_{\blob,s_0}(x, y;\; s) = g_{2D}(x, y;\; s_0 + s),
\end{equation}
then the scale-normalized Laplacian response according to
(\ref{eq-sc-norm-lapl}) and the scale-normalized determinant of the
Hessian response according to (\ref{eq-sc-norm-det-hess}) assume their
global extrema over space and scale at 
(Lindeberg \citeyear{Lin97-IJCV}, Equations~(36) and~(37))
\begin{align}
  \begin{split}
    \label{eq-lapl-sc-sel}
    (\hat{x}, \hat{y}, \hat{s}) 
    & = \operatorname{argmin}_{(x, y;\; s)}(\nabla^2 L_{\blob,s_0})(x, y;\; s)
  \end{split}\nonumber\\
  \begin{split}
    & = (0, 0, s_0),
  \end{split}\\
  \begin{split}
    \label{eq-det-hess-sc-sel}
    (\hat{x}, \hat{y}, \hat{s}) 
    & = \operatorname{argmax}_{(x, y;\; s)}(\det {\cal H} L_{\blob,s_0})(x, y;\; s)
  \end{split}\nonumber\\
  \begin{split}
    & = (0, 0, s_0).
  \end{split}
\end{align}
In this way, both a linear feature detector (the Laplacian) and a
non-linear feature detector (the determinant of the Hessian) can be
designed to respond in a scale-selective manner, with their maximum
response over scale at a scale that correspond to the inherent scale
in the input data.

\subsubsection{Edge detection}

Consider next the following idealized model of a diffuse edge
(Lindeberg \citeyear{Lin98-IJCV} Equation~(18))
\begin{equation}
\label{eq-def-ideal-blur-edge-s0}  
  f_{\edge,s_0}(x, y) = \operatorname{erg}(x;\; s_0),
\end{equation}
where $\operatorname{erg}(x;\; s_0)$ denotes the primitive function of
a 1-D Gaussian kernel
\begin{equation}
  \operatorname{erg}(x;\; s_0)
  = \int_{u = - \infty}^x g(u;\; s_0) \, du.
\end{equation}
Following a differential definition of edges, let us measure local
scale-normalized edge
strength by the scale-normalized gradient magnitude
(Lindeberg \citeyear{Lin98-IJCV} Equation~(15))
\begin{equation}
  \label{eq-sc-norm-grad-magn}
  L_{v,\norm} = s^{\gamma/2} \sqrt{L_x^2 + L_y^2},
\end{equation}
which for the scale-space representation of the idealized edge model
(\ref{eq-def-ideal-blur-edge-s0}) leads to a response of the form
\begin{equation}
  L_{v,\norm}(x, y;\; s) = s^{\gamma/2}  \, g(x;\; s_0 + s).
\end{equation}
Then, it can be shown that this scale-normalized edge response will, at
the spatial location of the edge at $x = 0$,
assume its maximum over scale at the scale
\begin{equation}
  \label{eq-grad-magn-sc-sel}
  \hat{s} = \operatorname{argmax}_s L_{v,\norm}(0, 0;\; s)  = s_0,
\end{equation}
provided that we choose the value of the scale normalization power
$\gamma$ as (Lindeberg \citeyear{Lin98-IJCV}, Equation~(23))
\begin{equation}
   \label{eq-gamma-edge-det}
  \gamma_{\edge} = \frac{1}{2}.
\end{equation}

\subsubsection{Ridge detection}

Let us next consider the following idealized model of a ridge
(Lindeberg \citeyear{Lin98-IJCV} Equation~(52))
\begin{equation}
  \label{eq-def-ideal-ridge-s0}  
  f_{\ridge,s_0}(x, y) = g(x;\; s_0).
\end{equation}
For a differential definition of ridges, consider a local coordinate
system $(p, q)$ aligned with the eigendirections of the Hessian
matrix, such that the mixed second-order derivative $L_{pq} = 0$.
Let us measure local scale-normalized ridge strength by the
scale-normalized second-order derivative in the direction $p$
(Lindeberg \citeyear{Lin98-IJCV}, Equations~(42) and~(47)):
\begin{multline}
  \label{eq-sc-norm-ridge-meas}
  L_{pp,\norm}
  = s^{\gamma} L_{pp} = \\
  = s^{\gamma}
      \left(
        L_{xx} + L_{yy} - \sqrt{(L_{xx} - L_{yy})^2 + 4 L_{xy}^2}
      \right),
\end{multline}
which for the idealized ridge model (\ref{eq-def-ideal-ridge-s0})
reduces to the form
\begin{equation}
  L_{pp,\norm}(x, y;\; s) = s^{\gamma} \, L_{xx}(x, y;\; s)
                                  = s^{\gamma} \, g_{xx}(x;\; s_0 + s).
\end{equation}
Then, it can be shown that, at the spatial location of the ridge at
$x = 0$, this scale-normalized ridge response will
assume its maximum over scale at the scale
\begin{equation}
   \label{eq-princ-curv-sc-sel}
  \hat{s} = \operatorname{argmax}_s L_{pp,\norm}(0, 0;\; s) = s_0,
\end{equation}
provided that we choose the value of the scale normalization power
$\gamma$ as (Lindeberg \citeyear{Lin98-IJCV}, Equation~(56))
\begin{equation}
  \label{eq-gamma-ridge-det}
  \gamma_{\ridge} = \frac{3}{4}.
\end{equation}

\subsection{Measures of scale selection performance}
\label{sec-meas-sc-sel-perf}

In the following, we will compare the results of using different ways
of discretizing the Gaussian derivative operators,
when applied
task of performing scale selection for
\begin{itemize}
\item
  Gaussian blobs of the form (\ref{eq-def-2D-gauss-blob-s0}),
\item
  idealized diffuse step edges of the form
  (\ref{eq-def-ideal-blur-edge-s0}), and
\item
  idealized Gaussian ridges of the form (\ref{eq-def-ideal-ridge-s0}).
\end{itemize}
To quantify the performance of the different scale normalization, we
will measure deviations from the ideal results in terms of:
\begin{itemize}
\item
  {\bf Relative scale estimation error:}
  The difference, between a computed scale estimate $\hat{s}$ and the
  ideal scale estimate $\hat{s}_{\scaleref} = s_0$, will be measured by the
  entity
  \begin{equation}
    \label{eq-def-sc-sel-rel-sc-err}
    E_{\scaleestrel}(s) = \sqrt{\frac{\hat{s}}{\hat{s}_{\scaleref}}} - 1.
    \end{equation}
   A motivation for measuring this entity in units of $\sigma =
   \sqrt{s}$ is to have the measurements in dimension of $[\mbox{length}]$.
\end{itemize}
In the ideal continuous case, with the scale-space derivatives
computed from continuous Gaussian derivatives, this error measure
should be zero. Any deviations from zero, when
computed from a discrete implementation based on discrete
approximations of Gaussian derivative kernels, do therefore
characterize the properties of the discretization.

\begin{figure}[hbpt]
  \begin{center}
    \begin{tabular}{c}
    {\em Selected scales for Laplacian scale selection \/} \\
    \includegraphics[width=0.45\textwidth]{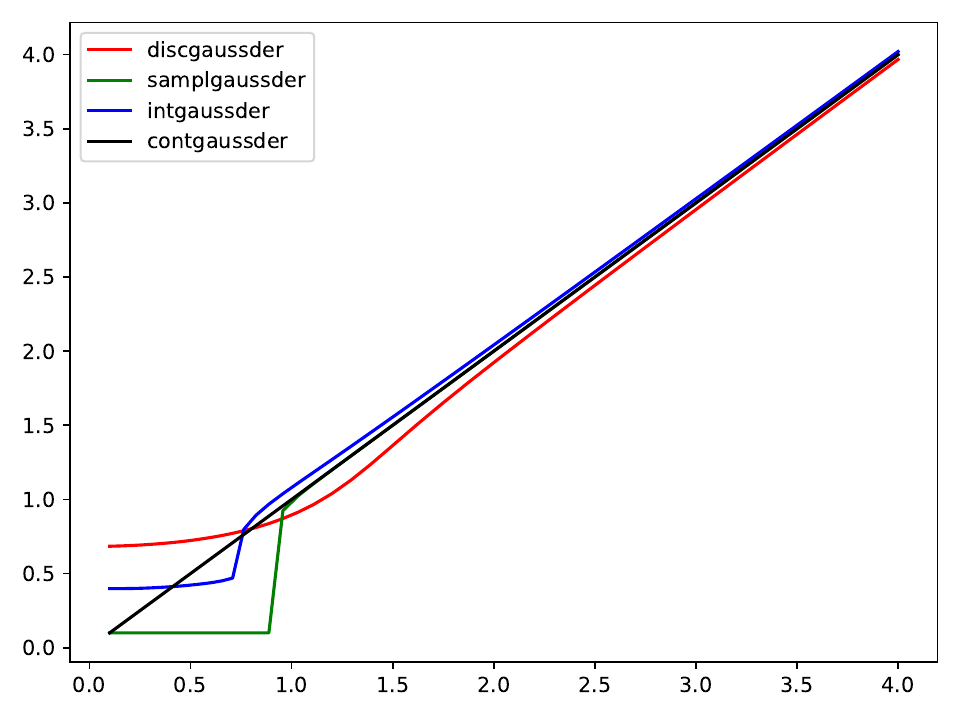}
    \end{tabular}
  \end{center}
  \caption{Graphs of the {\em selected scales\/} $\hat{\sigma} = \sqrt{\hat{s}}$,
    when applying scale selection from local extrema over scale of the
    {\em scale-normalized Laplacian\/} response according to (\ref{eq-lapl-sc-sel})
    to a set of Gaussian blobs of different size $\sigma_{\scaleref} =  \sigma_0$,
    for  different discrete approximations of the Gaussian
    derivative kernels, for either discrete analogues of Gaussian
    derivative kernels $T_{\disc,x^{\alpha}}(n;\; s)$
    according to (\ref{eq-disc-der-gauss}), sampled Gaussian
    derivative kernels $T_{\sampl,x^{\alpha}}(n;\, s)$ according to
    (\ref{eq-sampl-gauss-der}), or integrated Gaussian derivative
    kernels $T_{\intdisc,x^{\alpha}}(n;\, s)$ according to
    (\ref{eq-def-int-gauss-der}). For comparison, the reference scale
    $\sigma_{\scaleref} = \sqrt{s_{\scaleref}} = \sigma_0$ obtained in the continuous case for
    continuous Gaussian derivatives is also shown.
    (Horizontal axis: Reference scale
    $\sigma_{\scaleref} = \sigma_0 \in [0.1, 4]$.)}
  \label{fig-sel-sc-Lapl-scsel-gauss-blob}

  \bigskip

   \begin{center}
    \begin{tabular}{c}
    {\em Relative scale error for Laplacian scale selection\/} \\
    \includegraphics[width=0.45\textwidth]{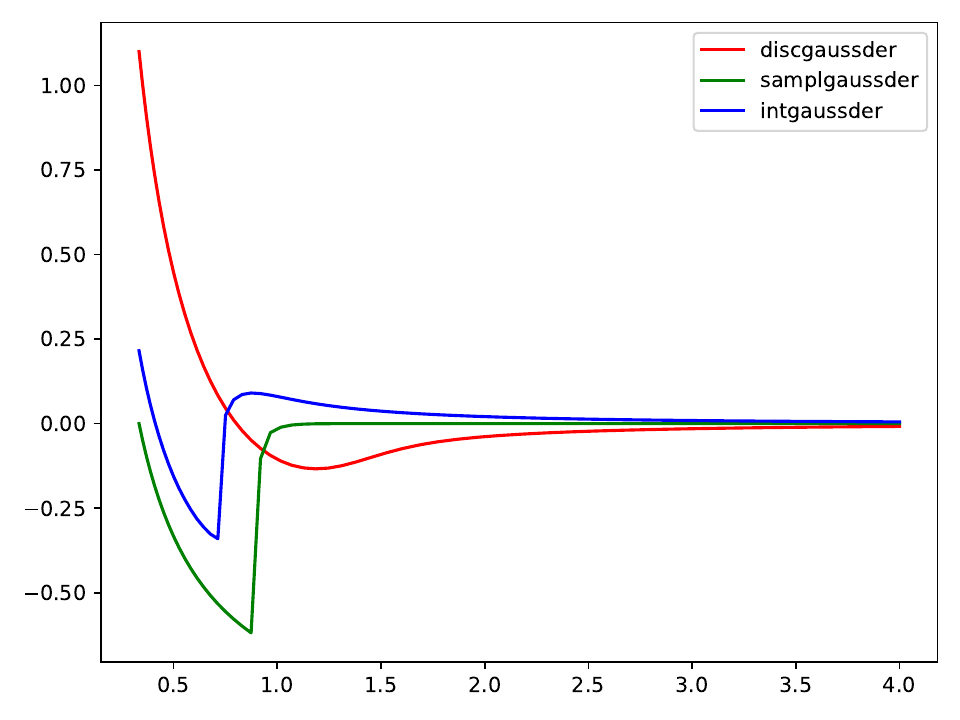}
    \end{tabular}
  \end{center}
  \caption{Graphs of the {\em relative scale estimation error\/}
    $E_{\scaleestrel}(\sigma)$, according to (\ref{eq-def-sc-sel-rel-sc-err}),
    when applying scale selection from local extrema over scale of the
    {\em scale-normalized Laplacian\/} response according to (\ref{eq-lapl-sc-sel})
    to a set of Gaussian blobs of different size $\sigma_{\scaleref} =  \sigma_0$,
    for  different discrete approximations of the Gaussian
    derivative kernels, for either discrete analogues of Gaussian
    derivative kernels $T_{\disc,x^{\alpha}}(n;\; s)$
    according to (\ref{eq-disc-der-gauss}), sampled Gaussian
    derivative kernels $T_{\sampl,x^{\alpha}}(n;\, s)$ according to
    (\ref{eq-sampl-gauss-der}), or integrated Gaussian derivative
    kernels $T_{\intdisc,x^{\alpha}}(n;\, s)$ according to
    (\ref{eq-def-int-gauss-der}). 
    (Horizontal axis: Reference scale
    $\sigma_{\scaleref} = \sigma_0 \in [1/3, 4]$.)}
  \label{fig-rel-sc-err-Lapl-scsel-gauss-blob}
\end{figure}

\begin{figure}[hbpt]
  \begin{center}
    \begin{tabular}{c}
    {\em Selected scales for detHessian scale selection \/} \\
    \includegraphics[width=0.45\textwidth]{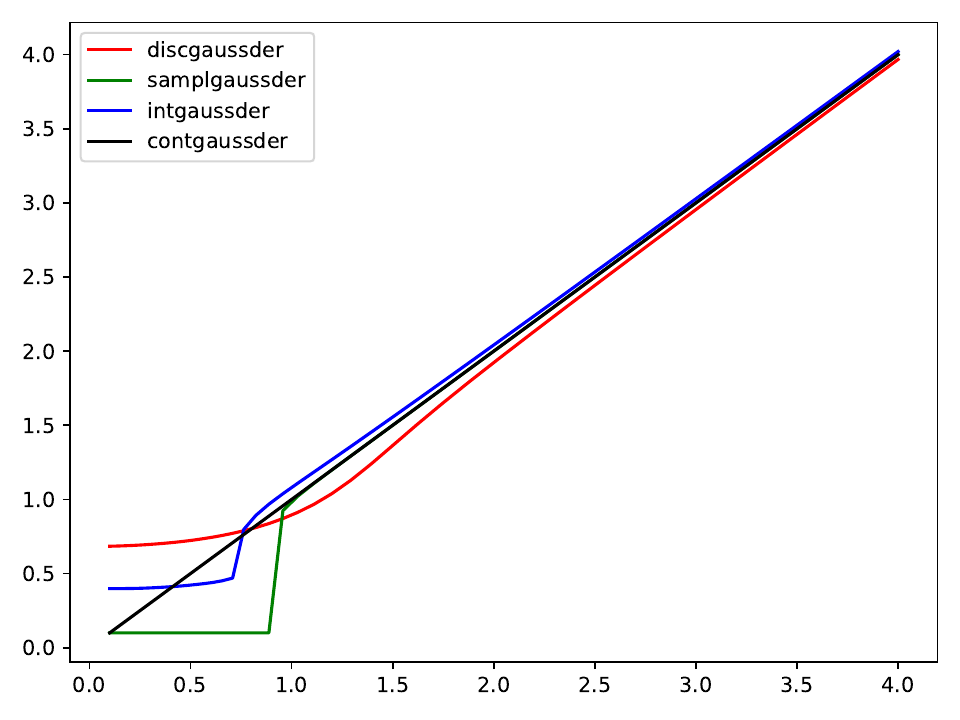}
    \end{tabular}
  \end{center}
  \caption{Graphs of the {\em selected scales\/} $\hat{\sigma} = \sqrt{\hat{s}}$,
    when applying scale selection from local extrema over scale of the
    {\em scale-normalized determinant of Hessian\/} response according to (\ref{eq-det-hess-sc-sel})
    to a set of Gaussian blobs of different size $\sigma_{\scaleref} =  \sigma_0$,
    for  different discrete approximations of the Gaussian
    derivative kernels, for either discrete analogues of Gaussian
    derivative kernels $T_{\disc,x^{\alpha}}(n;\; s)$
    according to (\ref{eq-disc-der-gauss}), sampled Gaussian
    derivative kernels $T_{\sampl,x^{\alpha}}(n;\, s)$ according to
    (\ref{eq-sampl-gauss-der}), or integrated Gaussian derivative
    kernels $T_{\intdisc,x^{\alpha}}(n;\, s)$ according to
    (\ref{eq-def-int-gauss-der}). For comparison, the reference scale
    $\sigma_{\scaleref} = \sqrt{s_{\scaleref}} = \sigma_0$ obtained in the continuous case for
    continuous Gaussian derivatives is also shown.
    (Horizontal axis: Reference scale
    $\sigma_{\scaleref} = \sigma_0 \in [0.1, 4]$.)}
  \label{fig-sel-sc-detHess-scsel-gauss-blob}

  \bigskip

   \begin{center}
    \begin{tabular}{c}
    {\em Relative scale error for detHessian scale selection\/} \\
    \includegraphics[width=0.45\textwidth]{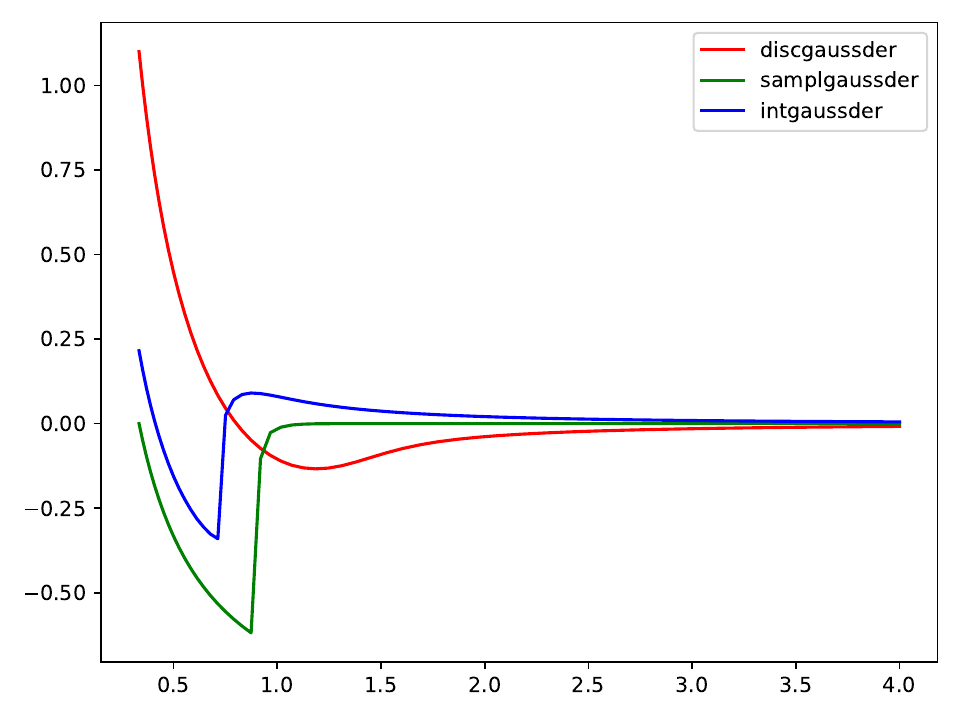}
    \end{tabular}
  \end{center}
  \caption{Graphs of the {\em relative scale estimation error\/}
    $E_{\scaleestrel}(\sigma)$, according to (\ref{eq-def-sc-sel-rel-sc-err}),
    when applying scale selection from local extrema over scale of the
    {\em scale-normalized determinant of the Hessian\/} response according to (\ref{eq-det-hess-sc-sel})
    to a set of Gaussian blobs of different size $\sigma_{\scaleref} =  \sigma_0$,
    for  different discrete approximations of the Gaussian
    derivative kernels, for either discrete analogues of Gaussian
    derivative kernels $T_{\disc,x^{\alpha}}(n;\; s)$
    according to (\ref{eq-disc-der-gauss}), sampled Gaussian
    derivative kernels $T_{\sampl,x^{\alpha}}(n;\, s)$ according to
    (\ref{eq-sampl-gauss-der}), or integrated Gaussian derivative
    kernels $T_{\intdisc,x^{\alpha}}(n;\, s)$ according to
    (\ref{eq-def-int-gauss-der}). 
    (Horizontal axis: Reference scale
    $\sigma_{\scaleref} = \sigma_0 \in [1/3, 4]$.)} 
  \label{fig-rel-sc-err-detHess-scsel-gauss-blob}
\end{figure}

\begin{figure}[hbpt]
  \begin{center}
    \begin{tabular}{c}
    {\em Selected scales for gradient magnitude scale selection \/} \\
    \includegraphics[width=0.45\textwidth]{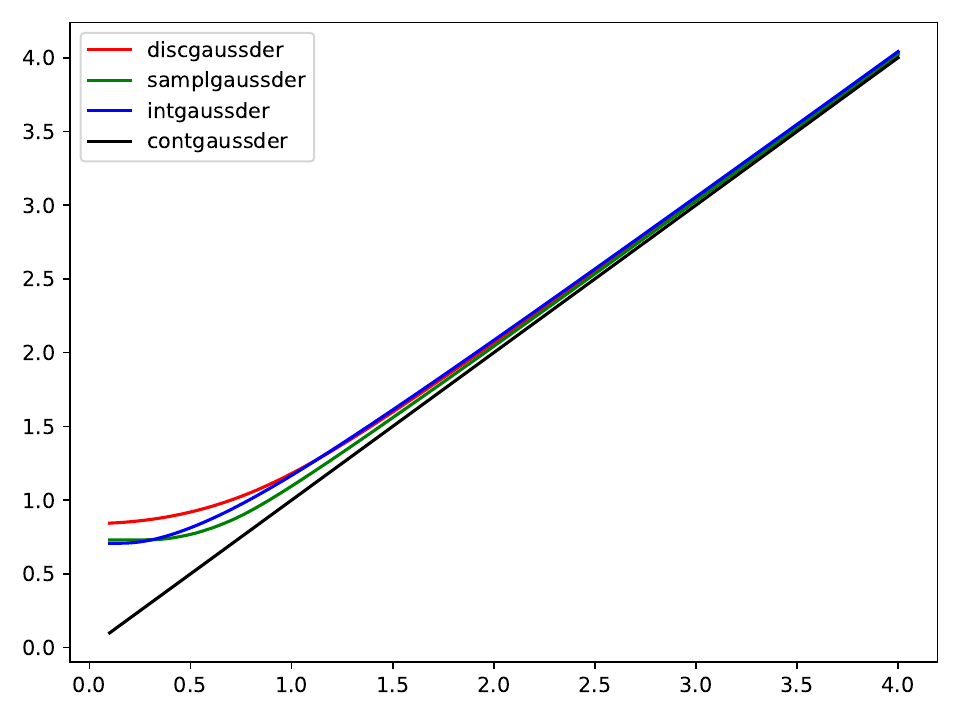}
    \end{tabular}
  \end{center}
  \caption{Graphs of the {\em selected scales\/} $\hat{\sigma} = \sqrt{\hat{s}}$,
    when applying scale selection from local extrema over scale of the
    {\em scale-normalized gradient magnitude\/} response according to (\ref{eq-grad-magn-sc-sel})
    to a set of diffuse step edges of different width $\sigma_{\scaleref} =  \sigma_0$,
    for  different discrete approximations of the Gaussian
    derivative kernels, for either discrete analogues of Gaussian
    derivative kernels $T_{\disc,x^{\alpha}}(n;\; s)$
    according to (\ref{eq-disc-der-gauss}), sampled Gaussian
    derivative kernels $T_{\sampl,x^{\alpha}}(n;\, s)$ according to
    (\ref{eq-sampl-gauss-der}), or integrated Gaussian derivative
    kernels $T_{\intdisc,x^{\alpha}}(n;\, s)$ according to
    (\ref{eq-def-int-gauss-der}). For comparison, the reference scale
    $\sigma_{\scaleref} = \sqrt{s_{\scaleref}} = \sigma_0$ obtained in the continuous case for
    continuous Gaussian derivatives is also shown.
    (Horizontal axis: Reference scale
    $\sigma_{\scaleref} = \sigma_0 \in [0.1, 4]$.)}
  \label{fig-sel-sc-gradmagn-scsel-diff-edge}

  \bigskip

   \begin{center}
    \begin{tabular}{c}
    {\em Relative scale error for gradient magnitude scale selection\/} \\
    \includegraphics[width=0.45\textwidth]{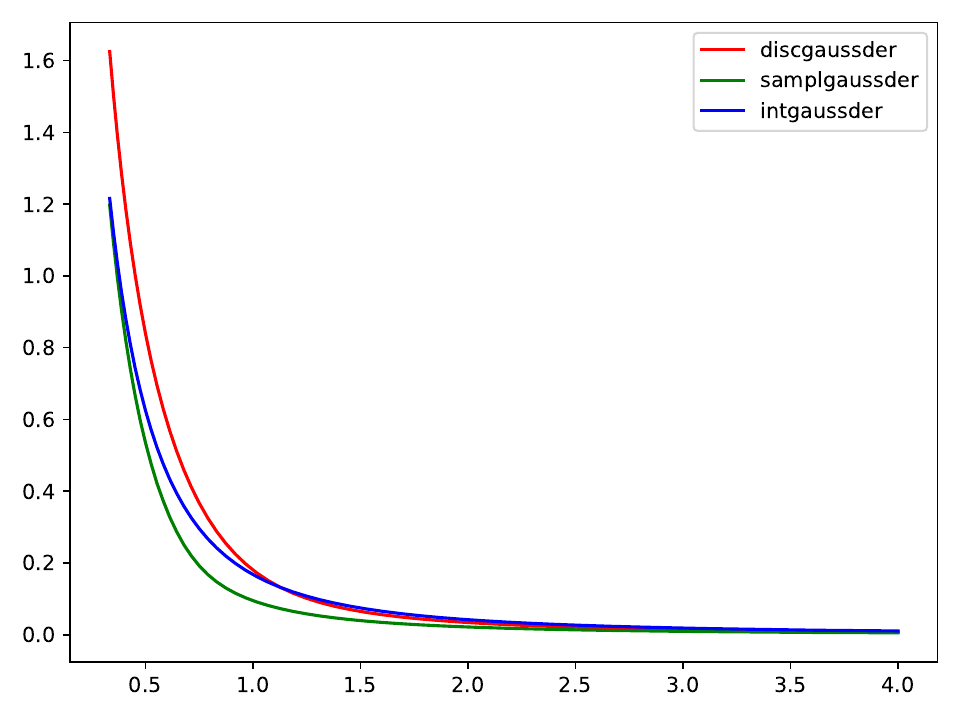}
    \end{tabular}
  \end{center}
  \caption{Graphs of the {\em relative scale estimation error\/}
    $E_{\scaleestrel}(\sigma)$, according to (\ref{eq-def-sc-sel-rel-sc-err}),
    when applying scale selection from local extrema over scale of the
    {\em scale-normalized gradient magnitude\/} response according to (\ref{eq-grad-magn-sc-sel})
    to a set of diffuse step edges of different width $\sigma_{\scaleref} =  \sigma_0$,
    for  different discrete approximations of the Gaussian
    derivative kernels, for either discrete analogues of Gaussian
    derivative kernels $T_{\disc,x^{\alpha}}(n;\; s)$
    according to (\ref{eq-disc-der-gauss}), sampled Gaussian
    derivative kernels $T_{\sampl,x^{\alpha}}(n;\, s)$ according to
    (\ref{eq-sampl-gauss-der}), or integrated Gaussian derivative
    kernels $T_{\intdisc,x^{\alpha}}(n;\, s)$ according to
    (\ref{eq-def-int-gauss-der}). 
    (Horizontal axis: Reference scale
    $\sigma_{\scaleref} = \sigma_0 \in [1/3, 4]$.)} 
  \label{fig-rel-sc-err-gradmagn-scsel-diff-edge}
\end{figure}

\begin{figure}[hbpt]
  \begin{center}
    \begin{tabular}{c}
    {\em Selected scales for principal curvature scale selection \/} \\
    \includegraphics[width=0.45\textwidth]{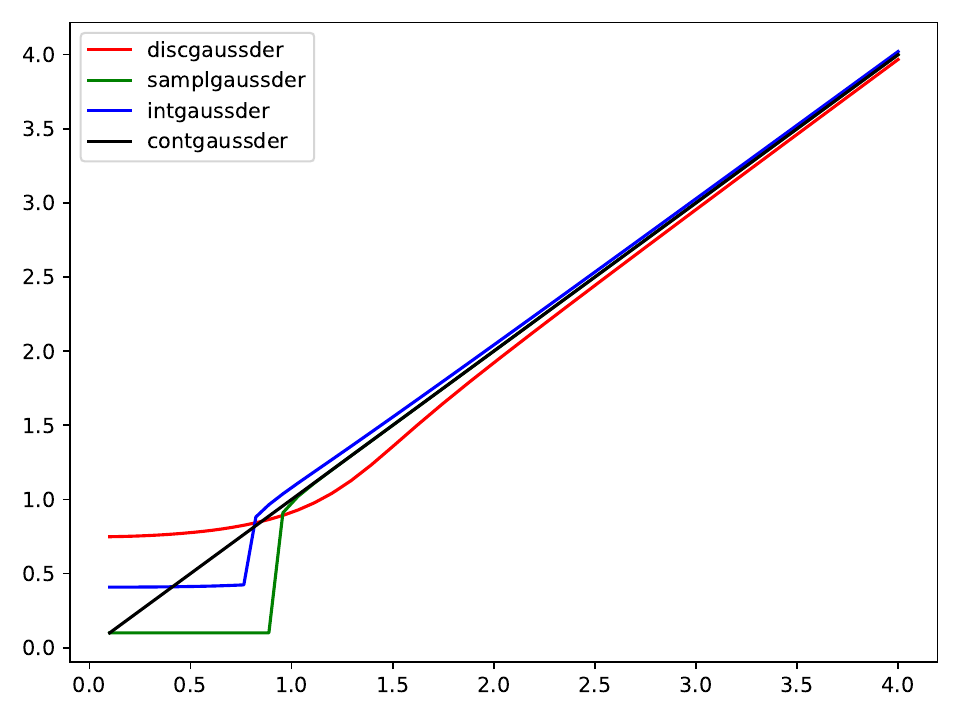}
    \end{tabular}
  \end{center}
  \caption{Graphs of the {\em selected scales\/} $\hat{\sigma} = \sqrt{\hat{s}}$,
    when applying scale selection from local extrema over scale of the
    {\em scale-normalized principal curvature\/} response according to (\ref{eq-princ-curv-sc-sel})
    to a set of diffuse ridges of different width $\sigma_{\scaleref} =  \sigma_0$,
    for  different discrete approximations of the Gaussian
    derivative kernels, for either discrete analogues of Gaussian
    derivative kernels $T_{\disc,x^{\alpha}}(n;\; s)$
    according to (\ref{eq-disc-der-gauss}), sampled Gaussian
    derivative kernels $T_{\sampl,x^{\alpha}}(n;\, s)$ according to
    (\ref{eq-sampl-gauss-der}), or integrated Gaussian derivative
    kernels $T_{\intdisc,x^{\alpha}}(n;\, s)$ according to
    (\ref{eq-def-int-gauss-der}). For comparison, the reference scale
    $\sigma_{\scaleref} = \sqrt{s_{\scaleref}} = \sigma_0$ obtained in the continuous case for
    continuous Gaussian derivatives is also shown.
    (Horizontal axis: Reference scale
    $\sigma_{\scaleref} = \sigma_0 \in [0.1, 4]$.)}
  \label{fig-sel-sc-princcurv-scsel-diff-edge}

  \bigskip

   \begin{center}
    \begin{tabular}{c}
    {\em Relative scale error for principal curvature scale selection\/} \\
    \includegraphics[width=0.45\textwidth]{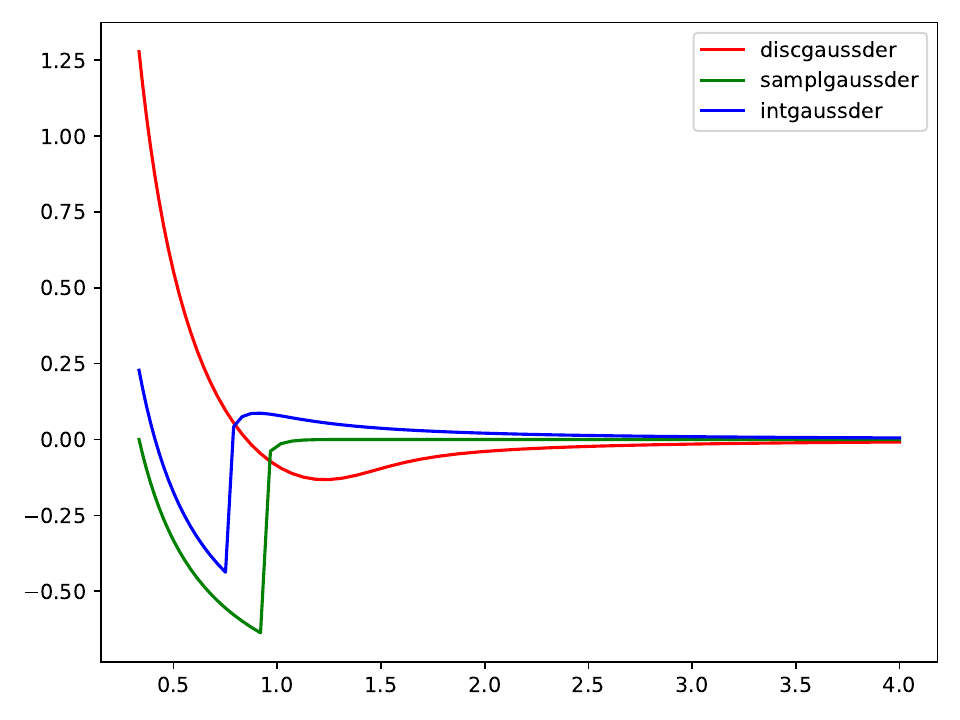}
    \end{tabular}
  \end{center}
  \caption{Graphs of the {\em relative scale estimation error\/}
    $E_{\scaleestrel}(\sigma)$, according to (\ref{eq-def-sc-sel-rel-sc-err}),
    when applying scale selection from local extrema over scale of the
    {\em scale-normalized principal curvature\/} response according to (\ref{eq-princ-curv-sc-sel})
    to a set of diffuse ridges of different width $\sigma_{\scaleref} =  \sigma_0$,
    for  different discrete approximations of the Gaussian
    derivative kernels, for either discrete analogues of Gaussian
    derivative kernels $T_{\disc,x^{\alpha}}(n;\; s)$
    according to (\ref{eq-disc-der-gauss}), sampled Gaussian
    derivative kernels $T_{\sampl,x^{\alpha}}(n;\, s)$ according to
    (\ref{eq-sampl-gauss-der}), or integrated Gaussian derivative
    kernels $T_{\intdisc,x^{\alpha}}(n;\, s)$ according to
    (\ref{eq-def-int-gauss-der}). 
    (Horizontal axis: Reference scale
    $\sigma_{\scaleref} = \sigma_0 \in [1/3, 4]$.)} 
  \label{fig-rel-sc-err-princcurv-scsel-diff-edge}
\end{figure}

\subsection{Numerical quantification of deviations from theoretical
  properties resulting from different discretizations of
  scale-normalized derivatives}
\label{sec-perf-meas-sc-norm-ders}

Our experimental investigation will focus on computing the relative scale estimation error for:
\begin{itemize}
\item
  scale selection based on the scale-normalized Laplacian operator (\ref{eq-sc-norm-lapl})
  for scale normalization power $\gamma = 1$, applied to an ideal
  Gaussian blob of the form (\ref{eq-def-2D-gauss-blob-s0}),
  \item
  scale selection based on the scale-normalized determinant of the
  Hessian operator (\ref{eq-sc-norm-det-hess})
  for scale normalization power $\gamma = 1$, applied to an ideal
  Gaussian blob of the form (\ref{eq-def-2D-gauss-blob-s0}),
\item
  scale selection based on the scale-normalized gradient magnitude
  (\ref{eq-sc-norm-grad-magn}) for scale normalization power
  $\gamma = 1/2$, applied to an ideal diffuse edge of the
  form (\ref{eq-def-ideal-blur-edge-s0}), and
\item
  scale selection based on the scale-normalized ridge strength measure
  (\ref{eq-sc-norm-ridge-meas}) for scale normalization power
  $\gamma = 3/4$, applied to an ideal Gaussian ridge of the form
  (\ref{eq-def-ideal-ridge-s0}).
\end{itemize}
With the given calibration of the scale normalization powers $\gamma$
to the specific feature detection tasks, the estimated scale level
$\hat{s}$ will in the ideal continuous case correspond to the scale
estimate reflecting the inherent scale of the feature model
\begin{equation}
  \hat{s}_{\scaleref} = s_0,
\end{equation}
for all of the cases of ideal Gaussian blobs, ideal diffuse edges or
ideal Gaussian ridges.
This bears relationships to the matched filter theorem
(Woodward \citeyear{Woo53-book}, Turin \citeyear{Tur60-matchedfilter}), in that the
scale selection mechanism will choose filters for detecting the
different types of image structures image data, that as well as possible match their size.

\subsubsection{Experimental methodology}

The experimental procedure, that we will follow in the following
experiments, consists of:
\begin{enumerate}
\item
  For a dense set of 50 logarithmically distributed scale levels $\sigma_{\refscale,i} = A_1 \, r_1^i$
  within the range $\sigma \in [0.1, 4.0]$, where $r_1 > 1$,
  generate an ideal model signal (a Gaussian blob, a diffuse step edge
  or a diffuse ridge) with scale parameter $\sigma_{\refscale,i}$, that represents the
  size in dimension $[\mbox{length}]$.
\item
  For a dense%
\footnote{In these experiments, we use a much denser distribution of
  the scale levels than would be used in an actual feature detection
  detection algorithm with automatic scale selection. The motivation
  for using such a dense distribution here, is to obtain accurate
  performance curves, that are not limited in accuracy because of a
  sparse scale sampling, but instead reflect the inherent properties
  of the underlying computational mechanisms.}
  set of 80 logarithmically distributed scale levels $\sigma_{\acc,j} = A_2 \, r_2^j$,
  within the range $\sigma \in [0.1, 6.0]$, where $r_2 > 1$,
  compute the scale-space signature, that is compute
  the scale-normalized response of the differential entity
  ${\cal D}_{norm} L$ at all scales $\sigma_{\acc,j}$.
\item
  Detect the local extrema over scale of the appropriate polarity (minima for
  Laplacian and principal curvature scale selection, and maxima for
  determinant of the Hessian and gradient magnitude scale selection)
  and select the local extremum that is closest to
  $\sigma_{\refscale,i}$.%
\footnote{In practice, this selection of a closest point should not be regarded
  as that important, since in most of the cases, there will be a dominant local
  extremum of the appropriate type. This selection is, however, a
  precaution, in case there would be multiple local extrema.}

  If there is no local extremum of the right polarity, include the
  boundary extrema into the analysis, and do then select the global extremum out of
  these.
\item
  Interpolate the scale value of the extremum to higher accuracy than
  grid spacing, by for each interior extremum fitting a second-order
  polynomial to the values at the central point and the values of the two adjacent neighbours.
  Find the extremum of the continuous polynomial, and let the scale
  value of the extremum of that interpolation polynomial be the scale
  estimate.%
\footnote{This is a similar scale refinement mechanism as used in the
  fully integrated algorithms for automatic scale selection in
  (Lindeberg \citeyear{Lin97-IJCV,Lin98-IJCV}).}
\end{enumerate}
Figures~\ref{fig-sel-sc-Lapl-scsel-gauss-blob}--\ref{fig-rel-sc-err-princcurv-scsel-diff-edge}
show graphs of scale estimates with associated relative scale errors
obtained in this way.

\subsubsection{Scale selection with the scale-normalized Laplacian
  applied to Gaussian blobs}

From Figure~\ref{fig-sel-sc-Lapl-scsel-gauss-blob}, which shows the
scale estimates obtained by detecting local extrema over scale of the scale-normalized
Laplacian operator, when applied to Gaussian blobs of different size,  we see that for all
the three approximation methods for Gaussian derivatives; discrete
analogues of Gaussian derivatives, sampled Gaussian derivatives and
integrated Gaussian derivatives, the scale estimates approach the
ideal values of the fully continuous model with increasing size of the
Gaussian blob used as input for the analysis.

For smaller scale values, there are, however, substantial deviations
between the different methods. When the scale parameter $\sigma$ is less than about 1, the results
obtained from sampled Gaussian derivatives fail to generate interior local
extrema over scale. Then, the extremum detection method instead resorts to
returning the minimum scale of the scale interval, implying
qualitatively substantially erroneous scale estimates.
For the integrated Gaussian derivatives, there is also a discontinuity
in the scale selection curve, while at a lower scale level, and not
leading to as low scale values as for the sampled Gaussian
derivatives.

For the discrete analogue of Gaussian derivatives, the behaviour is,
on the other hand, qualitatively different. Since these derivative
approximation kernels tend to regular central difference operators,
as the scale parameter tends to zero, their
magnitude is bounded from above in a completely different way than for
the sampled or integrated Gaussian derivatives. When this bounded
derivative response is multiplied by the scale parameter raised to the
given power, the scale-normalized feature strength measure cannot
assume as high values at very finest scales, as for the sampled or integrated Gaussian
derivatives. This means that the extremum over scale will be assumed
at a relatively coarser scale, when the reference scale is small,
compared to the cases for
the sampled or the integrated Gaussian kernels.

From Figure~\ref{fig-rel-sc-err-Lapl-scsel-gauss-blob}, which shows
the relative scale estimation error $E_{\scaleestrel}(\sigma)$
according to (\ref{eq-def-sc-sel-rel-sc-err}), we can see that
when the reference scale becomes larger, the scale estimates obtained
with the discrete analogues of Gaussian derivatives do, on the other
hand, lead to underestimates of the scale levels, whereas the scale
estimates obtained with integrated Gaussian kernels lead to
overestimates of the scale levels. For $\sigma_{\scaleref}$ a bit greater
than 1, the sampled Gaussian derivatives lead to the most accurate
scale estimates for Laplacian blob detection applied to Gaussian blobs.

\subsubsection{Scale selection with the scale-normalized determinant
  of the Hessian applied to Gaussian blobs}

For
Figures~\ref{fig-sel-sc-detHess-scsel-gauss-blob}--\ref{fig-rel-sc-err-detHess-scsel-gauss-blob},
which show corresponding results for determinant of the Hessian scale
selection applied to Gaussian blobs, the results are similar to the
results for Laplacian scale selection. These results are, however,
nevertheless reported here, to emphasize that the scale selection
method does not only apply to feature detectors that are linear in the
dependency of the Gaussian derivatives, but also to feature detectors
that correspond to genuinely non-linear
combinations of Gaussian derivative responses.

\subsubsection{Scale selection with the scale-normalized gradient
  magnitude applied to diffuse step edges}

From Figure~\ref{fig-sel-sc-gradmagn-scsel-diff-edge}, which shows the
selected scales obtained by detecting local extrema over scale of the
scale-normalized gradient magnitude applied to diffuse step edges of
different width, we can note that all the three discretization methods
for Gaussian derivatives are here bounded from below by an inner
scale. The reason why the behaviour is qualitatively different, in this
case based on first-order derivatives, compared to the previous case
with second-order derivatives, is that the magnitudes of the first-order
derivative responses are in all these cases bounded from above.
The lower bound on the scale estimates is, however, slightly higher
for the discrete analogues of the Gaussian derivatives compared to the
sampled or integrated Gaussian derivatives.

From Figure~\ref{fig-rel-sc-err-gradmagn-scsel-diff-edge}, we can also
see that the sampled Gaussian derivatives lead to slightly more accurate
scale estimates than the integrated Gaussian derivatives or the discrete
analogues of Gaussian derivatives, over the entire scale range.

\subsubsection{Scale selection with the second-order principal curvature
  measure applied to diffuse ridges}

From Figure~\ref{fig-sel-sc-princcurv-scsel-diff-edge}, which shows
the selected scales obtained by detecting local extrema over scale of the
scale-normalized principal curvature response according to
(\ref{eq-princ-curv-sc-sel}), when applied to a set of diffuse ridges
of different width, we can note that the behaviour is qualitatively
very similar to the previously treated second-order methods for scale
selection, based on extrema over scale of either the scale-normalized Laplacian or
the scale-normalized determinant of the Hessian.

There are clear discontinuities in the scale estimates obtained from
sampled or integrated Gaussian derivatives, when the reference scale
$\sigma_{\scaleref}$ goes down towards $\sigma = 1$, at slightly lower scale
values for integrated Gaussian derivatives compared to the sampled
Gaussian derivatives. For the discrete analogues of Gaussian
derivatives, the scale estimates are bounded from below at the finest
scales, whereas there are underestimates in the scale values 
near above $\sigma = 1$. Again, for scale values above about 1, the
sampled Gaussian derivatives lead to results that are closest to
those obtained in the ideal continuous case.

\subsection{Summary of the evaluation on scale selection experiments}

To summarize the results from this investigation, the sampled Gaussian
derivatives lead to the most accurate scale estimates, when the
reference scale $\sigma_{\scaleref}$ of the image features is somewhat above
1. For lower values of the reference scale, the behaviour is, on the
other hand, qualitatively different for the scale selection methods
that are based on second-order derivatives, or what could be expected
more generally, derivatives of even order. When the scale parameter
tends to zero, the strong influence from the fact that the derivatives
of even order of the continuous
Gaussian kernel tend to infinity at the origin, when the scale parameter
tends to zero, implies that the scale selection methods based on
either sampled or integrated Gaussian derivatives lead to
singularities when the reference scale is sufficiently low
(below 1 for the second-order derivatives in the above experiment).

If aiming at handling image data with discrete
approximations of Gaussian derivatives of even order for very low
scale values, it seems natural to then consider alternative discretization
approaches, such as the discrete analogues of Gaussian derivatives.
The specific lower bound on the scale values may, however, be
strongly dependent upon what tasks the Gaussian derivative responses are
to be used for, and also upon the order differentiation.

\begin{figure*}[hbtp]
  \begin{center}
    {\em The affine Gaussian kernel with its directional
      derivatives up to order 4}
  \end{center}
  \begin{center}
    \begin{tabular}{c}
      \hspace{-8mm} \includegraphics[width=0.23\textwidth]{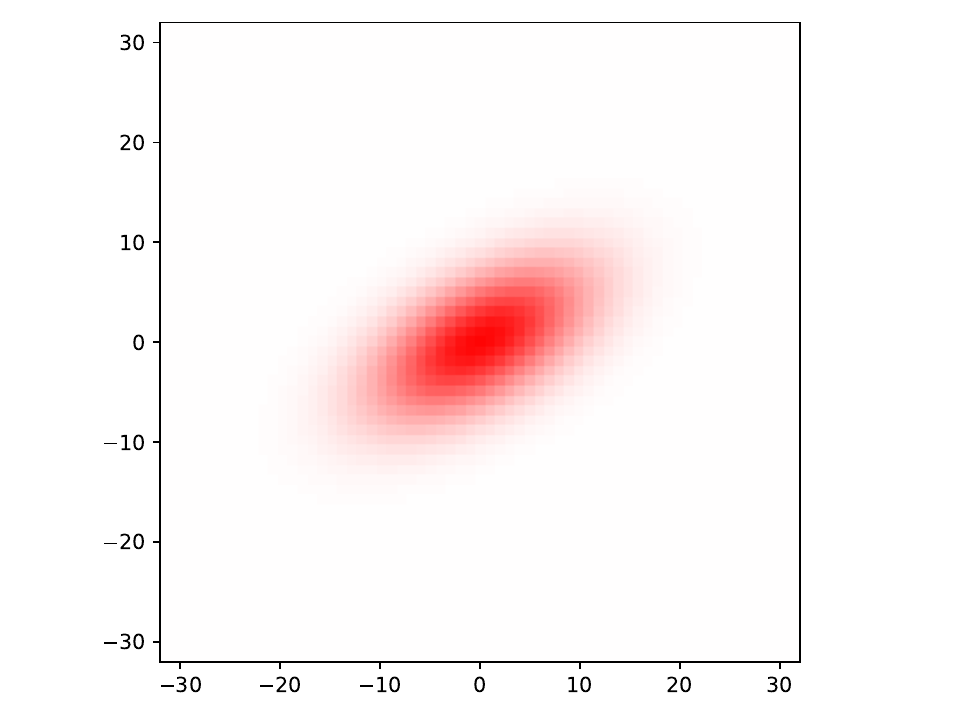}
    \end{tabular}
  \end{center}
  \begin{center}
    \begin{tabular}{cc}
      \hspace{-8mm} \includegraphics[width=0.23\textwidth]{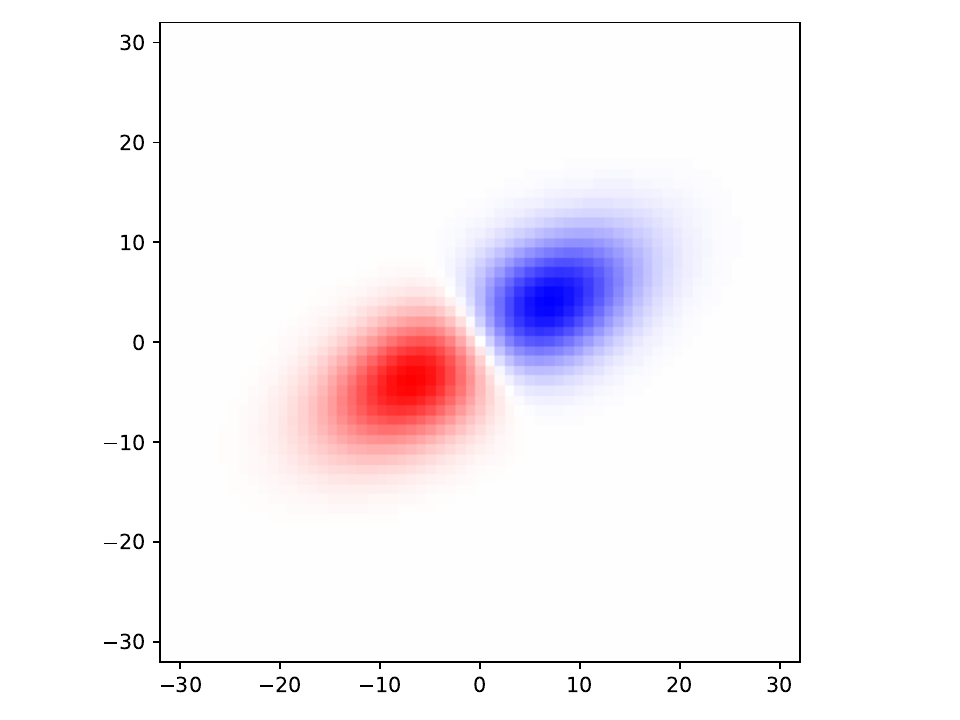} \hspace{-10mm}
      & \includegraphics[width=0.23\textwidth]{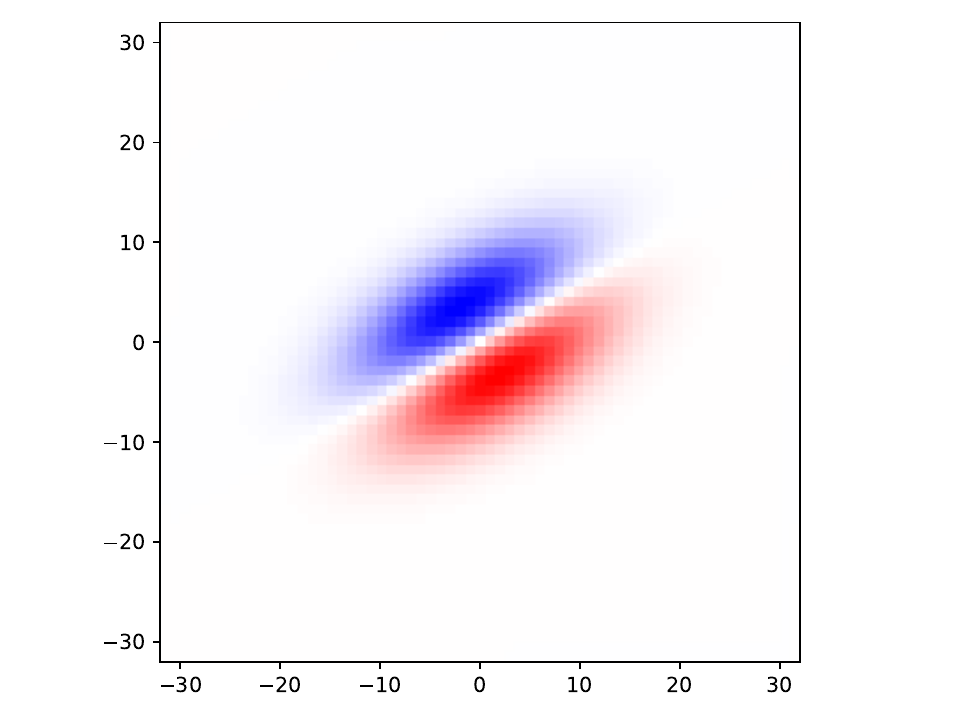}
    \end{tabular}
  \end{center}
  \begin{center}
    \begin{tabular}{ccc}
      \hspace{-8mm} \includegraphics[width=0.23\textwidth]{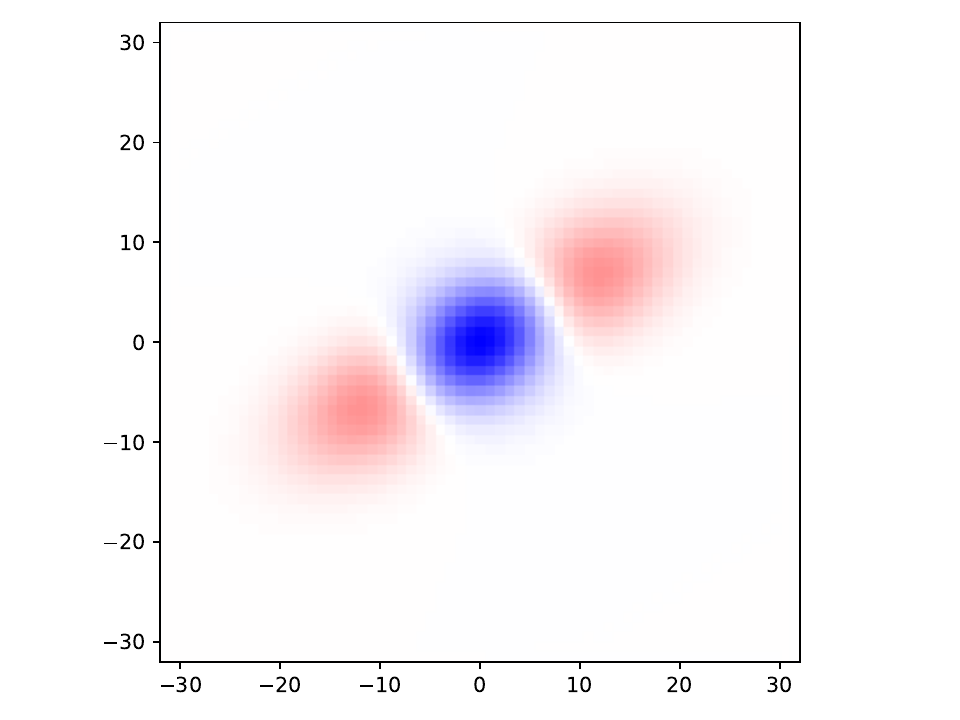} \hspace{-10mm}
      & \includegraphics[width=0.23\textwidth]{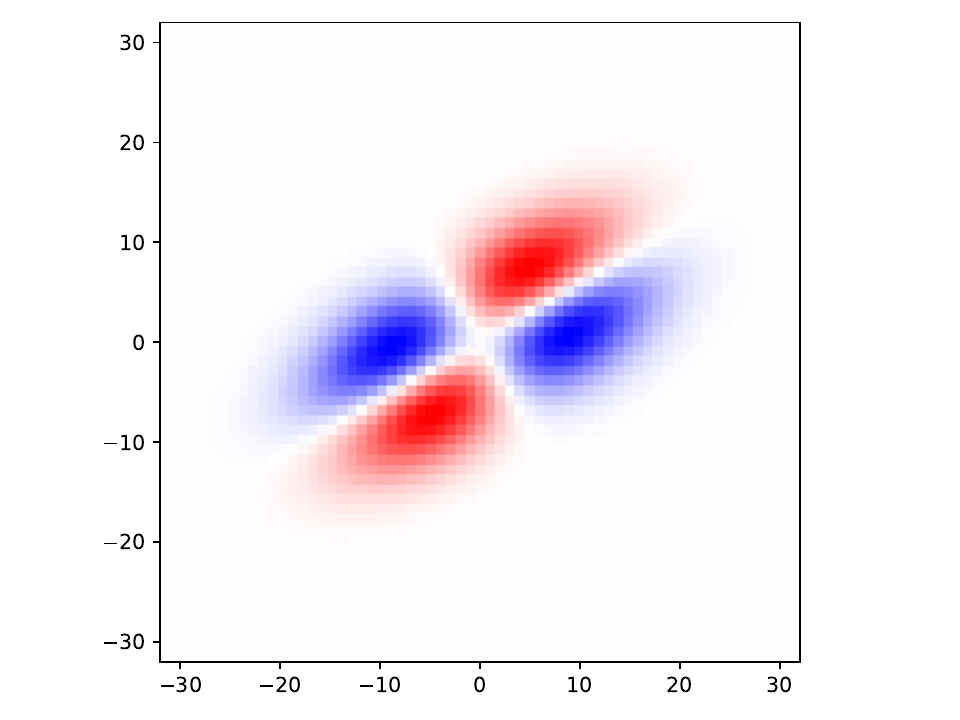} \hspace{-10mm}
       & \includegraphics[width=0.23\textwidth]{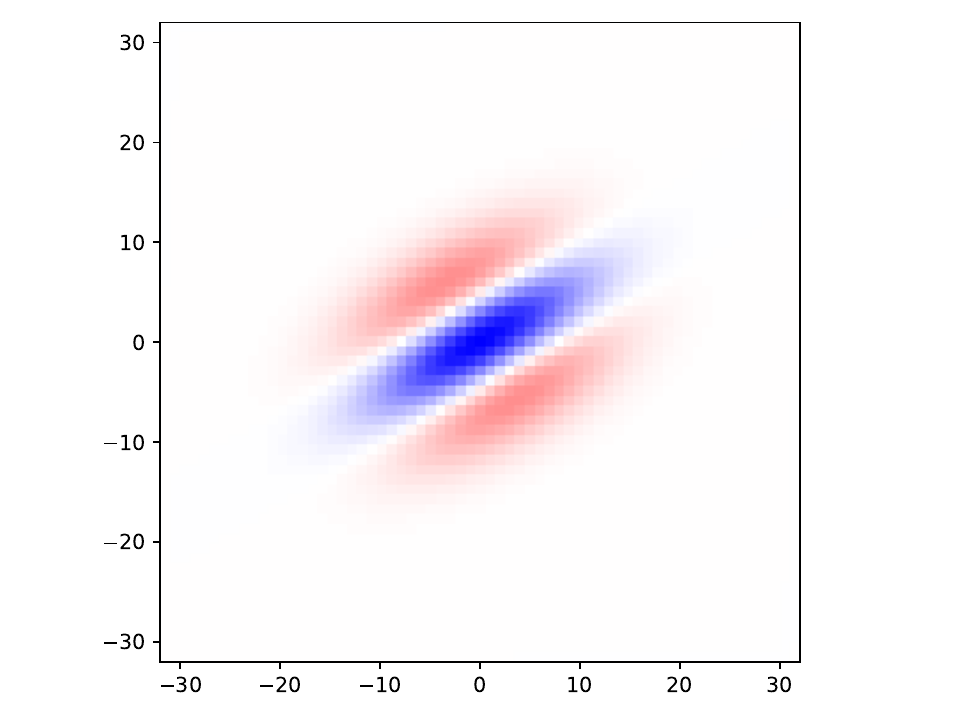}
    \end{tabular}
  \end{center}
  \begin{center}
    \begin{tabular}{cccc}
      \hspace{-8mm} \includegraphics[width=0.23\textwidth]{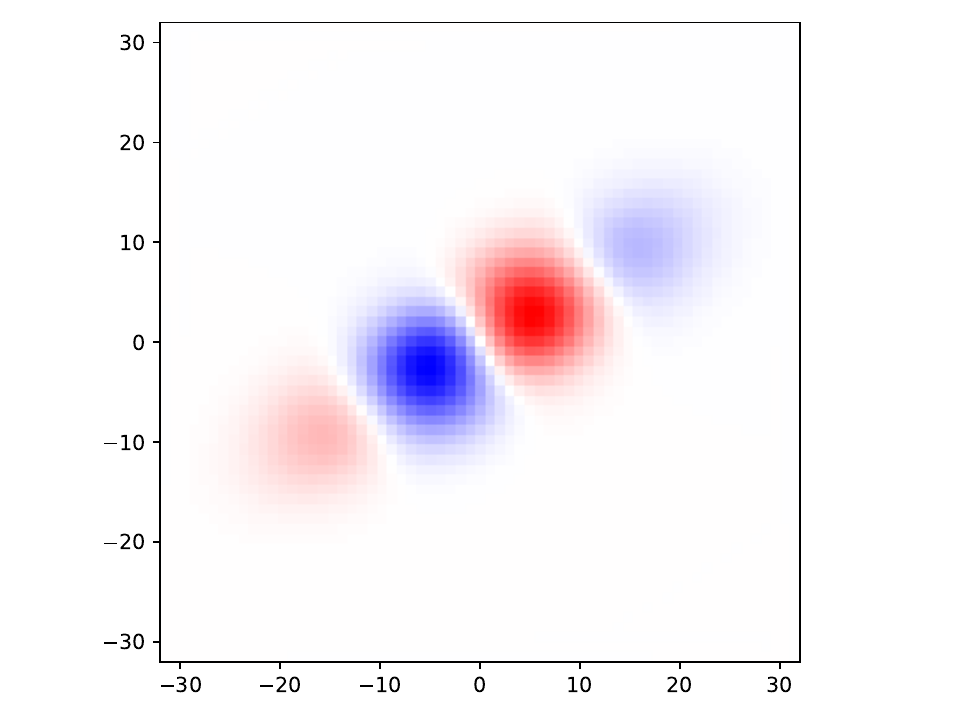} \hspace{-10mm}
      & \includegraphics[width=0.23\textwidth]{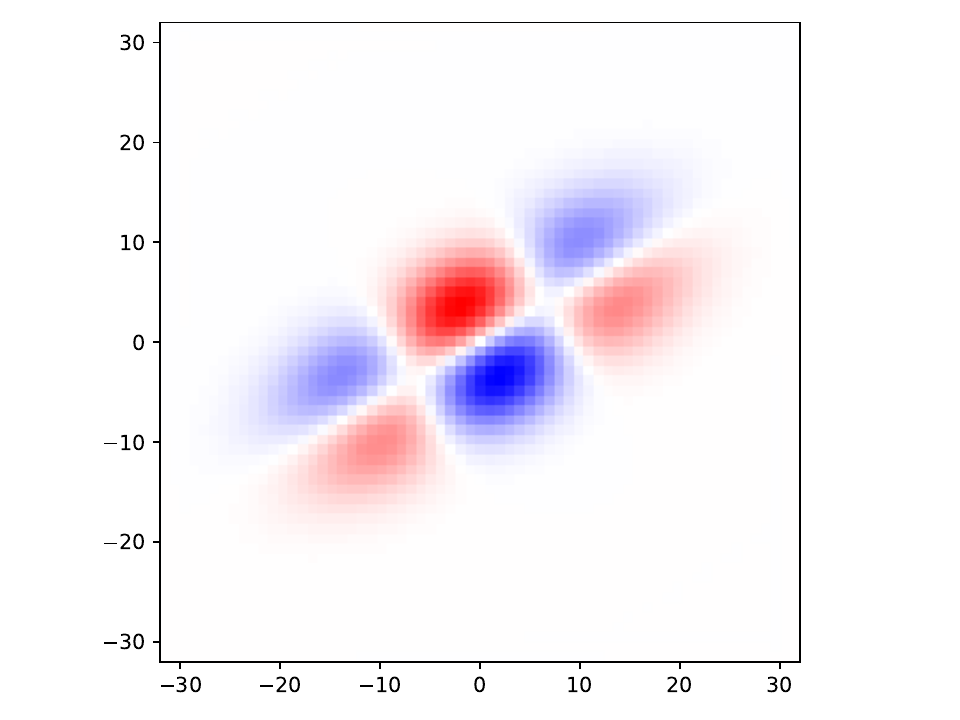} \hspace{-10mm}
       & \includegraphics[width=0.23\textwidth]{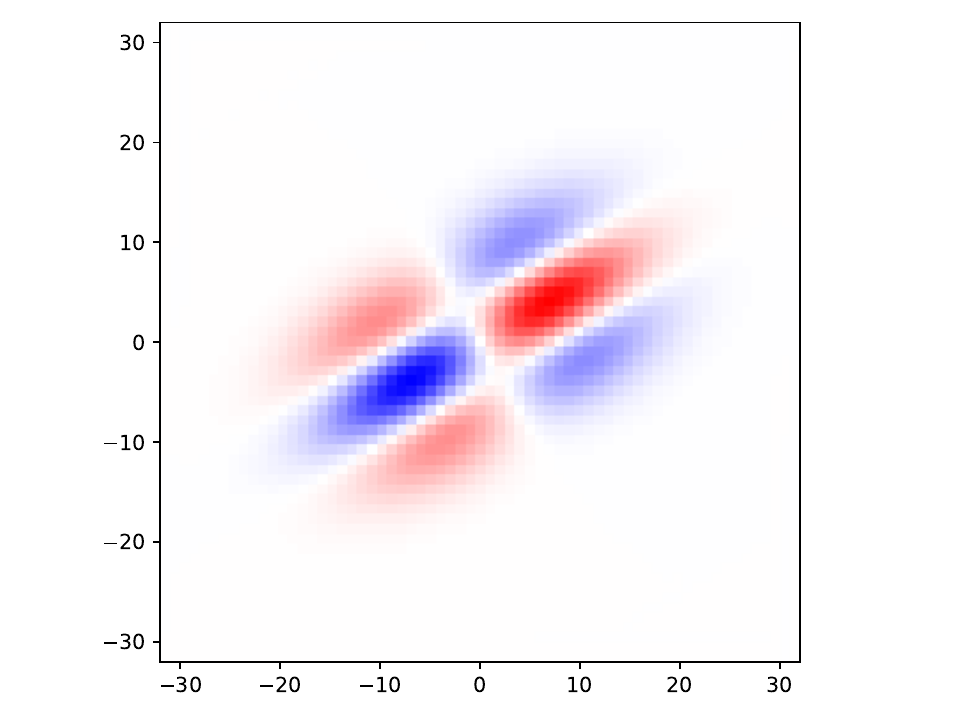} \hspace{-10mm}
       & \includegraphics[width=0.23\textwidth]{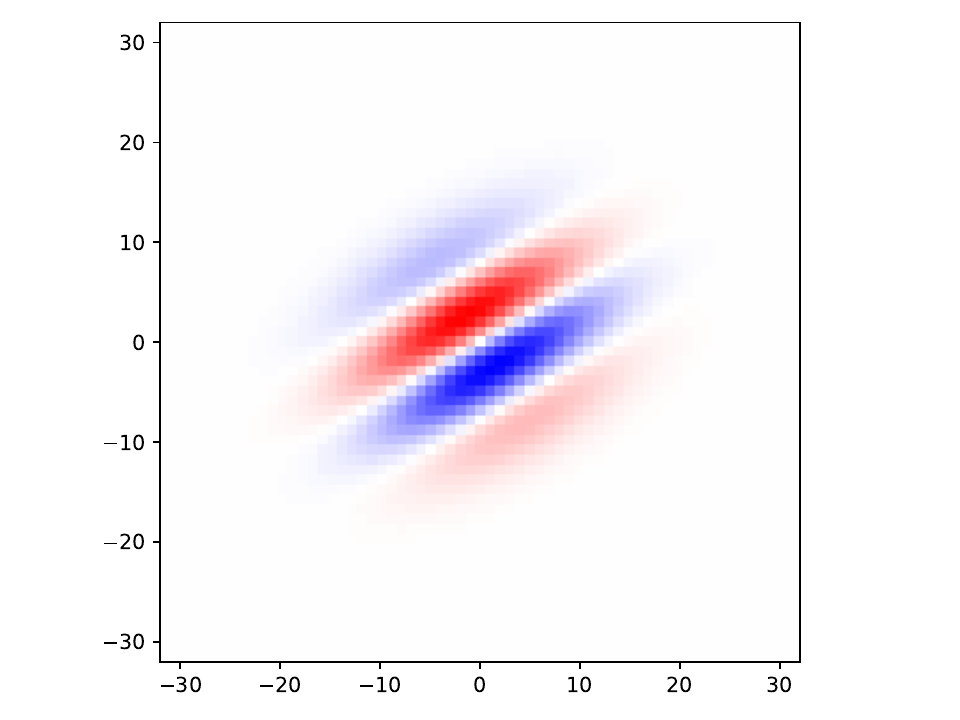}
    \end{tabular}
  \end{center}
  \begin{center}
     \begin{tabular}{ccccc}
      \hspace{-8mm} \includegraphics[width=0.23\textwidth]{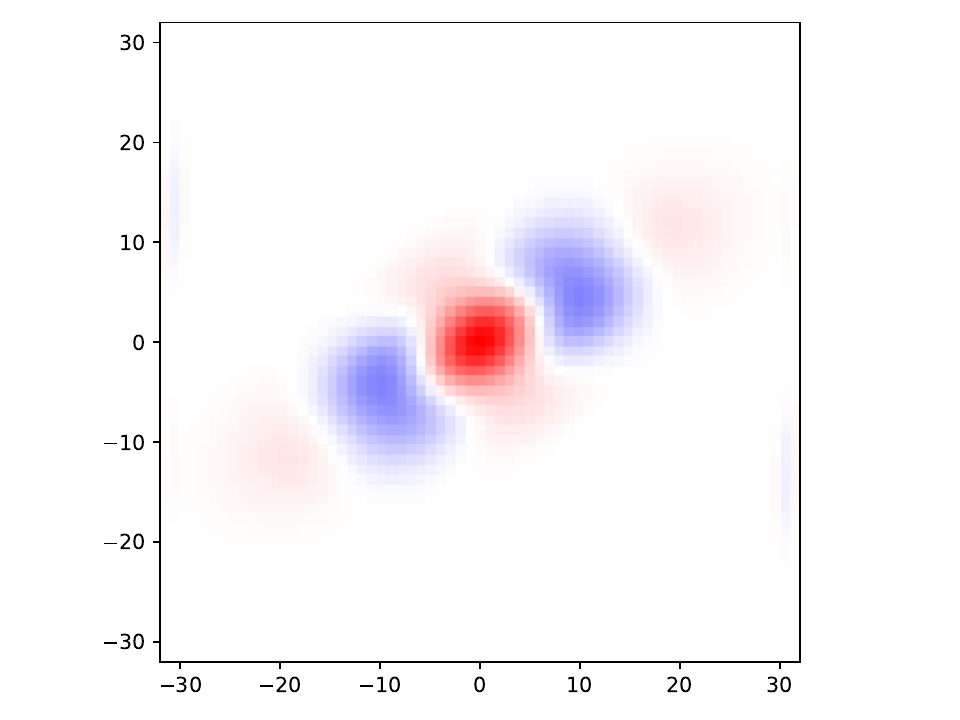} \hspace{-10mm}
        & \includegraphics[width=0.23\textwidth]{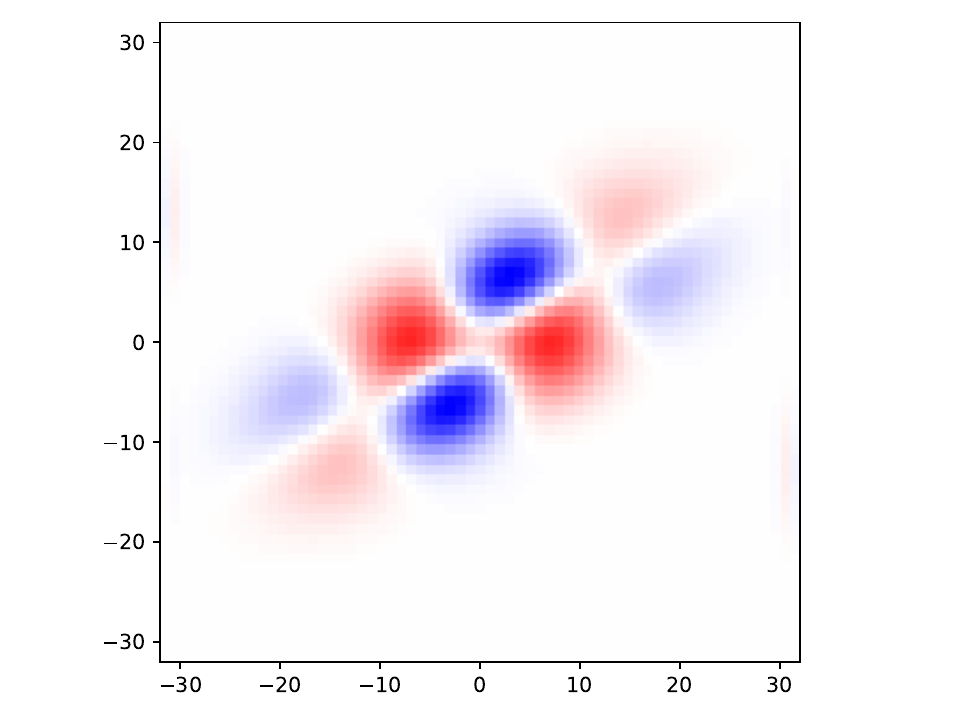} \hspace{-10mm}
       & \includegraphics[width=0.23\textwidth]{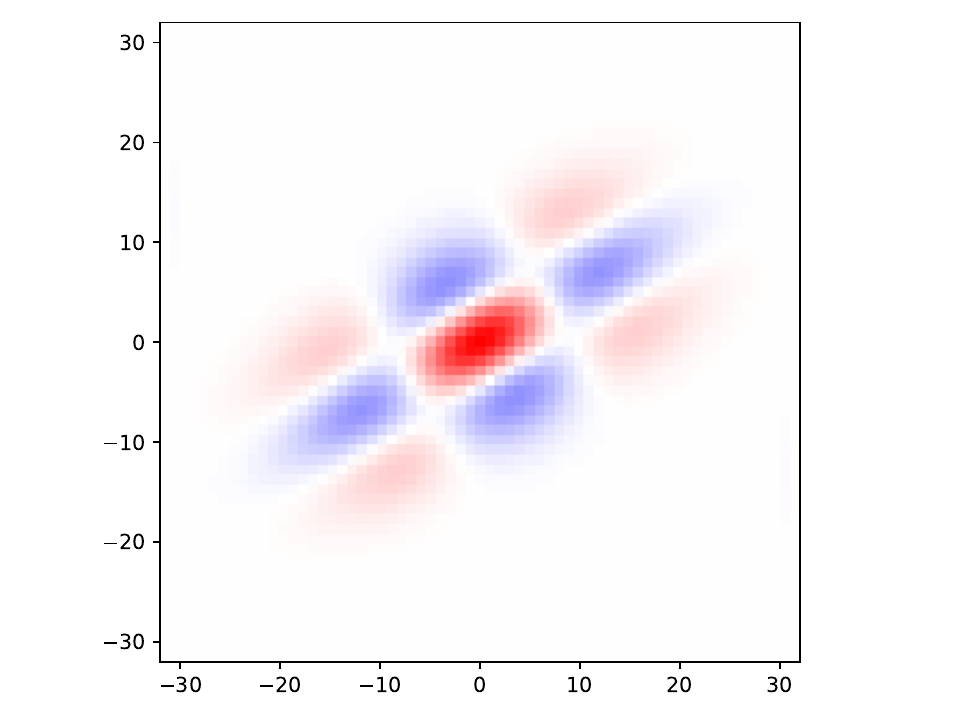} \hspace{-10mm}
       & \includegraphics[width=0.23\textwidth]{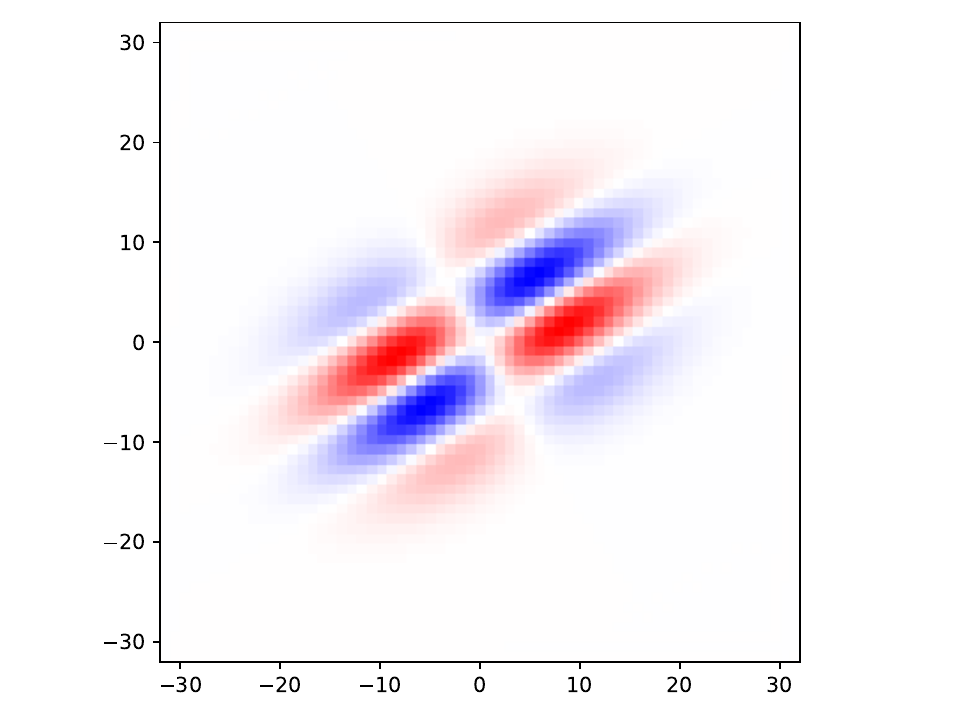} \hspace{-10mm}
       & \includegraphics[width=0.23\textwidth]{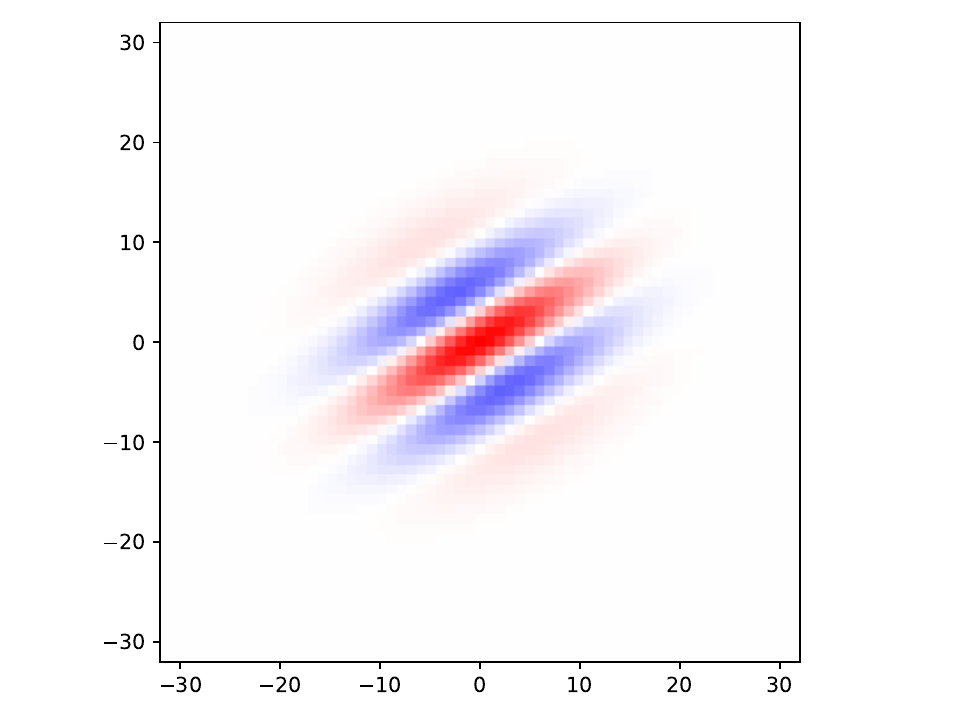}
    \end{tabular}
  \end{center}
  \caption{The affine Gaussian kernel for the spatial scale parameters
    $\sigma_1 = 8$ and $\sigma_2 = 4$
    and image orientation $\varphi = \pi/6$, with its 
    directional derivatives up to order 4, computed by applying small
    support directional derivative approximation masks of the form
    (\ref {eq-dir-der-mask-weight-expr}) to a sampled
    affine Gaussian kernel according to
    (\ref{eq-sampl-aff-gauss-kern}), based the explicit
    parameterization of affine Gaussian kernels according
    to Appendix~\ref{sec-app-aff-gauss-param-kern}.
    (Horizontal axes: $x$-coordinate $\in [-32, 32]$.
     Vertical axes: $y$-coordinate $\in [-32, 32]$.
     Colour coding: positive values in red, negative values in blue.)}
  \label{fig-aff-gauss-kernels}
\end{figure*}

\section{Discrete approximations of directional derivatives}
\label{disc-approx-dir-ders}

When operating on a 2-D spatial scale-space representation, generated
by convolutions with either rotationally symmetric Gaussian
kernels according to (\ref{eq-2D-gauss-conv}) or affine Gaussian
kernels,%
\footnote{See Appendix~\ref{sec-app-aff-gauss-scsp} for a description
  of the affine Gaussian scale-space concept that underlies the
  definition of affine Gaussian kernels.}
it is often
desirable to compute image features in terms of local directional
derivatives.

Given an image orientation $\varphi$ and its ortogonal direction
$\bot\varphi = \varphi + \pi/2$, we can express directional derivatives
along these directions in terms of partial derivative operators
$\partial_x$ and $\partial_y$ along the $x$- and
$y$-directions, respectively, according to
\begin{align}
  \begin{split}
    \label{eq-dphi-def}
    \partial_{\varphi}
    = \cos \varphi \, \partial_x + \sin \varphi \, \partial_y,
  \end{split}\\
  \begin{split}
    \label{eq-dorth-def}    
    \partial_{\bot\varphi}
    = - \sin \varphi \, \partial_x + \cos \varphi \, \partial_y.
  \end{split}
\end{align}
Higher-order directional derivatives of the scale-space representation
can then be defined according to
\begin{equation}
  \label{eq-dir-der-scsp-cont}
  L_{\varphi^{m_1} \bot\varphi^{m_2}} 
  = \partial_{\varphi}^{m_1} \, \partial_{\bot\varphi}^{m_2} \, L,
\end{equation}
where $L$ here denotes either a scale-space representation based on
convolution with a rotationally symmetric Gaussian kernel according to
(\ref{eq-2D-gauss-conv}), or convolution with an affine Gaussian kernel.

Image representations of this form are useful for modelling filter
bank approaches, for either purposes in classical computer vision or
in deep learning. It has also been demonstrated that the spatial
components of the receptive fields of simple cells in the primary
visual cortex of higher mammals can be modelled qualitatively
reasonably well in terms of such directional derivatives combined with
spatial smoothing using affine Gaussian kernels, then with the orientations
$\varphi$ and $\bot\varphi$ of the directional derivatives parallel to
the orientations corresponding to the eigendirections of the affine
spatial covariance matrix $\Sigma$, that underlies the definition of
affine Gaussian kernels
(see Equation~(23) in Lindeberg (\citeyear{Lin21-Heliyon})).

\subsection{Small-support directional derivative approximation masks}

If attempting to compute filter bank responses in terms of directional
derivatives in different image directions, and of different orders of
spatial differentiation, the amount of computational work will, however, grow
basically linearly with the number of different combinations of the
orders $m_1$ and $m_2$ of spatial differentiation and the different
image orientations $\varphi$, if we use an
underlying filter basis in terms of Gaussian derivative responses
along the coordinate axes based on the regular
Gaussian scale-space concept formed from convolutions with the
rotationally symmetric Gaussian kernel. If we instead base the filter
banks on elongated affine Gaussian kernels, the amount of
computational work will grow even more, since a non-separable
convolution would then have to be performed for each image
orientation, each combination of the scale parameters $\sigma_1$ and
$\sigma_2$, and for each order of spatial differentiation, as determined by
the parameters $m_1$ and $m_2$.

If we, on the other hand, base the analysis on the discrete scale-space
concept, by which derivative approximations can be computed from the
raw discrete scale-space representation, by applying small-support
central difference masks, then the amount of computational work can be
decreased substantially, since then for each new combination of the
orders $m_1$ and $m_2$ of differentiation, we only need to apply a new
small-support discrete filter mask. In the case, when the underlying
scale-space representation is based on convolutions with the rotationally symmetric
Gaussian kernel, we can use the same underlying, once and for all spatially
smoothed image as the input for computing filter bank responses for
all the possible orientations. In the case, when the underlying scale-space
representation is instead based on convolutions with affine Gaussian
kernels, we do, of course, have to redo the underlying spatial smoothing
operation for each combination of the parameters $\sigma_1$,
$\sigma_2$ and $\varphi$. We can, however, nevertheless reuse the same
underlying spatially smoothed image for all the combinations of the orders $m_1$ and
$m_2$ of spatial differentiation.

\subsection{Method for defining discrete directional derivative
  approximation masks}

To define a discrete derivative approximation mask
$\delta_{\varphi^{m_1} \bot\varphi^{m_2}}$,
for computing an
approximation of the directional derivative
$L_{\varphi^{m_1} \bot\varphi^{m_2}}$ from an already smoothed
scale-space representation $L$ according to
\begin{equation}
  \label{eq-dir-der-mask-concept-def}
  L_{\varphi^{m_1} \bot\varphi^{m_2}} = \delta_{\varphi^{m_1} \bot\varphi^{m_2}} L,
\end{equation}
for a given
image orientation $\varphi$ and two orders $m_1$ and $m_2$ of spatial
differentiation along the directions $\varphi$ and $\bot\varphi$,
respectively, we can proceed as follows:
\begin{enumerate}
\item
  Combine the continuous directional derivative operators
  (\ref{eq-dphi-def}) and (\ref{eq-dorth-def}) to a joint
  directional derivative operator of the form:
  \begin{equation}
     \label{eq-cont-dir-der-mask-combined}
     \partial_{\varphi^{m_1} \bot\varphi^{m_2}} 
     = \partial_{\varphi}^{m_1} \, \partial_{\bot\varphi}^{m_2}.
  \end{equation}
\item
   Expand the operator (\ref{eq-cont-dir-der-mask-combined}) by formal
   operator calculus over (\ref{eq-dphi-def}) and
   (\ref{eq-dorth-def}) to an expanded representation in terms of a
   linear combination of partial derivative operators
   $\partial_{x^{\alpha} y^{\beta}}$ along the Cartesian coordinate
   directions of the form:
   \begin{equation}
     \partial_{\varphi^{m_1} \bot\varphi^{m_2}} =
     \sum_{k = 0}^{m_1 + m_2}
       w_{k}^{(m_1, m_2)}(\varphi) \, \partial_{x^k y^{m_1 + m_2 - k}},
   \end{equation}
   where the directional weight functions $w_{k}^{(m_1, m_2)}(\varphi)$ are
   polynomials in terms of $\cos \varphi$ and $\sin \varphi$.
\item
  Transfer the partial directional derivative operator
  $\partial_{\varphi^{m_1} \bot\varphi^{m_2}}$ to a corresponding
  directional derivative approximation mask
  $\delta_{\varphi^{m_1} \bot\varphi^{m_2}}$, while simultaneously
  transferring all the Cartesian partial derivative operators
  $\partial_{x^{\alpha} y^{\beta}}$ to corresponding discrete derivative
  approximation masks $\delta_{x^{\alpha} y^{\beta}}$, which leads to:
  \begin{equation}
    \label{eq-dir-der-mask-weight-expr}
    \delta_{\varphi^{m_1} \bot\varphi^{m_2}} =
    \sum_{k = 0}^{m_1 + m_2}
      w_{k}^{(m_1, m_2)}(\varphi) \, \delta_{x^k y^{m_1 + m_2 - k}}.
   \end{equation}
\end{enumerate}
In this way, we obtain explicit expressions for compact discrete
directional derivative approximation masks, as depending on the orders
$m_1$ and $m_2$ of spatial differentiation and the image direction
$\varphi$.

Figure~\ref{fig-aff-gauss-kernels} shows corresponding equivalent
affine Gaussian derivative approximation kernels, computed according
to this scheme, by applying
small-support directional derivative approximation masks of these
forms to a sampled affine Gaussian kernel, as parameterized according to
the form  in Appendix~\ref{sec-app-aff-gauss-param-kern}.

Please note, however, that the resulting kernels obtained in this way,
are not in any way intended to be applied to actual image data. Instead,
their purpose is just to illustrate the equivalent effect of first
convolving the input image with a discrete approximation of the
Gaussian kernel, and then applying a set of small-support directional
derivative approximation masks, for different combinations of the
spatial orders of differentiation, to the spatially smoothed image
data. In situations when combinations of multiple orders of spatial
differentiation are to be used in a computer vision system, for
example, in applications involving filter banks, this form of
discrete implementation will be computationally much more efficient,
compared to applying a set of large-support filter kernels to the same image
data.

By the central difference operators
$\delta_{x^{\alpha} y^{\beta}}$ constituting numerical discrete approximations
of the corresponding partial derivative operators
$\partial_{x^{\alpha} y^{\beta}}$, it follows that the directional
derivative approximation mask
$\delta_{\varphi^{m_1} \bot\varphi^{m_2}}$ will be a numerical
approximation of the continuous directional derivative operator
$\partial_{\varphi^{m_1} \bot\varphi^{m_2}}$.
Thereby, the discrete analogue of the directional derivative operator
according to (\ref{eq-dir-der-mask-concept-def}), from a discrete
approximation $L$ of the scale-space representation of an input image
$f$, will constitute a numerical approximation of the corresponding
continuous directional derivative of the underlying continuous image,
provided that the input image has been sufficiently well sampled, and
provided that the discrete approximation of scale-space smoothing is a
sufficiently good approximation of the corresponding continuous
Gaussian smoothing operation.

In practice, the resulting directional derivative masks will be
of size $3 \times 3$ for first- and second-order derivatives
and of size $5 \times 5$ for third- and fourth-order derivatives.
Thus, once the underlying code for expressing these relationships
has been written, these directional derivative approximation masks
are extremely easy and efficient to apply in practice.

Appendix~\ref{sec-app-dir-der-ops} gives explicit expressions for the
resulting discrete directional derivative approximation masks
for spatial differentiation orders up to 4, whereas
Appendix~\ref{sec-app-der-approx-masks} gives explicit expressions for
the underlying Cartesian discrete derivative approximation masks up to
order 4 of spatial differentiation.

The conceptual construction of compact directional derivative
approximation masks performed in this way generalizes the notion of
steerable filters
(Freeman and Adelson \citeyear{FA91},
Perona \citeyear{Per92-ECCV,Per95-PAMI},
Beil \citeyear{Bei94-PRL},
Simoncelli and Farid \citeyear{SimFar96-IP},
Hel-Or and Teo \citeyear{HelTeo98-JMIV})
to a wide class of filter banks, that can
be computed in a very efficient manner, once an initial smoothing
stage by scale-space filtering, or some approximation thereof, has been computed.

\subsection{Scale-space properties of directional derivative
  approximations computed by applying small-support directional
  derivative approximation masks to smoothed image data}

Note, in particular, that if we compute discrete approximations of
directional derivatives based on a discrete scale-space representation
computed using the discrete analogue of the Gaussian kernel
according to Section~\ref{sec-disc-anal-gauss},
then discrete scale-space
properties will hold also for the discrete approximations of
directional derivatives, in the sense that: (i)~cascade smoothing
properties will hold between directional derivative approximations at
different scales, and (ii)~the discrete directional derivative
approximations will obey non-enhancement of local extrema with
increasing scale.

\section{Summary and conclusions}
\label{sec-summ-concl}

We have presented an in-depth treatment of different ways of
discretizing the Gaussian smoothing operation and the computation of
Gaussian derivatives, for purposes in scale-space analysis and deep learning.
Specifically, we have considered the following three main ways
of discretizing the basic scale-space operations, in terms of either:
\begin{itemize}
\item
  sampling the Gaussian kernel and the Gaussian derivative kernels,
\item
  integrating the Gaussian kernel and the Gaussian derivative
  kernels over the support regions of the pixels, or
\item
  using a genuinely discrete scale-space theory,
  based on convolutions with the discrete analogue of the Gaussian
  kernel, complemented with derivative approximations computed by
  applying small-support central difference operators to the
  spatially smoothed image data.
\end{itemize}
To analyze the properties of these different ways of discretizing the
Gaussian smoothing and Gaussian derivative computation operations, we
have in Section~\ref{disc-approx-gauss-smooth}
defined a set of quantifying performance measures, for which we
have studied their behaviour as function of the scale parameter from
very low to moderate levels of scale.

Regarding the purely spatial smoothing operation, the discrete
analogue of the Gaussian kernel stands out as having the best
theoretical properties over the entire scale range, from scale levels
approaching zero to coarser scales. The results obtained from the
sampled Gaussian kernel may deviate substantially from their
continuous counterparts, when the scale parameter $\sigma$ is
less than about 0.5 or 0.75. For $\sigma$ greater than about 1, the sampled Gaussian kernel
does, on the other hand, lead to numerically very good approximations
of results obtained from the corresponding continuous theory.

Regarding the computation of Gaussian derivative responses, we do also
in Sections~\ref{disc-approx-gauss-ders}
and~\ref{disc-approx-sc-norm-ders}
find that, when applied to polynomial input to reveal the accuracy of
the numerical approximations, the sampled Gaussian derivative kernels and the integrated
Gaussian derivative kernels do not lead to numerically accurate or consistent
derivative estimates, when the scale parameter $\sigma$ is less than
about 0.5 or 0.75. The integrated Gaussian kernels degenerate in somewhat
less strong ways, for very fine scale levels than the sampled
Gaussian derivative kernels, implying that the integrated Gaussian
derivative kernels may have better ability to handle very fine scales
than the sampled Gaussian derivative kernels. At coarser scales, the integrated
Gaussian kernels do, on the other hand, lead to numerically less
accurate estimates of the corresponding continuous counterparts, than
the sampled Gaussian derivative kernels.

At very fine levels of scale, the discrete analogues of the Gaussian
kernels stand out as giving the numerically far best estimates of derivative
computations for polynomial input.
When the scale parameter $\sigma$ exceeds about 1, the sampled
Gaussian derivative kernels do, on the other hand, lead to the
numerically closest estimates to those obtained from the fully
continuous theory.

The fact that the sampled Gaussian derivative kernels for sufficiently
coarse scales lead to the closest approximations of the corresponding
fully continuous theory should, however, not preclude from basing the
analysis on the discrete analogues of Gaussian derivatives at coarser
scales. If necessary, deviations between the results obtained from the
discrete analogues of Gaussian derivatives and the corresponding fully
continuous theory can, in principle, be compensated for by
complementary calibration procedures, or by deriving corresponding
genuinely discrete analogues of the relevant entities in the analysis.
Additionally, in situations
when a larger number of Gaussian derivative responses are to be
computed simultaneously, this can be accomplished with substantially
higher computational efficiency, if basing the scale-space analysis on
the discrete analogue of the Gaussian kernel, which only involves a
single spatial smoothing stage of large spatial support, from which each
derivative approximation can then be computed using a small-support
central difference operator.

As a complement to the presented methodologies of discretizing
Gaussian smoothing and Gaussian derivative computations,
we have also in Section~\ref{disc-approx-dir-ders}
presented a computationally very efficient ways of
computing directional derivatives of different orders and of different
orientations, which is highly useful for computing filter bank type
responses for different purposes in computer vision.
When using the discrete analogue of the Gaussian kernel for smoothing,
the presented discrete directional derivative approximation masks can
be applied at any scale. If using either sampled Gaussian kernels or
integrated Gaussian kernels for spatial smoothing, including extensions 
of rotationally symmetric kernels to anisotropic affine Gaussian kernels,
the discrete derivative approximation masks can be used, provided
that the scale parameter is sufficiently large in relation to the
desired accuracy of the resulting numerical approximation.

Concerning the orders of spatial differentiation, we have in this treatment,
for the purpose of presenting explicit expressions and quantitative
experimental results,
limited ourselves to spatial derivatives up to order 4. A motivation
for this choice is 
the observation by Young (\citeyear{You85-GM,You87-SV}) that receptive fields
up to order 4 have been observed in the primary visual cortex of
higher mammals, why this choice should then cover a majority of the
intended use cases.

It should be noted, however, that an earlier version of the theory for discrete
derivative approximations, based on convolutions with the discrete
analogue of the Gaussian kernel followed by central difference
operators, has, however, been demonstrated to
give useful results with regard to the sign of differential invariants
that depend upon derivatives up to order 5 or 6, for purposes of
performing automatic scale selection, when detecting edges or ridges
from spatial image data (Lindeberg \citeyear{Lin98-IJCV}).
Hence, provided that appropriate care is taken in the design of the
visual operations that operate on image data, this theory could also
be applied for higher orders of spatial differentiation.

\subsection{Extensions of the approach}


Concerning extensions of the approach, with regard to applications in
deep learning, for which the modified Bessel functions $I_n(s)$, underlying the
definition of the discrete analogue of the Gaussian kernel
$T_{\disc}(n;\; s)$ according to (\ref{eq-disc-gauss}), are
currently generally not available in standard frameworks for deep learning, a
possible alternative approach consists of instead replacing the previously
treated sampled Gaussian derivative kernels
$T_{\sampl,x^{\alpha}}(n;\; s)$ according to
(\ref{eq-sampl-gauss-der}) or the integrated
Gaussian derivative kernels $T_{\intdisc,x^{\alpha}}(n;\; s)$
according to (\ref{eq-def-int-gauss-der}) by the families of
{\em hybrid discretization approaches obtained by:
(i)~first smoothing the image with either the normalized sampled
Gaussian kernel $T_{\normsampl}(n;\; s)$ according to
(\ref{eq-def-norm-sampl-gauss}) or the integrated Gaussian kernel
$T_{\intdisc}(n;\; s)$ according to (\ref{eq-def-int-gauss-kern}),
and then applying central difference operators $\delta_{x^{\alpha}}$ of the form
(\ref{eq-def-cent-diff-op-arb-order}) to the spatially smoothed data\/}.

When multiple Gaussian
derivative responses of different orders are to be computed at the same scale level,
such an approach would, in analogy to the previously treated
discretization approach,
based on first smoothing the image with the discrete analogue of the
Gaussian kernel $T_{\disc}(n;\; s)$ according to (\ref{eq-disc-gauss})  and then applying central difference
operators $\delta_{x^{\alpha}}$ of the form (\ref{eq-def-cent-diff-op-arb-order})
to the spatially smoothed data, resulting in equivalent discrete
derivative approximation kernels $T_{\disc,x^{\alpha}}(n;\, s)$
according to  (\ref{eq-disc-der-gauss}),
to also be computationally much more efficient, compared to
explicit smoothing with a set of either sampled Gaussian derivative
kernels $T_{\sampl,x^{\alpha}}(n;\; s)$ according to (\ref{eq-sampl-gauss-der})
or integrated Gaussian derivative kernels $T_{\intdisc,x^{\alpha}}(n;\; s)$
according to (\ref{eq-def-int-gauss-der}).

In terms of equivalent convolution kernels for the resulting hybrid discretization
approaches, the corresponding discrete derivative approximation
kernels for these classes of kernels will then be given by
\begin{align}
   \begin{split}
      \label{eq-hybr-normsampl-disc-der}
      T_{\hybrnormsampl,x^{\alpha}}(n;\; s)
      & = (\delta_{x^{\alpha}} T_{\normsampl})(n;\; s),
    \end{split}\\
   \begin{split}
      \label{eq-hybr-int-disc-der}
      T_{\hybrint,x^{\alpha}}(n;\; s)
      & = (\delta_{x^{\alpha}} T_{\intdisc})(n;\; s),
   \end{split}        
\end{align}
with $T_{\normsampl}(n;\; s)$ and $T_{\intdisc}(n;\; s)$ according to
(\ref{eq-def-norm-sampl-gauss}) and (\ref{eq-def-int-gauss-kern}),
respectively.%
\footnote{In practice, these explicit forms for the derivative approximation
kernels would, however, never be used, since
it is computationally much more efficient to instead first perform the
spatial smoothing operation in an initial processing step, and then
applying different combinations of discrete derivative
approximations, in situations when multiple spatial derivatives of
different orders are to be computed at the same scale level.
With regard to theoretical analysis of the properties of these hybrid
discretization approaches, the corresponding equivalent convolution
kernels are, however, important when characterizing the properties of
these methods.}

Such an approach
is in a straightforward way compatible with learning of the scale
parameters by back propagation, based on
automatic differentiation in deep learning environments.
It would be conceptually very straightforward to
extend the theoretical framework and the experimental evaluations
presented in this paper to incorporating also a detailed
analysis of these two additional classes of discrete derivative approximations
for Gaussian derivative operators of hybrid type. For reasons of space constraints,
we have, however, not been able to include corresponding in-depth
analyses of those additional discretization methods here.%
\footnote{In particular, regarding the theoretical properties of
  these hybrid discretization approaches, it should
  be mentioned that, due
  to the approximation of the spatial derivative operators
  $\partial_{x^{\alpha}}$ by the central difference operators
    $\delta_{x^{\alpha}}$, in combination with the unit $l_1$-norm
      normalization of the corresponding spatial smoothing operations,
      the hybrid derivative approximation kernels
      $T_{\hybrnormsampl,x^{\alpha}}$ and
      $T_{\hybrint,x^{\alpha}}(n;\; s)$ according to
      (\ref{eq-hybr-normsampl-disc-der}) and (\ref{eq-hybr-int-disc-der})
      will
      obey similar response properties
      (\ref{eq-monom-resp-disc-gauss-equal-diff-order}) and
      (\ref{eq-monom-resp-disc-gauss-larger-diff-order}) to monomial
      input, as the previously treated approach, based on the
      combination of spatial smoothing using the discrete analogue
      of the Gaussian kernel with central difference operators
      $T_{\disc,x^{\alpha}}(n;\; s)$ according to (\ref{eq-disc-der-gauss}).
      In other words, the hybrid kernels $T_{\hybrnormsampl,x^{\alpha}}(n;\; s)$
      and $T_{\hybrint,x^{\alpha}}(n;\; s)$ according to
      (\ref{eq-hybr-normsampl-disc-der}) and
      (\ref{eq-hybr-int-disc-der})
      will with $p_k(x) = x^k$ obey
      $T_{\hybrnormsampl,x^M}(\cdot;\; s) * p_M(\cdot) = M!$ and
      $T_{\hybrint,x^M}(\cdot;\; s) * p_M(\cdot) = M!$, as well as
      $T_{\hybrnormsampl,x^M}(\cdot;\; s) * p_N(\cdot) = 0$ and
      $T_{\hybrint,x^M}(\cdot;\; s) * p_N(\cdot) = 0$ for $M > N$.
      In these respects, these hybrid discretization approaches
      $T_{\hybrnormsampl,x^{\alpha}}(n;\; s)$ and $T_{\hybrint,x^{\alpha}}(n;\; s)$ could
      be expected to constitute better approximations of Gaussian
      derivative operators at very fine scales, than their
      corresponding non-hybrid counterparts,
      $T_{\normsampl,x^{\alpha}}(n;\; s)$ and
      $T_{\intdisc,x^{\alpha}}(n;\; s)$ according to
      (\ref{eq-def-norm-sampl-gauss}) and
      (\ref{eq-def-int-gauss-kern}). Simultaneously, these hybrid
      approaches will also be computationally much more efficient than
      their non-hybrid counterparts, in situations where Gaussian
      derivatives of multiple orders are to be computed at the same
      scale.
      
      For very small values of the scale parameter,
      the spatial smoothing with the normalized
    sampled Gaussian kernel $T_{\normsampl,x^{\alpha}}(n;\; s)$ can,
    however, again be expected to lead to systematically too small
    amounts of spatial smoothing (see Figure~\ref{fig-gaussstddev}).
   For larger values of the scale parameter, the spatial smoothing
   with the integrated Gaussian kernel $T_{\intdisc,x^{\alpha}}(n;\;
   s)$ would, however, be expected to lead to a scale offset
   (see Figure~\ref{fig-gaussstddev}), influenced by the variance of a
   box filter over each pixel support region. Furthermore, these
   hybrid approaches will not be guaranteed to obey information
   reducing properties from finer to coarser levels of scale,
   in terms of either non-creation of new local
   extrema, or non-enhancement of
   local extrema, as the discrete analogues of Gaussian derivatives
   $T_{\disc,x^{\alpha}}(n;\;  s)$ according to
   (\ref{eq-disc-der-gauss}) obey.

   In situations, where the
   modified Bessel functions of integer order are immediately
   available, the approach, based on the combination of spatial
   smoothing with the discrete analogue of the Gaussian kernel with
   central differences, should therefore be preferable in relation to
   these hybrid approaches. In situations, where the modified Bessel
   functions of integer order are, however, not available as full-fledged
   primitive in a deep learning framework, these hybrid approaches
   could, on the other hand, be considered as interesting
   alternatives to the regular sampled or integrated Gaussian
   derivative kernels $T_{\sampl,x^{\alpha}}(n;\; s)$ and
   $T_{\intdisc,x^{\alpha}}(n;\; s)$,
       because of their substantially better
   computational efficiency, in situations when spatial derivatives of
   multiple orders are to be computed at the same scale.

   What remains to explore, is how these hybrid discretization
   approaches compare to the previously treated three main classes of
   discretization methods, with respect to the set of quantitative
   performance measures defined and then evaluated experimentally in
   Sections~\ref{sec-perf-meas-gauss-ders}--\ref{sec-num-perf-meas-gauss-ders} and
   Sections~\ref{sec-meas-sc-sel-perf}--\ref{sec-perf-meas-sc-norm-ders}
   (see (Lindeberg \citeyear{Lin24-JMIV-hybrdisc}) for a treatment about this topic)
   as well as with respect to integration in different types of computer
   vision and/or deep learning algorithms.}


Concerning the formulation of discrete approximations of affine
Gaussian derivative operators, it would also be straightforward to
extend the framework in Section~\ref{disc-approx-dir-ders} to replacing the initial
spatial smoothing step, based on convolution with the sampled affine Gaussian derivative
kernel, by instead using an {\em integrated affine Gaussian derivative
kernel\/}, with its filter coefficients of the form
\begin{multline}
  \label{eq-int-aff-gauss-kern}
  T_{\affint}(m, n;\; \sigma_1, \sigma_2, \varphi) = \\
  = \int_{x = m-1/2}^{m+1/2} \int_{y = n-1/2}^{n+1/2}
          g_{\aff}(x, y;\; \sigma_1, \sigma_2, \varphi) \, dx \, dy,
\end{multline}
with $g_{\aff}(x, y;\; \sigma_1, \sigma_2, \varphi)$ denoting the
continuous affine Gaussian kernel according to
(\ref{eq-def-aff-gauss-cont}) and (\ref{eq-def-aff-gauss-cont-arg}),
with the spatial scale parameters $\sigma_1$ and $\sigma_2$ in the two
orthogonal principal directions of affine Gaussian kernel with
orientation $\varphi$,
and where the integral 
can in a straightforward way be approximated by numerical
integration.

In analogy with the previously presented results,
regarding the spatially isotropic Gaussian scale-space representation,
based on discrete approximations of rotationally symmetric Gaussian
kernels, for which the spatial covariance matrix $\Sigma$ in the
matrix-based formulation of the affine Gaussian kernel
$g_{\aff}(p;\; \Sigma)$ according to (\ref{eq-aff-gauss-def}) is equal to
the identity matrix $I$, such an approach, based on integrated affine
Gaussian kernels, could be expected to have clear advantages compared to the
sampled affine Gaussian kernel
$T_{\affsampl}(m, n;\; \sigma_1, \sigma_2, \varphi)$
according to (\ref{eq-sampl-aff-gauss-kern}), for
very small values of the
spatial scale parameters $\sigma_1$ and $\sigma_2$.


A third line of extensions concerns to evaluate the
influence with regard to performance of using the different types of treated
discrete approximations of the Gaussian derivative
operators, when used in specific computer vision algorithms and/or
deep learning architectures,
which we will address in future work.

\section{Acknowledgements}

Python code, that implements a subset of the discretization methods for
Gaussian smoothing and Gaussian derivatives in this paper, is available
in the pyscsp package, available at GitHub:
\begin{quote}
  https://github.com/tonylindeberg/pyscsp
\end{quote}
as well as through PyPi:
\begin{quote}
  \tt pip install pyscsp
\end{quote}

\appendix

\section{Appendix}
\label{sec-app}

\subsection{Explicit expressions for Gaussian derivative kernels}
\label{sec-app-expl-gauss-ders}

This section gives explicit expressions for the 1-D Gaussian derivative
kernels, that we derive discrete approximations for in
Section~\ref{disc-approx-gauss-ders}. For simplicity, we here parameterize
the kernels in terms of the standard deviation $\sigma$, instead to the
variance $s = \sigma^2$.

Consider the probabilistic Hermite polynomials
$\operatorname{He}_n(x)$, defined by
\begin{equation}
\operatorname{He}_n(x) = (-1)^n e^{x^2/2} \, \partial_{x^n} \left( e^{-x^2/2} \right),
\end{equation}
which implies that
\begin{equation}
\partial_{x^n} \left( e^{-x^2/2} \right) =  (-1)^n \operatorname{He}_n(x) \, e^{-x^2/2} 
\end{equation}
and
\begin{equation}
  \partial_{x^n} \left( e^{-x^2/2\sigma^2} \right)
  =  (-1)^n \operatorname{He}_n(\frac{x}{\sigma}) \,
       e^{-x^2/2\sigma^2} \frac{1}{\sigma^n}.
\end{equation}
This means that the $n$:th-order Gaussian derivative kernel in 1-D can be
written as
\begin{align}
\begin{split}
  \partial_{x^n} \left( g(x;\; \sigma) \right) 
  = \frac{1}{\sqrt{2 \pi} \sigma} \partial_{x^n} \left( e^{-x^2/2\sigma^2} \right) 
\end{split}\nonumber\\
\begin{split}
  \label{eq-gauss-der-herm-pol}
  = \frac{1}{\sqrt{2 \pi} \sigma}  \frac{(-1)^n}{\sigma^n}
      \operatorname{He}_n(\frac{x}{\sigma}) \, e^{-x^2/2\sigma^2}
  = \frac{(-1)^n}{\sigma^n} \operatorname{He}_n(\frac{x}{\sigma}) \, g(x;\; \sigma).
\end{split}
\end{align}
For $n$ up to fourth order of spatial differentiation, we have
\begin{align}
  \begin{split}
      g(x;\; \sigma)
      & = \frac{1}{2 \pi \sigma} \, e^{-x^2/2\sigma^2},
  \end{split}\\
  \begin{split}
      g_x(x;\; \sigma)
      & = -\frac{x}{\sigma^2} \, g(x;\; \sigma),
  \end{split}\\
  \begin{split}
      g_{xx}(x;\; \sigma)
      & = \frac{(x^2 - \sigma^2)}{\sigma^4} \, g(x;\; \sigma),
  \end{split}\\
 \begin{split}
      g_{xxx}(x;\; \sigma)
      & = - \frac{(x^3 - 3 \ \sigma^2 \, x)}{\sigma^6} \, g(x;\; \sigma),
  \end{split}\\
 \begin{split}
      g_{xxxx}(x;\; \sigma)
      & = \frac{(x^4 - 6 \, \sigma^2 \, x^2 + 3 \, \sigma^4)}{\sigma^8} \, g(x;\; \sigma).
  \end{split}
\end{align}

\subsection{Derivation of the integrated Gaussian kernel and the
  integrated Gaussian derivatives}
\label{app-deriv-int-gauss} 

In this appendix, we will give a hands-on derivation of how
convolution with integrated Gaussian kernels or integrated Gaussian
derivative kernels arises from an assumption of extending any discrete
signal to a piecewise constant continuous signal over each pixel
support region. 

Consider a Gaussian derivative kernel of order $\alpha$, where the
special case $\alpha = 0$ corresponds to the regular zero-order
Gaussian kernel. For any one-dimensional continuous signal $f_c(x)$, the
Gaussian derivative response of order $\alpha$ is given by
\begin{equation}
   \label{eq-expl-conv-int-gauss-der}
   L_{x^{\alpha}}(x;\; s)
   = \int_{\xi \in \bbbr} g_{x^{\alpha}}(x - \xi;\, s) \, f_c(\xi) \, d\xi.
\end{equation}
Let us next assume that we have a given discrete input signal $f(n)$,
and from this discrete signal define a step-wise constant continuous
signal $f_c(x)$  according to
\begin{equation}
  f_c(x) = f(n) \quad\quad \mbox{if $-\tfrac{1}{2} < x - n \leq \tfrac{1}{2}$},
\end{equation}
where $n$ denotes the integer nearest to the real-valued coordinate $x$.

The result of subjecting this continuous signal to the continuous Gaussian
derivative convolution integral (\ref{eq-expl-conv-int-gauss-der})
at any integer grid point $x = n$ can therefore be written as
\begin{equation}
   L_{x^{\alpha}}(n;\; s)
   = \sum_{m = -\infty}^{\infty}
         \int_{\xi = m-1/2}^{m+1/2} g_{x^{\alpha}}(n - \xi;\, s) \, f_c(\xi) \, d\xi.
\end{equation}
Now, since $f_c(n - \xi) = f(m)$ within the pixel support region
$m-1/2 < \xi \leq m+1/2$, we can also write this relation as
\begin{equation}
  \label{eq-app-deriv-int-gauss-intermed1}
   L_{x^{\alpha}}(n;\; s)
   = \sum_{m = -\infty}^{\infty} f(m)
         \int_{\xi = m-1/2}^{m+1/2} g_{x^{\alpha}}(n - \xi;\, s) \, d\xi.
\end{equation}
Next, by defining the integrated Gaussian derivative kernel as
\begin{equation}
  T_{\intdisc, x^{\alpha}}(n - m;\; s)
  = \int_{\xi = m-1/2}^{m+1/2} g_{x^{\alpha}}(n - \xi;\, s) \, d\xi,
\end{equation}
it follows that the relation (\ref{eq-app-deriv-int-gauss-intermed1})
can be written as
\begin{multline}
   L_{x^{\alpha}}(n;\; s)
   = \sum_{m = -\infty}^{\infty} f(m) \, T_{\intdisc, x^{\alpha}}(n - m;\; s) = \\
   = \sum_{m = -\infty}^{\infty}T_{\intdisc, x^{\alpha}}(n - m;\; s) \,  f(m),
\end{multline}
which shows that construction of applying the continuous
Gaussian derivative convolution to a stepwise constant signal,
defined as being equal to the discrete signal over each pixel support
region, at the discrete grid points $x = n$ corresponds to discrete convolution
with the integrated Gaussian derivative kernel.

\subsection{$L_1$-norms of Gaussian derivative kernels}
\label{app-L1-norms-gauss-der-kernels}

This appendix gives explicit expressions for the $L_1$-norms of
1-D Gaussian derivative kernels
\begin{equation}
  N_{\alpha}
  = \| g_{\alpha}(\cdot;\; \sigma) \|_1
    = \int_{x \in \bbbr} \left| g_{x^{\alpha}}(x;\; \sigma) \right| \, dx
\end{equation}
for differentiation orders up to 4,
based on Equations~(74)--(77) in Lindeberg (\citeyear{Lin97-IJCV}), while
with the scale normalization underlying those equations, to constant
$L_1$-norms over scale, removed:
\begin{align}
  \begin{split}
    N_0(\sigma)
    & = 1,
  \end{split}\\
  \begin{split}
    N_1(\sigma)
    & = \frac{1}{\sigma} \sqrt{\frac{2}{\pi}} \approx \frac{0.798}{\sigma},
  \end{split}\\
  \begin{split}
    N_2(\sigma)
    & = \frac{1}{\sigma^2} \sqrt{\frac{8}{e \, \pi}} \approx \frac{0.968}{\sigma^2},
  \end{split}\\
  \begin{split}
    N_3(\sigma)
    & = \frac{1}{\sigma^3}
           \left( 1 + \frac{4}{e^{3/2}} \right)
           \sqrt{\frac{2}{\pi}} \approx \frac{1.510}{\sigma^3},
  \end{split}\\
  \begin{split}
    N_4(\sigma)
    & = \frac{1}{\sigma^4}
           \frac{4 \sqrt{3}}{\left (e^{3/2} + e^{\sqrt{3/2}} \sqrt{\pi} \right)}
           \left(
              \sqrt{3 - \sqrt{6}} \, e^{\sqrt{6}} +\sqrt{3 + \sqrt{6}}
            \right)
 \end{split}\nonumber\\
  \begin{split}
    & \approx \frac{2.801}{\sigma^4}.
  \end{split}
\end{align}

\begin{figure*}[hbtp]
  \begin{equation}
    S_4(\sigma)
    = \frac{3 \sqrt[4]{27+11 \sqrt{6}} \sqrt{\frac{5880 \sqrt{2}+4801 \sqrt{3}+267
   \sqrt{485+198 \sqrt{6}}+109 \sqrt{6 \left(485+198 \sqrt{6}\right)}+\left(27+11
   \sqrt{6}+267 \sqrt{49-20 \sqrt{6}}+109 \sqrt{6 \left(49-20 \sqrt{6}\right)}\right)
   e^{\sqrt{6}}}{5+2 \sqrt{6}+\sqrt{49+20 \sqrt{6}}+\left(\sqrt{2}+\sqrt{3}+\sqrt{5+2
   \sqrt{6}}\right) e^{\sqrt{6}}}} }{\left(3+\sqrt{6}\right)^{11/4}}
\times \sigma 
   \end{equation}
  \caption{Exact expression for the spread measure $S_4(\sigma)$ of the 4:th
    order derivative (\ref{eq-Salpha-def-app}) of a 1-D Gaussian kernel.}
  \label{eq-S4-expl-expr}
\end{figure*}

\subsection{Spatial spread measures for Gaussian derivative kernels}
\label{app-spread-measures}

This appendix gives explicit expressions for spread measures in terms
of the standard deviations of the
absolute values of the 1-D Gaussian derivative kernels%
\footnote{Note that since these kernels are symmetric, we can avoid
  the compensation with respect to the mean values.}
\begin{equation}
  \label{eq-Salpha-def-app}
  S_{\alpha}
    = \sqrt{\frac{\int_{x \in \bbbr} x^2 \left| g_{x^{\alpha}}(x;\; \sigma) \right| \, dx}
                         {\int_{x \in \bbbr} \left| g_{x^{\alpha}}(x;\; \sigma) \right| \, dx}}
  \end{equation}
for differentiation orders up to 4:
\begin{align}
  \begin{split}
    S_0(\sigma)
    & = \sigma,
  \end{split}\\
  \begin{split}
    S_1(\sigma)
    & = \sqrt{2} \, \sigma \approx 1.414 \, \sigma,
  \end{split}\\
  \begin{split}
    S_2(\sigma)
    & = \sqrt[4]{\frac{e \, \pi }{2}} \,
           \sqrt{1 + 3 \, \sqrt{\frac{2}{e \, \pi}}
                     - 2\operatorname{erf}\left(\frac{1}{\sqrt{2}}\right)}
     \times \sigma 
  \end{split}\nonumber\\
  \begin{split}
    & \approx 1.498 \, \sigma,
  \end{split}\\
  \begin{split}
    S_3(\sigma)
    & = \sqrt{\frac{28-2 \, e^{3/2}}{4+e^{3/2}}} \times \sigma
        \approx 1.498 \, \sigma,
  \end{split}\\
  \begin{split}
    S_4(\sigma)
    & \approx 1.481 \, \sigma.
  \end{split}
\end{align}
An exact expression for $S_4(\sigma)$ is given in
Figure~\ref{eq-S4-expl-expr}.
The calculations have been performed in Mathematica.

\subsection{Diffusion polynomials in the 1-D continuous case}
\label{sec-app-cont-diff-poly}

This appendix lists diffusion polynomials, that satisfy the 1-D
diffusion equation
\begin{equation}
  \partial_s L
  = \frac{1}{2} \, \partial_{xx} L 
\end{equation}
with initial condition $L(x;\; 0) = f(x)$, for $f(x)$ being monomials
\begin{equation}
  f(x) = x^k
\end{equation}
of orders up to $k = 4$:
\begin{align}
  \begin{split}
    \label{eq-diff-pol-0}
    q_0(x;\; s) = 1,
   \end{split}\\
 \begin{split}
    q_1(x;\; s) = x,
   \end{split}\\
 \begin{split}
    q_2(x;\; s) = x^2 + s,
   \end{split}\\
 \begin{split}
    q_3(x;\; s) = x^3 + 3 \, x \, s,
   \end{split}\\
  \begin{split}
    \label{eq-diff-pol-4}
   q_4(x;\; s) = x^4 + 6 \, x^2 \, s + 3 \, s^2.
   \end{split}
\end{align}
These diffusion polynomials do, in this respect, describe how a monomial
input function $f(x) = x^k$ is affected by convolution with the
continuous Gaussian kernel for standard deviation $\sigma =
\sqrt{s}$.

\subsection{Affine Gaussian scale space}
\label{sec-app-aff-gauss-scsp}

As a more general spatial scale-space representation for 2-D images,
consider the affine Gaussian scale-space representation
\begin{equation}
  L(x, y;\;  \Sigma)
  = (g_{\aff}(\cdot, \cdot;\; \Sigma) * f(\cdot, \cdot))(x, y;\; \Sigma),
\end{equation}
where $g_{\aff}(x, y;\; \Sigma) = g_{\aff}(p;\; \Sigma)$ for $p = (x, y)^T$
represents a 2-D affine Gaussian kernel of the form
\begin{equation}
  \label{eq-aff-gauss-def}
  g_{\aff}(p;\; \Sigma)
  = \frac{1}{2 \pi \sqrt{\det \Sigma}} \, e^{-p^T \Sigma^{-1} p/2},
\end{equation}
with $\Sigma$ denoting any positive definite $2 \times 2$ matrix.
In terms of diffusion equations, this affine scale-space
representation along each ray $\Sigma = s \, \Sigma_0$
in affine scale-space satisfies the affine diffusion equation
\begin{equation}
  \partial_s = \frac{1}{2} \nabla^T (\Sigma_0 \, \nabla L)
\end{equation}
with initial condition $L(\cdot, \cdot;\; 0) = f(\cdot, \cdot)$.

A general rationale for studying and making use of this affine
scale-space representation is that it is closed under affine
transformations, thus leading to affine covariance or affine
equivariance
(Lindeberg \citeyear{Lin93-Dis}; Lindeberg and G{\aa}rding \citeyear{LG96-IVC}).

This closedness under affine transformations has been used for
computing more accurate estimates of local surface orientation from
monocular och binocular cues
(Lindeberg and G{\aa}rding \citeyear{LG96-IVC},
Rodr{\'\i}guez {\em et al.\/} \citeyear{RodDelMor18-SIAM}),
for computing affine invariant image features for
image matching under wide baselines
(Baumberg \citeyear{Bau00-CVPR},
Mikolajczyk and Schmid \citeyear{MikSch04-IJCV},
Mikolajczyk {\em et al.\/} \citeyear{MikTuySchZisMatSchKadGoo05-IJCV},
Tuytelaars and van Gool \citeyear{TuyGoo04-IJCV},
Lazebnik {\em et al.\/}\ \citeyear{LazSchPon05-PAMI},
Rothganger {\em et al.\/}\ \citeyear{RotLazSchPon06-IJCV,RotLazSchPon07-PAMI},
Lia {\em et al.\/}\ \citeyear{LiaLiuHui13-PRL},
Eichhardt and Chetverikov \citeyear{EicChe18-ECCV},
Dai {\em et al.\/} \citeyear{DaiJinZhaNgu20-IP}),
for performing affine invariant segmentation 
(Ballester and Gonz{\'a}lez \citeyear{BalGon98-JMIV}),
for constructing affine covariant SIFT descriptors 
(Mo\-rel and Yu \citeyear{MorGuo09-SIAM-JIS},
 Yu and Morel \citeyear{YuMor09-ASSP};
Sadek {\em et al.\/}\ \citeyear{SadConMeiBalCas12-SIM-JIS}),
for modelling receptive fields in biological vision
(Lindeberg \citeyear{Lin13-BICY,Lin21-Heliyon}),
for affine invariant tracking
 (Giannarou {\em et al.\/}\ \citeyear{GiaVisYan13-PAMI}),
and for formulating affine covariant metrics
 (Fedorov {\em et al.\/} \citeyear{FedAriSadFacBal15-SIAM}).
Affine Gaussian kernels with their related affine Gaussian derivatives
have also been used as a general filter family for a large number of
purposes in computer vision
(Lampert and Wirjadi \citeyear{LamWir06-IP},
Li and Shui \citeyear{LiShu20-SP},
Keilmann \citeyear{KeiGodMogRedSch23-arXiv}).

\subsection{A convenient parameterization of affine Gaussian kernels
  over a 2-D spatial domain}
\label{sec-app-aff-gauss-param-kern}

For the 2-D case, which we will restrict ourselves to henceforth,
we may parameterize the spatial covariance matrix
\begin{equation}
  \Sigma
  = \left(
        \begin{array}{cc}
          C_{xx} & C_{xy} \\
          C_{xy} & C_{yy}
        \end{array}
    \right),
  \end{equation}
underlying the definition of affine Gaussian kernels in the affine
Gaussian scale-space theory according to
Appendix~\ref{sec-app-aff-gauss-scsp},
in terms of its eigenvalues $\lambda_1 >0$ and $\lambda_2 > 0$ as well as an
 image orientation $\varphi$, as
 \begin{align}
    \begin{split}
       \label{eq-expl-par-Cxx}
       C_{xx} & = \lambda_1 \cos^2 \varphi + \lambda_2 \sin^2 \varphi,
    \end{split}\\
    \begin{split}
          \label{eq-expl-par-Cxy}
        C_{xy} & = (\lambda_1 - \lambda_2)  \cos \varphi  \, \sin \varphi,
    \end{split}\\
    \begin{split}
       \label{eq-expl-par-Cyy}
        C_{yy} & = \lambda_1  \sin^2 \varphi + \lambda_2  \cos^2 \varphi,
   \end{split}
\end{align}
and where we may additionally parameterize the eigenvalues in terms
of corresponding standard deviations $\sigma_1$ and $\sigma_2$
according to
\begin{align}
  \begin{split}
       \label{eq-expl-par-lambda1}    
       \lambda_1 & = \sigma_1^2,
    \end{split}\\
    \begin{split}
       \label{eq-expl-par-lambda2}        
       \lambda_2 & = \sigma_2^2,
   \end{split}
\end{align}
which then leads to the following explicit expression for the affine
Gaussian derivative kernel
\begin{equation}
  \label{eq-def-aff-gauss-cont}
  g_{\aff}(x, y;\; \sigma_1, \sigma_2, \varphi) =
  \frac{1}{2 \pi \sigma_1 \sigma_2} \,
  e^{-A/2 \, \sigma_1^2 \, \sigma_2^2},
\end{equation}
where
\begin{multline}
  \label{eq-def-aff-gauss-cont-arg}  
  A = (\sigma_2^2 \, x^2 + \sigma_1^2 \, y^2)  \cos^2 \varphi  
        + (\sigma_1^2 \, x^2 + \sigma_2^2 \, y^2) \sin^2 \varphi  \\
        - 2 \, (\sigma_1^2 - \sigma_2^2) \, \cos \varphi \, \sin \varphi  \, x \, y.
\end{multline}
The sampled affine Gaussian kernel is then given by
\begin{equation}
  \label{eq-sampl-aff-gauss-kern}
  T_{\affsampl}(m, n;\; \sigma_1, \sigma_2, \varphi)
  = g_{\aff}(m, n;\; \sigma_1, \sigma_2, \varphi).
\end{equation}
At very fine scales, this discrete kernel will suffer from similar problems as
the sampled rotationally symmetric Gaussian kernel according to
(\ref{eq-sampl-gauss}), in the sense that:
(i)~the filter coefficients may exceed 1 for very small values of
  $\sigma_1$ or $\sigma_2$, 
(ii)~the filter coefficients may sum up to a value larger than 1 for
very small
values of $\sigma_1$ or $\sigma_2$, and
(iii)~it may not be a sufficiently good numerical approximation of a
spatial differentiation operator.
For sufficiently large values of the scale parameters $\sigma_1$ and
$\sigma_2$, however, this kernel can nevertheless be expected to
constitute a reasonable approximation of the continuous affine
Gaussian scale-space theory, for purposes of
computing coarse-scale receptive field responses, for filter-bank
approaches to {\em e.g.\/} visual recognition.

Alternatively, it is also possible to define a genuine discrete theory
for affine kernels (Lindeberg \citeyear{Lin17-arXiv-discaff}).
A limitation of that theory, however, is that a positivity requirement
on the resulting spatial discretization imposes an upper bound on the
eccentricities of the shapes of the kernels (as determined by the
ratio between the eigenvalues $\lambda_1$ and $\lambda_2$ of $\Sigma$),
and implying that the kernel shapes must not be too eccentric, to be
represented within the theory. For this reason, we do not consider
that theory in more detail here, and refer the reader to the original
source for further details.

\subsection{Explicit expressions for discrete directional derivative
  approximation masks}
\label{sec-app-dir-der-ops}

This appendix gives explicit expressions for directional derivative
operator masks $\delta_{\varphi^{m_1} \bot\varphi^{m_2}}$ according to
(\ref{eq-dir-der-mask-concept-def}) and
(\ref{eq-dir-der-mask-weight-expr}), in terms of underlying discrete
derivative approximation masks along the Cartesian coordinate
directions according to Appendix~\ref{sec-app-der-approx-masks}.

\medskip

\noindent{\bf Of order 1:}
\begin{align}
  \begin{split}
    \delta_{\varphi}
     = & \cos \varphi \, \delta_x + \sin \varphi \, \delta_y,
  \end{split}\\
  \begin{split}
   \delta_{\bot\varphi}
    = & - \sin \varphi \, \delta_x + \cos \varphi \, \delta_y.
  \end{split}
\end{align}

\medskip

\noindent{\bf Of order 2:}
\begin{align}
  \begin{split}
    \delta_{\varphi\varphi}
    = & \cos^2 \varphi \, \delta_{xx} + 
            2 \cos \varphi \, \sin \varphi \, \delta_{xy} + 
            \sin^2 \varphi \, \delta_{yy},
     \end{split}\\
   \begin{split}
    \delta_{\varphi\bot\varphi}
    = & \cos \varphi \, \sin \varphi \, (\delta_{yy} - \delta_{xx})
           + (\cos^2 \varphi - \sin^2 \varphi) \, \delta_{xy},
  \end{split}\\
  \begin{split}
    \delta_{\bot\varphi\bot\varphi}
     = & \sin^2 \varphi \,  \delta_{xx} 
             - 2 \cos \varphi \, \sin \varphi \, \delta_{xy} 
             + \cos^2 \varphi \,  \delta_{yy}.
  \end{split}
\end{align}

\medskip

\noindent{\bf Of order 3:}
\begin{align}
   \begin{split}
     \delta_{\varphi\varphi\varphi}
     =  & \cos^3 \varphi \,  \delta_{xxx} + 
           3 \cos^2 \varphi \, \sin \varphi \, \delta_{xxy} 
    \end{split}\nonumber\\
    \begin{split}
        & + 3 \cos \varphi \, \sin^2 \varphi \, \delta_{xyy} +
           \sin^3 \varphi \, \delta_{yyy},
    \end{split}\\
   \begin{split}
     \delta_{\varphi\varphi\bot\varphi}
     = & - \cos^2 \varphi \, \sin \varphi \,  \delta_{xxx}
     \end{split}\nonumber\\
    \begin{split}
          & + (\cos^3 \varphi - 2 \, \cos \varphi \,  \sin^2 \varphi)  \, \delta_{xxy}
     \end{split}\nonumber\\
     \begin{split}
        & - (\sin^3 \varphi - 2 \,  \cos^2 \varphi \, \sin \varphi) \, \delta_{xyy}
    \end{split}\nonumber\\
    \begin{split}
         &  + \cos \varphi \, \sin^2 \varphi \, \delta_{yyy},
     \end{split}\\
    \begin{split}
     \delta_{\varphi\bot\varphi\bot\varphi}
     = & \cos \varphi \,  \sin^2 \varphi \, \delta_{xxx}
    \end{split}\nonumber\\
    \begin{split}
         & + (\sin^3 \varphi - 2 \, \cos^2 \varphi \,  * \sin \varphi) \, \delta_{xxy}
     \end{split}\nonumber\\
     \begin{split}
         &  + (\cos^3 \varphi - 2 \, \cos \varphi \, \sin^2 \varphi2) \, \delta_{xyy}
    \end{split}\nonumber\\
    \begin{split}
          & + \cos^2 \varphi \,  \sin \varphi \, \delta_{yyy},
     \end{split}\\
   \begin{split}
     \delta_{\bot\varphi\bot\varphi\bot\varphi}
     = & - \sin^3 \varphi \, \delta_{xxx}
             + 3 \, \sin^2 \varphi \, \cos \varphi \, \delta_{xxy}
     \end{split}\nonumber\\
     \begin{split}
         &  - 3 \, \sin \varphi \, \cos^2 \varphi \, \delta_{xyy}
           + \cos^3 \varphi \, \delta_{yyy}.
     \end{split}
\end{align}

\medskip

\noindent{\bf Of order 4:}
\begin{align}
   \begin{split}
     \delta_{\varphi\varphi\varphi\varphi}
     =  & \cos^4 \varphi \,  \delta_{xxxx} 
           + 4 \,  \cos^3 \varphi \,  \sin \varphi \,  \delta_{xxxy} 
    \end{split}\nonumber\\
    \begin{split}
          & + 6 \, \cos^2 \varphi \, \sin^2 \varphi \, \delta_{xxyy} 
    \end{split}\nonumber\\
    \begin{split}
           & + 4 \, \cos \varphi \, \sin^3 \varphi \, \delta_{xyyy} 
              + \sin^4 \varphi \, \delta_{yyyy},
    \end{split}\\
  \begin{split}
     \delta_{\varphi\varphi\varphi\bot\varphi}
     =  & - \cos^3 \varphi \, \sin \varphi \, \delta_{xxxx} 
    \end{split}\nonumber\\
    \begin{split}
          & + (\cos^4 \varphi - 3 \, \cos^2 \varphi \, \sin^2 \varphi) \, \delta_{xxxy}
    \end{split}\nonumber\\
    \begin{split}
          & + 3 \,  (\cos^3 \varphi \, \sin \varphi - \cos \varphi \, \sin^3 \varphi) \,
           \delta_{xxyy} 
    \end{split}\nonumber\\
    \begin{split}
           & + (3 \, \cos^2 \varphi \, \sin^2 \varphi - \sin^4 \varphi) \, \delta_{xyyy}
    \end{split}\nonumber\\
    \begin{split}
           & + \cos \varphi \, \sin^3 \varphi \, \delta_{yyyy},
    \end{split}\\
  \begin{split}
     \delta_{\varphi\varphi\bot\varphi\bot\varphi}
     =  & \cos^2 \varphi \, \sin^2 \varphi \, \delta_{xxxx}
     \end{split}\nonumber\\
    \begin{split}
          & + 2 \,  (\cos \varphi \, \sin^3 \varphi - \cos^3 \varphi \, \sin \varphi) \,
             \delta_{xxxy}
    \end{split}\nonumber\\
    \begin{split}
           & + (\cos^4 \varphi - 4 \, \cos^2 \varphi \, \sin^2 \varphi + \sin^4 \varphi) \,
              \delta_{xxyy} 
    \end{split}\nonumber\\
    \begin{split}
           & + 2 \, (\cos^3 \varphi \, \sin \varphi - \cos \varphi \, \sin^3 \varphi) \,
           \delta_{xyyy}
    \end{split}\nonumber\\
    \begin{split}
           & + \cos^2 \varphi \, \sin^2 \varphi \, \delta_{yyyy},
    \end{split}\\
    \begin{split}
       \delta_{\varphi\bot\varphi\bot\varphi\bot\varphi}
       =  & - \cos \varphi \, \sin^3 \varphi \, \delta_{xxxx}
    \end{split}\nonumber\\
    \begin{split}
           & + (3 \, \cos^2 \varphi \, \sin^2 \varphi - \sin^4 \varphi) \, \delta_{xxxy}
    \end{split}\nonumber\\
    \begin{split}
           & + 3 \, (\cos \varphi \, \sin^3 \varphi - \cos^3 \varphi \, \sin \varphi)  \,
            \delta_{xxyy}
   \end{split}\nonumber\\
    \begin{split}
            & + (\cos^4 \varphi - 3 \, \cos^2 \varphi \, \sin^2 \varphi) \, \delta_{xyyy}
    \end{split}\nonumber\\
    \begin{split}
            & + \cos^3 \varphi \, \sin \varphi \, \delta_{yyyy},
    \end{split}\\
    \begin{split}
       \delta_{\bot\varphi\bot\varphi\bot\varphi\bot\varphi}
       =  & \sin^4 \varphi \, \delta_{xxxx}
    \end{split}\nonumber\\
    \begin{split}
            & - 4 \, \sin^3 \varphi \,  \cos \varphi \, \delta_{xxxy}
    \end{split}\nonumber\\
    \begin{split}
            & + 6 \, \sin^2 \varphi \, \cos^2 \phi \, \delta_{xxyy}
    \end{split}\nonumber\\
    \begin{split}
             & - 4 \, \sin \varphi \, \cos^3 \varphi \, \delta_{dxyyy}
    \end{split}\nonumber\\
    \begin{split}
             & + \cos^4 \varphi \, \delta_{yyyy}.
    \end{split}
\end{align}

\subsection{Explicit expressions for discrete derivative approximation masks}
\label{sec-app-der-approx-masks}

This appendix gives explicit expressions for discrete derivative
approximation masks $\delta_{x^{\alpha} y^{\beta}}$ up to order 4 for
the case of a 2-D spatial image domain.

\medskip

\noindent{\bf Of order 1 embedded in masks of size $3 \times 3$:}

\begin{align}
  \begin{split}
     \delta_x &
     = \left(
           \begin{array}{ccc}
              0       &  0 &   0 \\
              -\tfrac{1}{2} &  0 & +\tfrac{1}{2} \\
              0       & 0  & 0
           \end{array}
         \right)
   \end{split}\\
  \begin{split}
     \delta_y &
     = \left(
           \begin{array}{ccc}
               0 & +\tfrac{1}{2} & 0 \\
               0 &  0                 & 0 \\
               0 & -\tfrac{1}{2} & 0
           \end{array}
         \right)
   \end{split}
\end{align}

\medskip

\noindent{\bf Of order 2 embedded in masks of size $3 \times 3$:}

\begin{align}
  \begin{split}
     \delta_{xx} &
     = \left(
           \begin{array}{ccc}
              0   & 0   & 0 \\
              +1 & -2 & 1 \\
              0   & 0   & 0 
           \end{array}
         \right)
   \end{split}\\
  \begin{split}
     \delta_{xy} &
     = \left(
           \begin{array}{ccc}
              -\tfrac{1}{4} & 0 & +\tfrac{1}{4} \\
              0                  & 0 &  0 \\
              +\tfrac{1}{4} & 0 & -\tfrac{1}{4} 
           \end{array}
         \right)
   \end{split}\\
  \begin{split}
     \delta_{yy} &
     = \left(
           \begin{array}{ccc}
               0 & +1 & 0 \\
               0 & -2 & 0 \\
               0 & +1 & 0
           \end{array}
         \right)
   \end{split}
\end{align}

\medskip

\noindent{\bf Of order 3 embedded in masks of size $5 \times 5$:}

\begin{align}
 \begin{split}
     \delta_{xxx} &
     = \left(
           \begin{array}{ccccc}
              0                  &  0  & 0 & 0   & 0 \\
              0                  &  0  & 0 & 0   & 0 \\
              -\tfrac{1}{2} & +1 & 0 & -1 & \tfrac{1}{2} \\
              0                  &  0  & 0 & 0   & 0 \\
              0                  &  0  & 0 & 0   & 0 
           \end{array}
         \right)
   \end{split}\\
  \begin{split}
     \delta_{xxy} &
     = \left(
           \begin{array}{ccccc}
              0 &   0                & 0     & 0                  & 0 \\
              0 & +\tfrac{1}{2} &  -1 & +\tfrac{1}{2} & 0 \\
              0 &   0                & 0     & 0                  & 0 \\
              0 & -\tfrac{1}{2} &  +1 & -\tfrac{1}{2} & 0 \\
              0 &   0                & 0     & 0                  & 0 
           \end{array}
         \right)
   \end{split}\\
   \begin{split}
     \delta_{xyy} &
     = \left(
           \begin{array}{ccccc}
              0 &   0                & 0 & 0                  & 0 \\
              0 & -\tfrac{1}{2} & 0 & +\tfrac{1}{2} & 0 \\
              0, & +1               & 0 & -1                & 0 \\
              0 & -\tfrac{1}{2} & 0 & +\tfrac{1}{2} & 0 \\
              0 &   0                & 0 & 0                  & 0 
           \end{array}
         \right)
   \end{split}\\
   \begin{split}
     \delta_{yyy} &
     = \left(
           \begin{array}{ccccc}
              0 & 0 & +\tfrac{1}{2} & 0 & 0 \\
              0 & 0 & -1                & 0 & 0 \\
              0 & 0 & 0                  & 0 & 0 \\
              0 & 0 & +1                & 0 & 0 \\
              0 & 0 & -\tfrac{1}{2} & 0 & 0 
           \end{array}
         \right)
   \end{split}
\end{align}

\medskip

\noindent{\bf Of order 4 embedded in masks of size $5 \times 5$:}

\begin{align}
 \begin{split}
     \delta_{xxxx} &
     = \left(
           \begin{array}{ccccc}
              0 & 0   & 0 & 0  & 0 \\
              0 & 0   & 0 & 0  & 0 \\
              1 & -4 & 6 & -4 & 1 \\
              0 & 0   & 0 & 0  & 0 \\
              0 & 0   & 0 & 0  & 0 
           \end{array}
         \right)
   \end{split}\\
  \begin{split}
     \delta_{xxxy} &
     = \left(
           \begin{array}{ccccc}
              0                  & 0                  & 0 & 0                  & 0 \\
              -\tfrac{1}{4} & +\tfrac{1}{2} & 0 & -\tfrac{1}{2} & +\tfrac{1}{4} \\
              0                  & 0                  & 0 & 0                  & 0 \\
              +\tfrac{1}{4} & -\tfrac{1}{2} & 0 & +\tfrac{1}{2} & -\tfrac{1}{4} \\
              0                  & 0                  & 0 & 0                  & 0 
           \end{array}
         \right)
   \end{split}\\
   \begin{split}
     \delta_{xxyy} &
     = \left(
           \begin{array}{ccccc}
               0 & 0    & 0  & 0    & 0 \\
               0 & +1 & -2 & +1 & 0 \\
               0 & -2  & +4 & -2 & 0 \\
               0 & +1 & -2 & +1 & 0 \\
               0 & 0    & 0  & 0    & 0 
           \end{array}
         \right)
   \end{split}\\
   \begin{split}
     \delta_{xyyy} &
     = \left(
           \begin{array}{ccccc}
              0 & -\tfrac{1}{4} & 0 & +\tfrac{1}{4} & 0 \\
              0 & +\tfrac{1}{2} & 0 & -\tfrac{1}{2} & 0 \\
              0 & 0                   & 0  & 0                 & 0 \\
              0 & -\tfrac{1}{2} & 0 & +\tfrac{1}{2} & 0 \\
              0 & +\tfrac{1}{4} & 0 & -\tfrac{1}{4} & 0 
           \end{array}
         \right)
   \end{split}\\
   \begin{split}
     \delta_{yyyy} &
     = \left(
           \begin{array}{ccccc}
              0 & 0 & +1 & 0 & 0 \\
              0 & 0 & -4 & 0 & 0 \\
              0 & 0 & +6 & 0 & 0 \\
              0 & 0 & -4 & 0 & 0 \\
              0 & 0 & +1 & 0 & 0 
           \end{array}
         \right)
   \end{split}
\end{align}

\subsection{Clarifications in relation to an evaluation of what is
  referred to as ``Lindeberg's smoothing method'' by
  Rey-Otero and Delbracio (\citeyear{OteDel16-IPOL})}
\label{app-clarif-lindebergs-smoothing-method}

With regard to an evaluation of a method, referred to as
  ``Lindeberg's smoothing method' in
  (Rey-Otero and Delbracio \citeyear{OteDel16-IPOL}), some
  clarifications would be needed concerning the experimental
  comparison they perform in that work, since that comparison is not
  made in relation to any of the best methods that arise from the
  discrete scale-space theory introduced in
  (Lindeberg \citeyear{Lin90-PAMI}) and then extended in
  (Lindeberg \citeyear{Lin93-Dis} Chapters~3 and~4).

  Rey-Otero and Delbracio (\citeyear{OteDel16-IPOL}) 
  compare to an Euler-forward discretization of the semi-discrete
  diffusion equation (Equation~(4.30) in Lindeberg
  \citeyear{Lin93-Dis})
  \begin{equation}
    \partial_t L
    = \frac{1}{2} \,
    (
      (1 - \gamma) \, \nabla_5^2 + \gamma \, \nabla_{\times}^2 
    ) \, L
  \end{equation}
  with initial condition $L(x, y;\; 0) = f(x, y)$,
  that determines the evolution of a 2-D discrete
  scale-space representation over scale. The corresponding
  discrete scale-space representation, according to Lindeberg's
  discrete scale-space theory, can, however, be computed more
  accurately, using the explicit
  expression for the Fourier transform of the underlying discrete
  family of scale-space kernels, according to Equation~(4.24) in
  (Lindeberg \citeyear{Lin93-Dis}), and thus without using any {\em a priori\/}
  restriction to discrete levels in the scale direction, as used by
  Rey-Otero and Delbracio (\citeyear{OteDel16-IPOL}).
  
  Additionally, concerning the choice of the
  parameter value $\gamma$, that determines the relative weighting
  between the contributions from the five-point $\nabla_5^2$ and
  cross-point $\nabla_{\times}^2$ 
  discretizations of the Laplacian operator, Rey-Otero and 
  Delbracio (\citeyear{OteDel16-IPOL}) use a
  non-optimal value for this relative weighting parameter $(\gamma = 1/2$), instead of
  using either $\gamma = 1/3$, which gives the best numerical
  approximation to rotational symmetry (Proposition~4.16 in
  Lindeberg \citeyear{Lin93-Dis}), or $\gamma = 0$, which leads
  to separable convolution operators on a Cartesian grid
  (Proposition~4.14 in Lindeberg \citeyear{Lin93-Dis}), which then
  also implies
  that the discrete scale-space representation can be computed
  using separable convolution with the 1-D discrete analogue of the
  Gaussian kernel (\ref{eq-disc-gauss}) that is
  used in this work.
  
{\footnotesize
\bibliographystyle{abbrvnat}
\bibliography{bib/defs,bib/tlmac}}

\end{document}